%% file: main.tex
\documentclass{report}

\usepackage[english]{babel}

\usepackage[backend=biber, style=nature]{biblatex}
\usepackage{svg}
\usepackage{graphicx}
\usepackage{subcaption}
\usepackage{amsmath, amssymb}
\usepackage{hyperref}
\hypersetup{colorlinks=true,linkcolor=blue,citecolor=blue,urlcolor=black,
linktocpage=true}
\usepackage[capitalise]{cleveref}
\usepackage[version=4]{mhchem}
\usepackage{float}
\usepackage{todonotes}

\usepackage{bbm}
\usepackage{gensymb}
\usepackage{algorithm}
\usepackage{geometry}
\usepackage{algpseudocode}
\usepackage{fancyhdr}
\usepackage{makecell}
\usepackage{xcolor}
\usepackage{booktabs}
\usepackage{indentfirst}

\usepackage{subfiles}

\title{  	{ \includegraphics[scale=.5]{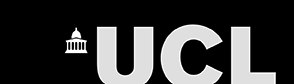}}\\
{{\Huge Climate Change Policy Exploration \\using Reinforcement Learning}}\\
{\large Learning and Interpreting Trajectories in a World-Earth System model}\\
		}
\date{Submission date: September 2022}
\author{Authored by\\\textbf{Theodore Wolf}\thanks{
{\bf Disclaimer:}
This report is submitted as part requirement for the MSc Machine Learning at UCL. It is
substantially the result of my own work except where explicitly indicated in the text.
 The report may be freely copied and distributed provided the source is explicitly acknowledged.
}
\\ \\ \\ \\ \\ \\ \\ \\
Research Project for\\
\textbf{MSc Machine Learning}
\\ \\ 
Supervised by \\
\textbf{Dr. Mar\'{i}a P\'{e}rez Ortiz, Prof. John Shawe-Taylor}}

\begin{document}

\maketitle

\newpage
\thispagestyle{empty}
\section*{Acknowledgement}
I would like to thank Mar\'{i}a and John for their unwavering support throughout this work. Special thanks go to Felix, who helped me understand his work better, which was invaluable for my research. I also thank all my peers at the UCL Centre for Artificial Intelligence for valuable input and interesting discussions.

\newpage

\begin{abstract}

\subfile{chapters/0.abstract}
\end{abstract}
\maketitle

\tableofcontents
\newpage
\chapter{Introduction}\label{chapter:intro}

\subfile{chapters/1.introduction}

\chapter{Background}\label{chapter:background}

\subfile{chapters/2.reinforcement_learning}

\chapter{Methods}\label{chapter:methods}

\subfile{chapters/3.methods}

\chapter{Experiments}\label{chapter:experiments}

\subfile{chapters/4.experiments}

\chapter{Discussion and Conclusion}\label{chapter:discussion}

\subfile{chapters/5.discussion}

\sloppy
\printbibliography

\subfile{chapters/6.appendix}

\end{document}

%% file: chapters/0.abstract.tex
Climate Change is an incredibly complicated problem that humanity faces. When many variables interact with each other, it can be difficult for humans to grasp the causes and effects of the very large-scale problem of climate change. The climate is a dynamical system, where small changes can have considerable and unpredictable repercussions in the long term. Understanding how to nudge this system in the right ways could help us find creative solutions to climate change. 

In this research, we combine Deep Reinforcement Learning and a World-Earth system model to find, and explain, creative strategies to a sustainable future. This is an extension of the work from Strnad et al. \cite{Strnad2019DeepStrategies}, where we extend on the method and analysis, by taking multiple directions. We use four different Reinforcement Learning agents varying in complexity to probe the environment in different ways and to find various strategies. The environment is a low-complexity World Earth system model where the goal is to reach a future where all the energy for the economy is produced by renewables by enacting different policies. We use a reward function based on planetary boundaries that we modify to force the agents to find a wider range of strategies. To favour applicability, we slightly modify the environment, by injecting noise and making it fully observable, to understand the impacts of these factors on the learning of the agents. 

We discover that our Reinforcement Learning agents learn to reach a carbon-free future, but for some initial conditions, strong feedback loops in the system prevent the agents from being able to control the environment. We find that in this simplistic model, the growth of the economy is a significant feature for the agents when deciding which policies to enact. While this approach shows some promise, there is much more to be done for this framework to be applicable to climate related policy. We hope this novel research will inspire more work on this topic.

%% file: chapters/1.introduction.tex
\section{Motivation}
\subsection{A Safe Space for Humanity}
“It is unequivocal that human influence has warmed the atmosphere, ocean, and land. Widespread and rapid changes in the atmosphere, ocean, cryosphere, and biosphere have occurred.” according to the latest \textit{Intergovernmental Panel on Climate Change} (IPCC) report \cite{IPCC2021SummaryPolicymakers}. 

The relatively stable climate of the last 10-12 millennia, the Holocene, has seen the majority of human advancement: from the first agricultural societies to the start of the industrial revolution. The geological period before the Holocene, is known as the Pleistocene, colloquially known as the \textit{Ice Age} \cite{Walker2018FormalStages/subseries}. Recently, due to the consequential effect of humans on the climate, scientists have started naming the era of human industrialisation the \textit{Anthropocene} \cite{Crutzen2002GeologyMankind, Waters2016TheHolocene}. This proposed geological era is characterised by the effects of humans on the environment and the climate. 

\textit{Tipping Points} refer to certain thresholds at which small perturbations can drastically change a system \cite{Lenton2008TippingSystem}. For the climate, different tipping elements can cause dramatic changes to the overall climate balance. According to scientists, the climate of the Holocene has a delicate balance \cite{Lenton2008TippingSystem, Steffen2018TrajectoriesAnthropocene}, one that should not be perturbed lest disastrous consequences for life on Earth, which would be difficult to counteract. It is understood that the climate is a part of a complex coupled socio-ecological system that is also adaptive
\cite{Berkes1998LinkingSustainability}. So, any actions  taken by humans can have many cascading and unpredictable effects \cite{Steffen2018TrajectoriesAnthropocene}. Therefore, the coupled system has to be considered when constructing policy that relates to the climate \cite{Levin2013Social-ecologicalImplications}.

To mitigate the effects of future climate change, scientists recommend staying within \textit{Planetary Boundaries} \cite{Rockstrom2009AHumanity, Steffen2015PlanetaryPlanet}. These boundaries define limits for measurable quantities, that are linked with significant tipping points, such as keeping warming below $+2\degree C$ above pre-industrial levels. Crossing planetary boundaries can provoke \textit{feedback loops} (or “snowballing” effects) in the climate system that can lead to rapid and uncontrollable climate destabilisation \cite{Steffen2018TrajectoriesAnthropocene}.

\subsection{Climate Change as a Control Problem}

The IPCC uses SSP (Shared Socioeconomic Pathways) to make predictions on the possible pathways for the climate given different behaviours of humans. There are five pathways described, ranging from heavily reducing Green House Gas (GHG) emissions, to heavily increasing them. Each potential behaviour has consequences for the Earth and the life on it. Predicting these pathways requires models that have incredibly high uncertainties attached to them, since modelling all the complexities of the Earth is infeasible. The IPCC does not give exact probabilities for each scenario; instead, they predict using terms such as “very high probability”, “low probability”, “medium probability”. These SSP are generated by \textit{Integrated Assessment Models} (IAM). These models encompass a wide range of scientific knowledge in different fields such as climate, biology, and economics \cite{Wang2017IntegratedEconomics}.

We know that the weather system and, consequently, the climate are dynamical systems where small changes in initial conditions can lead to startlingly different answers over time \cite{EdwardN.Lorenz1963DeterministicFlow,IPCC2021ClimateSummary}. Dynamical systems are deterministic. If the initial conditions are perfectly known, then we can predict outcomes accurately. Having fully accurate data and models is not technically possible, so meteorologists start multiple simulations at different initial conditions and then average the answers, also called ensembles models. This is the approach also taken by the IPCC to generate SSP, they use information from multiple IAMs from the \href{https://www.iamconsortium.org/}{\textit{Integrated Assessment Model Consortium}} and average them out. In dynamical systems, this can yield an incredibly wide range of answers with enough time, which is why weather predictions are often inaccurate further in the future.

Climate is defined as the “general weather conditions usually found in a particular place”\footnote{Cambridge Dictionary}. The global climate is then the general weather patterns of the Earth. The global climate is easier to predict than the weather over long periods, since it is an average of the weather over long periods. However, due to human influence, this system is being perturbed very fast \cite{Donges2017ClosingScience:}. Adding perturbations to a dynamical system makes the system even harder to predict. Perturbations can include a change in human carbon emissions, which has cascading and unpredictable effects. Ideally, if we can predict accurately, we can control accurately. We would like to control the climate to negate its most disastrous effects for humans by staying within planetary boundaries. Control would be done by policymakers and worldwide organisations by changing the actions of humans to enable a sustainable future.

The task of keeping a dynamical system within certain boundaries is called \textit{Optimal Control Theory} (OCT), often referred to more simply as \textit{control} \cite{Kamien2012DynamicManagement}. This sub-field of mathematics looks at methods that aim to optimise some objective function in a system by controlling the variables of the system. In the case of the climate, we would like to minimise climate destabilisation by staying within planetary boundaries. 

\section{Context of this Work}
\subsection{Reinforcement Learning for Control}

Recently, considerable leaps forward in control have been achieved thanks to Reinforcement Learning (RL). RL is a subfield of machine learning where an \textit{agent} learns to interact with an \textit{environment}, with \textit{actions}, to maximise a \textit{reward} signal. The leaps forward can be attributed to the advances in Deep Learning \cite{Krizhevsky2012ImageNetNetworks}. Due to this, RL agents have started performing superhuman feats, first in computer games that have easily definable rewards (score) and actions (controller inputs) \cite{Mnih2015Human-levelLearning}. More complexity has been added and is enabling agents to learn more and more intricate environments, such as the game of Go \cite{Silver2016MasteringSearch}, where the number of possible states is larger than the number of atoms in the universe. As Deep RL gains popularity, it starts being successfully used in real-world scenarios, such as the magnetic control of hydrogen plasma in a fusion reactor \cite{Degrave2022MagneticLearning}. RL can be used in any simulation environment where a goal can be defined. These agents are increasingly complex and execute a series of creative actions for control that even experts do not find intuitive \cite{Vargas2019CreativityLearning}.

RL's superhuman ability to find pathways in systems and flexibility has spurred scientists to use it in a range of tasks such as video compression \cite{Lu2021ReinforcementSensing} and power grid optimisations \cite{Marot2020L2RPN:Design}. It is easily applicable to any control environment or simulation. This includes climate models. 

By applying RL to climate models, there is the potential to find solutions to the challenge of climate change that are difficult to predict by human intuition: it has been shown that humans often underestimate the effects of exponential growth, known as Exponential Growth Bias \cite{Levy2017Exponential-growthOverconfidence, Schonger2021IntuitionComplexity}. Exponential growth is the solution to the most basic dynamic equation: 
\begin{align}
    \frac{dA(t)}{dt} = kA(t)
    \Longleftrightarrow A(t) = Ce^{kt},
\end{align}
where $k, C$ are constants. Recently, Exponential Growth Bias has been shown to have an impact on the real world with the Covid-19 pandemic \cite{Lammers2020CorrectingDistancing, Banerjee2021Exponential-growthCOVID-19}, where the exponential increase in infections was underestimated by the public and consequently affected individual response, which is believed to have contributed to higher infection rates. Human intuition struggles with the concept of exponential growth and consequently dynamical systems, so having algorithms inform climate change related policy could be vital.

\subsection{Previous Work}
Combining Machine Learning and climate research is an emerging field, that shows tremendous promise. Rolnick et al. \cite{Rolnick2019TacklingLearning} summarise many approaches for combining the two fields in a landmark paper written by leaders in their respective fields, both from academia and industry. Chapter 7 of this paper details approaches for constructing data-driven climate models for better prediction and that reduce the huge computational cost of the current climate models. Data-driven models, that can accurately predict outcomes, are possible thanks to the enormous wealth of data generated by satellites and the current climate models. Bury et al. \cite{Bury2021DeepPoints} use Deep Learning techniques to identify early warning signs for climate tipping points. Work such as from Mansfield et al. \cite{Mansfield2020PredictingLearning}, who use Gaussian Processes for climate prediction, show that advanced Machine Learning techniques can indeed be used for climate modelling. Making climate models computationally cheaper with Machine Learning enables more intricate components to be included, such as social variables, to create \textit{Social-Climate} models \cite{Moore2022DeterminantsSystem}. Including social variables could be vital for constructing policy relating to climate, where additionally, game theory can be included to create a wide range of scenarios \cite{Bury2019ChartingModel, Menard2021WhenInequality, MOTESHARREI201490}. 

As far as we know, the idea of combining Deep RL and Social-Climate models was only ever explored by Strnad et al. \cite{Strnad2019DeepStrategies} using \textit{World-Earth System} models, a different nomenclature for Social-Climate models. Two models were used by Strnad et al., one from Kittel et al. \cite{Kittel2017FromManagement}, and a higher dimension environment adapted from Nitzbon et al. \cite{Nitzbon2017SustainabilityModel}. The authors of Strnad et al. apply variants of the Deep RL algorithm \textit{DQN} (Deep Q-Learning) to these models and analyse how the agents learn strategies that lead to a sustainable future. The agents learn to “solve” the environments and achieve a sustainable future through various actions, but lack interpretability, as it is not made clear why some actions are taken rather than others. 

\section{Approach}
Overall, the work we conduct here follows strongly from Strnad et al. as we are using the same environment and a similar approach. In this work, we take a deeper dive into the decision-making of the RL agents compared to Strnad et al., as well as introducing a different set of agents: actor critics. By analysing the policy and value functions of these agents, we hope to draw meaningful conclusions. In addition, we present experiments that aim to extend the real-world applications of our agents. We do this by changing components of the environment without altering its fundamental dynamics:
\begin{itemize}
    \item We use the same environment and reward setup as Strnad et al. for comparison.
    \item Then, we add a cost to using actions that manage the environment to see how the agent learns in an environment where enacting policy has a cost attached to it.
    \item Next, the reward function is radically changed to make rewards more sparse. This affects the learning of the agents in various ways.
    \item To test the applicability of our framework, noise is introduced in the model to see how the agents learn with non-deterministic environment parameters.
    \item We make the environment fully observable to compare with the previous partially observed environment from Strnad et al. and see how the learning of the agents changes.
\end{itemize}
These experiments aim to test the limits of the learning of the agents. We present many plots to evaluate the learning of our agents, and understand their decision-making.
This is a very novel research area, and therefore we will not be able to answer all the questions this work will raise. We will focus on the following questions; 

\begin{itemize}
    \item To what extent can Reinforcement Learning agents inform us about climate change?
    \item Can they help us explain how to navigate its feedback loops? 
\end{itemize}

\subsubsection{Document Outline}
In this research, we first present in Chapter \ref{chapter:background} the basics of Reinforcement Learning that are required for this work. We attempt to be as succinct as possible while giving all the relevant background. In this same chapter, we introduce dynamical systems and World-Earth system models, which are useful for a more profound understanding of this work. 

\noindent Then, in Chapter \ref{chapter:methods}, we explain our methods in detail: we describe the agents we use as well as summarise the environment we use for our agents: the AYS model. We show the similarities of our methods to those used in Strnad et al. 

\noindent We follow this in Chapter \ref{chapter:experiments} by showing the results of our experiments, and analysing these in detail, highlighting specific trends. 

\noindent We will discuss these results in Chapter \ref{chapter:discussion}, attempt to draw meaningful conclusions from them, and propose extensions to this work in the final chapter.

%% file: chapters/2.reinforcement_learning.tex
\section{A Primer on Reinforcement Learning}
In this section, we will go through the essentials of Reinforcement Learning (RL) that are required for this work but, for a full introduction to the topic, \textit{Sutton \& Barto's} book \cite{Sutton2020ReinforcementIntroduction} is highly recommended.

\subsection{Markov Decision Process}
The Markov Decision Process (MDP) is the basis of most RL and is defined as a tuple $(\mathcal{S}, \mathcal{A}, p, r, \gamma)$, where:
\begin{itemize}
    \item $\mathcal{S}$ is the set of all possible states,
    \item $\mathcal{A}$ is the set of all actions,
    \item $p(s'|s, a)$ is the probability of transition to $s'$, given a state $s$ and action $a$,
    \item $r:\mathcal{S}\times\mathcal{A}\to \mathbb{R}$ is the expected reward achieved on a transition starting in $(s,a)$:
    \begin{align*}
        r=\mathbb{E}[R|s,a],
    \end{align*}
    \item $\gamma \in[0,1)$ is a discount factor that trades off later rewards to earlier ones
\end{itemize}
MDP's follow the Markov property: the present is independent of the past. Each state has all the relevant information from the history. The goal in RL is to solve the MDP, equivalent to finding the optimal \textit{policy} $\pi^*$ to maximise the reward that an \textit{agent} can collect in an \textit{environment}. A policy is a mapping $\pi : \mathcal{S}\times\mathcal{A}\to[0,1]$ that for every state \textit{s} assigns for each action $a\in\mathcal{A}$: the probability of taking that action in state \textit{s}. We denote it by $\pi(a|s)$ such that $a_t\sim\pi(\cdot|s_t)$ or $a_t=\pi(s_t)$ for deterministic policies. Solving an MDP requires finding the best return:
\begin{align}\label{eq:2}
    G_t = \sum_{\tau=0}^{\infty}\gamma^\tau R_{t+\tau},
    \intertext{which we can define recursively:}
    G_t = R_t + \gamma G_{t+1}.
\end{align}
We can then assign a value to each state: the value function of a certain state is the expected return from that state, given a policy $\pi$:
\begin{align}
    V^\pi(s) = \mathbb{E}[G_t|S_t=s, \pi].
\end{align}
And the optimal value is the value that follows the optimal policy:
\begin{align}
    V^*(s) = \mathbb{E}[G_t|S_t=s, \pi^*].
\end{align}
We often call the sequential decision-making of the agent and feedback from the environment the \textit{agent-environment interface} and is summarised in the figure below.
\begin{figure}[H]
    \centering
    \includegraphics[scale=0.3]{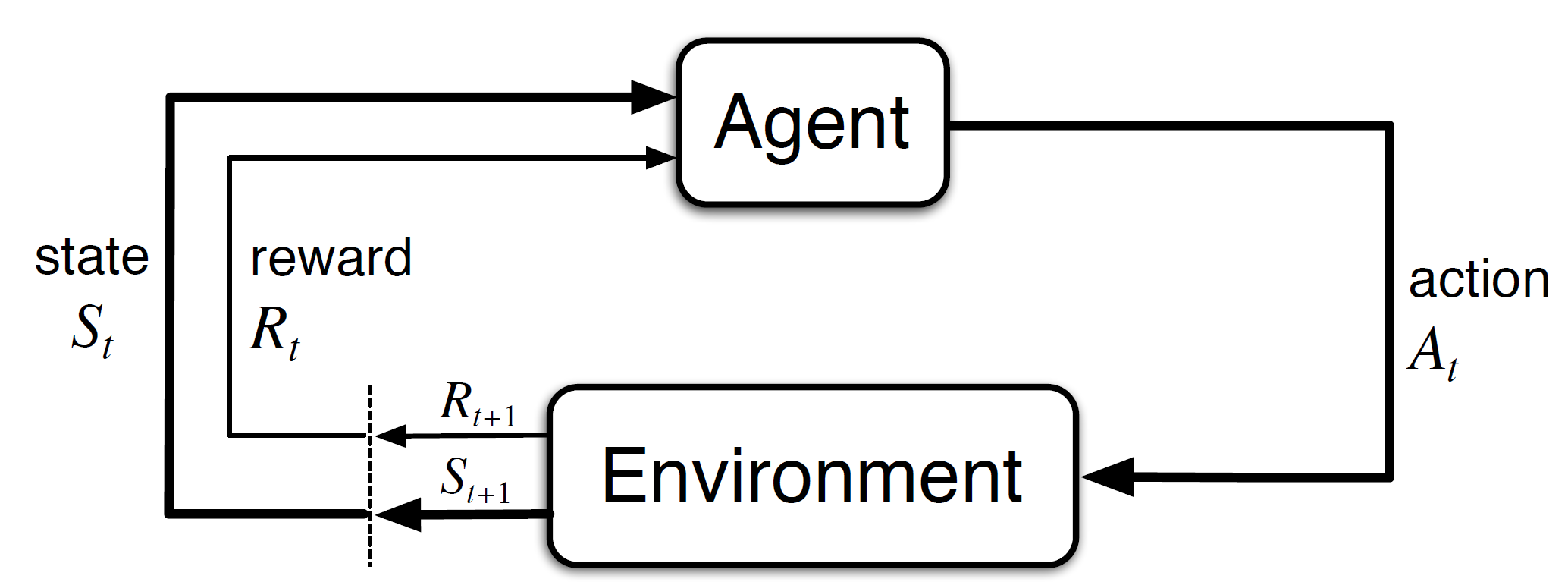}
    \caption{The agent-environment interface in a Markov Decision Process from Sutton \& Barto \cite{Sutton2020ReinforcementIntroduction}. At every time step, the agent observes a state and a reward from the environment and outputs an action.} 
    \label{fig:mdp}
\end{figure}
MDPs usually have \textit{terminal states}, when the agent reaches this state, the \textit{episode} ends.

\subsection{Bellman Equations}
Bellman's equations for optimality can solve an MDP: find the optimal policy in tabular cases \cite{Bellman1957DynamicStudy}. These equations use Dynamic Programming: they solve smaller sub-problems and then use the separate solutions to solve a larger problem. For an MDP, Bellman's equation has the following form, for a certain policy $\pi$:
\begin{align}
    V^\pi(s) = \sum_a\pi(s,a)\big[r(s,a) +\gamma\sum_{s'}p(s'|s, a)V^\pi(s')\big].
\end{align}
The value of a state $s$ is defined as the sum of the average reward obtained under the policy $\pi$ plus the discounted value of the next state under the same policy. The optimal version of this equation is found by taking a maximum over actions rather than an expectation, as we want to find the best possible return:
\begin{align}
    V^*(s) = \max_a[r(s,a) +\gamma\sum_{s'}p(s'|s, a)V^*(s')],
\end{align}
Dynamic Programming methods follow from the Bellman optimality equations. A common Dynamic Programming algorithm for RL is \textit{Value Iteration}:
\begin{align}
    v_{k+1}(s) \gets \max_a \mathbb{E}[R_{t+1}+\gamma v_k(s_{t+1})|S_t=s, A_t=a] \quad \forall s.
\end{align}
This algorithm always converges to the optimal value under tabular conditions (finite state space) and $\gamma<1$. $v_k(s)$ eventually converges, which implies that the optimal value function has been found. Acting greedily with respect to these values will yield the optimal policy. 

For this algorithm to converge, it needs to be applied to all states. As the state space increases, this becomes computationally infeasible, a common problem in machine learning known as the \textit{curse of dimensionality}, a term coined by Richard Bellman himself \cite{Bellman1957DynamicStudy}. Other methods need to be applied, such as sampling methods.

\subsection{State-Action Values}
Values are often represented with respect to both an action and a state, denoted $Q(s,a)$, this value is named the state-action value (or more simply Q-values) and is intimately related to the state value above:
\begin{align}
    V^\pi(s) = \sum_a \pi(a|s)Q^\pi(s,a) = \mathbb{E}[Q^\pi(s_t,a_t)|s_t=s, \pi]. \label{eq:vtoq}
\end{align}
We often use Q-values, as they provide insight as to which action will yield higher reward, helping the agent make a better choice. The Bellman equations can also be written with Q-values:
\begin{align}
    Q^\pi(s,a) &= r(s,a) + \gamma\sum_{s'}p(s'|a,s)\underbrace{\sum_{a'} \pi(s',a')Q^\pi(s',a') },\\
    &=r(s,a) + \gamma\sum_{s'}p(s'|a,s)\qquad\qquad V^\pi(s'),
    &\intertext{and similarly for optimality:}
    Q^{*}(s,a) &= r(s,a) + \gamma\sum_{s'}p(s'|a,s)\max_{a'}Q^{*}(s',a').
\end{align}

\subsection{Temporal Difference Learning}
One solution to break the curse of dimensionality is to sample states and then update with a one-step lookahead \cite{580874}:
\begin{align}
    \delta_t =& R_t +\gamma V^\pi(s_{t+1}) - V^\pi(s_t) \label{eq:td}\\
    V^\pi(s_t) \gets& V^\pi(s_t) + \alpha\delta_t,
\end{align}
where $\alpha$ is a learning rate that reduces the variance of updates. This is TD(0), a low variance but high bias update to the state value function. One option to effectively balance bias and variance is to bootstrap later than one step ahead. This is \textit{n-step TD}:
\begin{align}
    V^\pi(s_t) \gets V^\pi(s_t)+ \alpha\Big(G_{t:t+n} - V^\pi(s_t)\Big),
\end{align}
Where $G_{t:t+n}$ is the \textit{n-step return} and is defined similarly to the return in equation \eqref{eq:2} but bootstrapping earlier such that:
\begin{align}
    G_{t:t+n} = R_t +\gamma R_{t+1} +\gamma^2 R_{t+2}+...+\gamma^{n-1}R_{t+n-1} +\gamma^n V^\pi(s_{t+n}),
\end{align}
using the value estimate at state $s_{t+n}$ following policy $\pi$. The n-step return enables the agent to update the value of a state with what it has learned about the environment since visiting that state without having to wait for the end of the episode. 
The version that generalises these algorithms is TD($\lambda$) \cite{Sutton2020ReinforcementIntroduction}, that uses a parameter $\lambda$ to balance short and long-term rewards. We define the $\lambda$-return:
\begin{align}
    G_{t}^\lambda \dot{=}  (1-\lambda)\sum_{n=1}^\infty \lambda^{n-1}G_{t:t+n}.
\end{align}
The $\lambda$-return can be rewritten as a sum of TD(0)-errors:
\begin{align}
    G^\lambda_{t} = V^\pi(s_t)  + \sum_{i=t}^{\infty} (\gamma\lambda)^{i-t}\delta_i, \label{eq:lreturn}
\end{align}
where $\delta$ is the TD(0) error defined in Equation \eqref{eq:td}. 
By tuning $\lambda$ effectively, we can balance between the high variance/low bias of long horizon rewards and the low variance/high bias of short horizon rewards. When setting $\lambda=0$, we recover the traditional TD(0) and when $\lambda=1$, we recover the infinite horizon return from Equation \eqref{eq:2}.

\subsection{Q-Learning}
One of the oldest and most popular algorithms for Reinforcement Learning is Q-learning \cite{Watkins1989LearningRewards}. Q-learning emulates the Bellman Optimality equation directly as an update to state-action values, but breaks the curse of dimensionality by sampling instead of updating all states:
\begin{align*}
    Q(s_t, a_t) \gets R_t+\gamma\max_aQ(s_{t+1}, a)-Q(s_t, a_t).
\end{align*}
Q-learning is an \textit{off-policy} algorithm because the update is independent of what action was taken at the next time step (the action sampled by the policy). Therefore, the algorithm can be used for \textit{off-line} updates, which are much more data-efficient. The policy for choosing the action to take is an exploratory policy, usually $\epsilon$-greedy, which selects actions stochastically:
\begin{align*}
    \pi(a|s) = 
    \begin{cases}
    1-\epsilon + \epsilon/|\mathcal{A}| \qquad&\textrm{if } Q(s,a)=\max_bQ(s,b)\\
    \epsilon/|\mathcal{A}|   &\textrm{otherwise}.
    \end{cases}
\end{align*}
With $\epsilon\in [0,1]$ and $\lim_{t\to\infty} \epsilon= 0$ to converge to a better solution as exploration is needed for better convergence \cite{Watkins1992Q-learning,Watkins1992TechnicalQ-Learning}. In the tabular case (finite state space), there are many guarantees for optimal convergence of RL algorithms, but not for the continuous case, where the state space is infinite. 

\subsection{Policy Gradients}
The goal of RL is to learn the optimal policy, Q-learning is all about learning state-action values then acting greedy with respect to them. Why not learn the policy directly? This is the idea behind policy gradient methods. These methods parameterise the policy and then try to optimise it with gradient ascent. This can be done with the score function trick. We define $d(\mathcal{S})$ as the distributions of states and $\pi_\theta(a|s)$ as the parametrised policy, we want to maximise the expected reward given the distribution of states and the policy:
\begin{align}
    \max_\theta \mathbb{E}[r(s,a)|d(\mathcal{S}), \pi_\theta(a|s)] \nonumber
    &=\nabla_\theta\mathbb{E}[r(s,a)|d(\mathcal{S}), \pi_\theta(a|s)]\nonumber\\
    &=\nabla_\theta \sum_{s\in\mathcal{S}} d(s) \sum_{a\in\mathcal{A}} \pi_\theta(a|s) r(s,a)\nonumber\\
    &=\sum_{s\in\mathcal{S}} d(s) \sum_{a\in\mathcal{A}} \nabla_\theta \pi_\theta(a|s) r(s,a)\nonumber\\
    &=\sum_{s\in\mathcal{S}} d(s) \sum_{a\in\mathcal{A}} r(s,a)\pi_\theta(a|s) \frac{\nabla_\theta\pi_\theta(a|s)} {\pi_\theta(a|s)}\nonumber \\
    &=\mathbb{E}[r(s,a)\nabla_\theta\log \pi_\theta(a|s)|d(S), \pi_\theta(a|s)].
\end{align}

Where we have managed to push the gradient operator inside the expectation, meaning that we can sample this gradient to get an estimate of it. We can replace the reward $r(s,a)$ with the return $G_t$ (which we calculate at the end of an episode) or with a Q-value $Q(s,a)$. The intuition here is that we want to scale the update to the policy with the reward that was obtained. If we obtain a large reward, we incur a larger update. However, a large negative reward causes a large update in the opposite direction. This is the REINFORCE algorithm \cite{Williams1992SimpleLearning, Sutton1999PolicyApproximation}. Like other policy gradient algorithms, it is an \textit{on-policy} algorithm; the actions that the policy samples change the updates.

A common application of policy gradients are actor critics \cite{Sutton2020ReinforcementIntroduction}. They are a well-researched subcategory of hybrid policy gradient methods that lend themselves nicely to deep learning. They combine the ideas of value function learning and policy learning. The policy is parameterised (\textit{actor}) and receives additional information from a parameterised state value function (\textit{critic}). The actor takes actions and gets feedback on these actions based on how “good” the critic judges them to be. We will be using both actor critics and Q-learning inspired methods for our agents.

\section{World-Earth System Models}

\subsection{Dynamical Systems}
We briefly introduce Dynamical systems, as we will use the jargon specific to this field throughout this work.
Dynamical systems are models that use differential equations to model interactions \cite{EdwardN.Lorenz1963DeterministicFlow}. Their discovery is attributed to Lorenz, a meteorologist who named these types of systems \textit{non-periodic deterministic flows}, as they behave very erratically but are still fully deterministic. This behaviour is dubbed \textit{chaotic}, and is the origin of chaos theory. Below are plots of the Lorenz system, with variables $(x,y,z)$, this is called a \textit{phase space}, where variable values are plotted over time to form a \textit{trajectory}. The full system of equations is in Appendix \ref{Tables}.
\begin{figure}[H]
    \centering
    \begin{subfigure}[b]{0.49\textwidth}
        \centering
        \includegraphics[width=\textwidth]{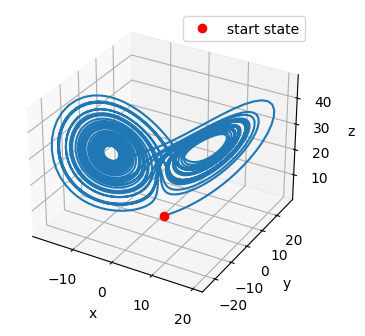}
        \caption{$\beta=2.7$}
        \label{fig:lorentz1}
    \end{subfigure}
    \hfill
    \begin{subfigure}[b]{0.49\textwidth}
    \centering
        \includegraphics[width=\textwidth]{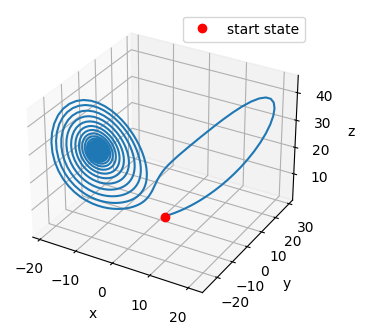}
        \caption{$\beta=5$}
        \label{fig:lorentz2}
    \end{subfigure}
    \caption{Phase space plots of the Lorenz system trajectory for a single initial state but with different $\beta$ values. Also called the Lorenz butterfly.}
    \label{fig:lorentz}
\end{figure}
    
In Figure \ref{fig:lorentz1}, we see that there are two regions that attract the trajectories, they are called \textit{attractors}. Depending on the system, some attractors are a single point and so are called \textit{fixed points}. Once a trajectory reaches a fixed point, the system stops evolving, it reaches an equilibrium, this is the case in Figure \ref{fig:lorentz2}. As shown here, changing parameters in dynamical systems can lead to very different attractors, here we changed one of the parameters that has caused the system to go from infinitely oscillating to reaching an equilibrium. 

\subsection{Earth to World-Earth}
\textit{Earth system models} (ESM) aim to model the interactions of the various components of the Earth using dynamical systems. These models use our current understanding of chemistry, physics, and biology to represent different processes that happen on Earth \cite{ESM}. For example, the effects of life on Earth are included in atmospheric projection. As such, they are more advanced than Global Climate Models (GCM), which only model the purely physical climate and ocean processes, without considering the impact of biological processes. 

ESM's are missing a crucial component. Geologists are currently considering calling the current period the Anthropocene due to the unprecedented and growing impact that humans are having on the environment. For models to be accurate, the human factor has to be taken into account. Hence, the human world is sometimes integrated in the ESM to create a different class of models called \textit{World-Earth system models}. The earliest model of this type is from the 1974 book; the “Limits to Growth” \cite{meadows1972limits}. The book describes different possible pathways for humanity with a dynamical model: \textit{World3}, that considers population growth, economy, and pollution. “World-Earth system model” is the nomenclature used by the Potsdam Institute for Climate Impact Research (PIK). This institute has produced much of the research for World-Earth system modelling, such as the AYS model \cite{Kittel2017FromManagement}, which we use in this work. Another nomenclature exists, Social-Climate models \cite{Bury2019ChartingModel, Moore2022DeterminantsSystem}. This research is still quite uncommon, and the naming has yet to be completely established. For clarity, we will call them World-Earth systems, as our work follows from Strnad et al. \cite{Strnad2019DeepStrategies}, who use this naming scheme. 

The AYS model from Kittel et al. \cite{Kittel2017FromManagement} is a low complexity World Earth system model. The AYS is mathematically defined and can be plotted in phase space due to only having three dimensions of variables, which adds to interpretability. The dynamics are fairly simple and depend on very few parameters. The few parameters that are present are justified coherently with data in the original paper. We will thus be using this model for the experiments.

%% file: chapters/3.methods.tex
In this chapter, we describe our methods. We attempt to be as descriptive as possible in order for our experiments to be reproducible. To assist this, the code was uploaded on  \href{https://github.com/TheodoreWolf/climate_policy_RL}{\textcolor{blue}{GitHub}}. A short demo is also available on \href{https://colab.research.google.com/drive/1pnklz9uI8NG2ZW_HKQsv9GVOgm42s0qq?usp=sharing}{\textcolor{blue}{Google Colaboratory}}.

\section{The Agents}
Following Strnad et al. \cite{Strnad2019DeepStrategies}, we use different agents to probe the Environment and discover strategies leading to sustainable development. We use a DQN agent \cite{Mnih2015Human-levelLearning} as well as one of its variants with improvements: Duelling Double DQN with Prioritised Experience Replay and Importance sampling \cite{Hessel2017Rainbow:Learning}. We call this algorithm \textit{DuelDDQN} for conciseness. We want to compare these to Actor Critic agents: A2C \cite{Mnih2016AsynchronousLearning} and PPO \cite{Schulman2017ProximalAlgorithms}. Variety in our population of agents enables us to explore and discover different strategies. The agents vary in both fundamentals and complexity. This also serves to compare off-policy and on-policy algorithms: DQN agents are off-policy and actor critics are on-policy.
These are all implemented with \textit{PyTorch} and hyperparameters were tuned using a Bayesian optimisation scheme. 

\subsection{DQN-Based Agents}
\subsubsection{Deep Q-Networks}
\begin{algorithm}[H]
\caption{Deep Q-learning with Experience Replay}\label{alg:dqn}
\begin{algorithmic}[1]
\State Initialise replay memory $\mathcal{D}$ to capacity \textit{N}
\State Initialise policy action-value function $Q$ with random weights $\theta$
\State Initialise target action-value function $\hat{Q}$ with weights $\theta^-=\theta$
\For{episode = 1, M}
\State Initialise sequence $s_0$ 
\For{t = 1, T}
\State With probability $\epsilon$ select a random action $a_t$
\State otherwise select $a_t = \max_a Q (s_t, a | \theta)$
\State Execute action $a_t$ and observe reward $r_t$, state $s_{t+1}$ and whether episode is \textit{done}
\State Store transition $(s_t, a_t, r_t, s_{t+1}, done_t)$ in $\mathcal{D}$
\State $s_{t} \gets s_{t+1}$
\State Sample random minibatch of transitions $(s_j, a_j, r_j, s_{j+1}, done_j)$ of size $J$ from $\mathcal{D}$ 
\State Set $y_j = \begin{cases}
r_j &\textrm{if $done_j$}\\
r_j + \gamma \max_{a'} \hat{Q}(s_{j+1}, a'|\theta^-) &\textrm{otherwise}
\end{cases}$
\State Perform a gradient descent step on $\frac{1}{J}\sum_j^J(y_j - Q(s_j , a_j | \theta))^2$
\State Every $C$ steps, $\theta^-\gets\theta$
\EndFor
\EndFor
\end{algorithmic}
\end{algorithm}
One of the first highly successful algorithms that combine deep learning and RL is “Deep Q-networks” (DQN) by Mnih et al. \cite{Mnih2015Human-levelLearning}, presented in Algorithm \ref{alg:dqn}. This algorithm keeps two Q-functions parameterised as neural networks, and selects actions with an $\epsilon$-greedy policy. The two Q-functions are called the policy network and the target network. The policy network selects actions and is updated at every iteration with both new experiences and old ones thanks to a replay buffer. The parameters of the policy network are periodically copied to the target network. This helps to reduce overestimation bias, a common problem in Q-learning \cite{VanHasselt2010DoubleQ-learning}.

The DQN agent we implement is very close to Algorithm \ref{alg:dqn}. We keep the $\epsilon$-greedy policy, but introduce a decay to the $\epsilon$ parameter to slowly converge to a greedy policy over time. This decay is controlled by the parameter $\rho$ such that $\epsilon(t) = \epsilon_0\times\frac{1}{t^\rho} + 0.01$ where $t$ is a variable that is incremented by one every time the agent takes an action and $\epsilon_0$ is chosen to be one. This scheme is inspired from Strnad et al. and DeepMind's UCL Reinforcement Learning course \cite{VanHasselt2021DeepMindYouTube}. It enables the agents to be entirely exploratory at first, and slowly converge to the greedy policy while still exploring occasionally. The exploration decay rate to a greedy policy $\rho$, is optimised as a hyperparameter. The periodic copying of the policy network parameters to the target network was optimised as $\tau$; the target network update frequency. The policy network is copied to the target network every $\frac{1}{\tau}$ updates. The batch size and the buffer size are also treated as hyperparameters.

\subsubsection{DuelDDQN with Prioritised Experience Replay}
Many subsequent iterations of the DQN algorithm have surfaced, improving various aspects of it \cite{Hessel2017Rainbow:Learning}. One of the most popular improvements is the addition of a Prioritised Experience Replay Buffer with Importance sampling (PER-IS) \cite{Schaul2015PrioritizedReplay}, where actions that incur a larger update (that are the most surprising) are more likely to be picked to be replayed. This does entail that importance sampling weights must be included to compensate for picking the same transition multiple times. We use $done_t$ as a boolean variable that indicates whether the episode is finished at time step $t$. The probability of a tuple of experience $(s_t, a_t, r_t, s_{t+1}, done_t)$ being sampled is its absolute error:
\begin{align}
    p_t = |\delta_t| = |r_t + \mathbbm{1}[\overline{done_t}]\,\gamma \max_a  Q(s_{t+1}, a) - Q(s_t, a_t)|.
\end{align}
Where we use the $\overline{done}_t$ notation as the negation of the $done_t$ boolean variable and $\mathbbm{1}[\cdot]$ as the indicator function\footnote{$\mathbbm{1}[x]=1$ if $x$ is True else 0 if $x$ is False}, a shorthand notation of line 13 in Algorithm \ref{alg:dqn}. A parameter $\alpha$ is used as an exponent: $P_t = p_t^\alpha$ with $\alpha\in[0,1]$. This is to tune the prioritisation with $\alpha=0$ corresponding to no prioritisation (equivalent to the buffer in algorithm \ref{alg:dqn}) and $\alpha=1$ to full prioritisation. These tuned probabilities are then normalised by taking the sum of the tuned probabilities in the replay buffer. Importance sampling weights then are:
\begin{align}
    w_t = \frac{(N  P_i)^{-\beta}}{\max_i w_i}\label{eq:weights},
\end{align}
with $N$ the number of experiences in the buffer and $\beta$ another hyperparameter $\beta\in[0,1]$, that is converged to 1 over time. Using PER-IS can speed up convergence and improve stability.

Another improvement is Double Q-Learning \cite{VanHasselt2010DoubleQ-learning}, in this version of the algorithm, the action selection and evaluation are decoupled in the update as such:
\begin{align}
    L_{Double}(\theta) = \Big(Q_\theta(s_t,a_t) - R_t +\mathbbm{1}[\overline{done_t}]\, \gamma \big[\big[Q_\theta(s_{t+1}, \arg\max_a Q_{\theta^-}(s_{t+1},a))\big]\big]\Big)^2 .
\end{align}
Here $[[.]]$ denotes stopping the gradient, so $\nabla_\theta[[f_\theta]]=0$, and we have simplified notation $Q(\cdot, \cdot|\theta) = Q_\theta(\cdot, \cdot)$
This type of update helps reduce overestimation bias in Q-learning and can dramatically improve performance \cite{VanHasselt2015DeepQ-learning}. 

The last improvement we look at, is the duelling network architecture \cite{Wang2015DuelingLearning}. Duelling architectures are branching neural networks that calculate values and advantages in two different streams. Advantages represent how much better an action is compared to the possible others, they are defined as:
\begin{align}
    A(s,a) &= Q(s,a) - V(s),
\end{align}
and the network outputs:
\begin{align}
    Q_{\theta, \psi, \omega}(s,a) &= V_{\theta,\psi}(s) + A_{\theta,\omega}(s,a) - \frac{1}{|\mathcal{A}|}\sum_{i}^{|\mathcal{A}|} A_{\theta,\omega}(s,a_i) \label{eq:advantage}
\end{align}
Where $\omega$ is the set of parameters associated with the advantage stream, $\psi$, the set of parameters for the value stream and $\theta$, the parameters shared across both streams. Regular DQN neural network architecture and duelling architecture are represented below.
\begin{figure}[H]
    \centering
    \includegraphics[scale=0.4]{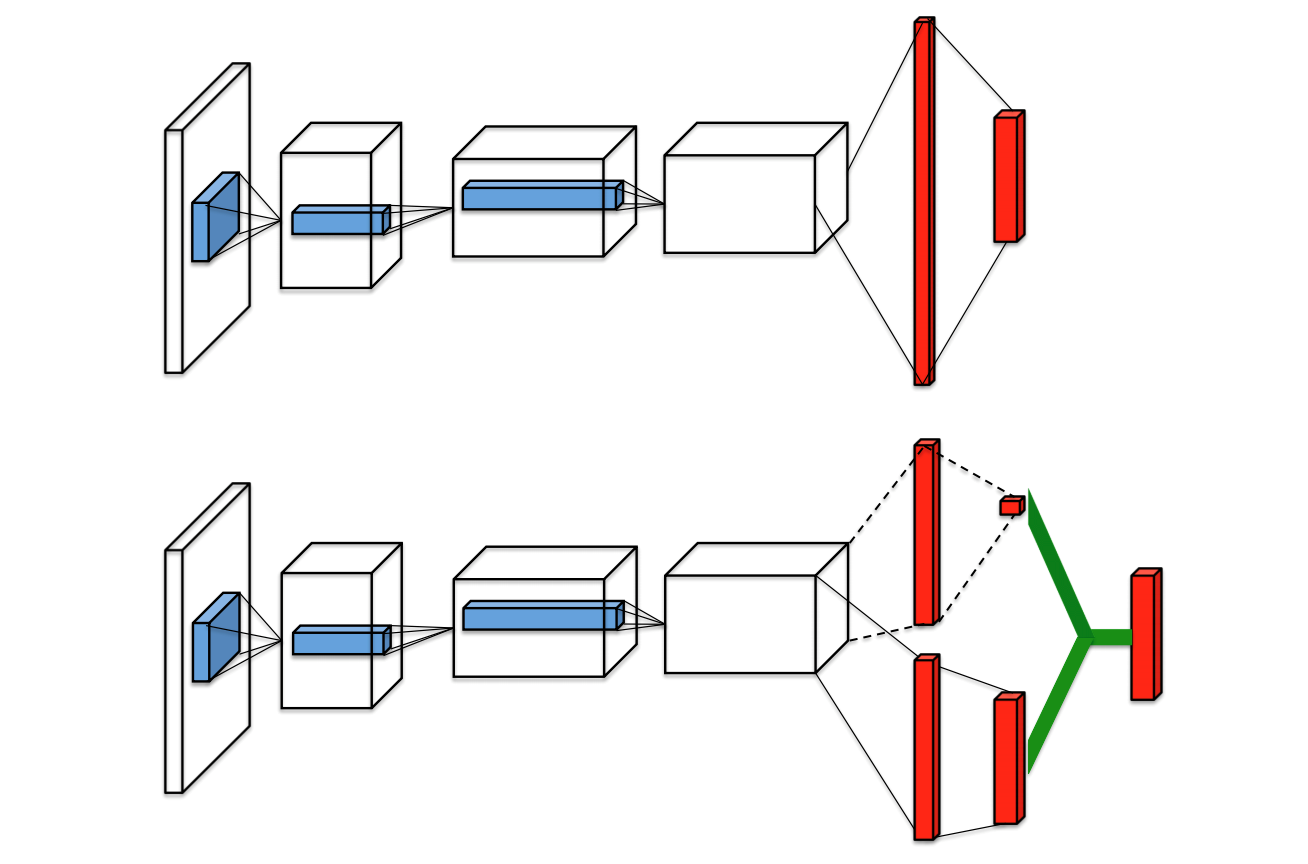}
    \caption{Architecture of a regular Deep Q-network (top) and duelling architecture (bottom) from Wang et al. \cite{Wang2015DuelingLearning}. In the bottom panel, the network branches into two streams, the top stream outputs the scalar state value $v(s_t)$ and the bottom  stream outputs the $|\mathcal{A}|$-length vector of advantages $A(s_t, a_t)$. The streams are then combined according to Equation \eqref{eq:advantage} to output $Q(s_t,a_t)$.}
    \label{fig:duel_net}
\end{figure}
This improvement is small, but leads to better policy evaluation when there are multiple similar valued actions \cite{Wang2015DuelingLearning}.

We take our DQN implementation as it is and add the upgrades described above. We add a duelling network architecture and the double Q-learning update. The algorithm was also run with a Prioritised Experience Replay with Importance Sampling buffer. This was implemented with two binary trees where the root of each tree respectively saved the sum of all the probabilities, and the minimum of all the probabilities. This was to compute the normalised probabilities and the importance sampling weights efficiently. Despite this, the training time of this agent scaled significantly with increasing buffer size. The agent is thus very similar to RAINBOW from Hessel et al. \cite{Hessel2017Rainbow:Learning}. The buffer prioritisation $\alpha$ and importance sampling exponent $\beta$ were optimised as hyperparameters. The resulting algorithm is a Duelling Double-DQN with Prioritised Experience Replay and Importance Sampling agent, which we call \textit{DuelDDQN} for short.

\subsection{Actor Critic Agents}\label{section:deep actor critics}
\subsubsection{Advantage Actor Critic}

The parametrisation of both the policy and the value function in actor critics can easily be converted to deep learning. Some of the most famous algorithms are Advantage Actor Critic (A2C) and Asynchronous Advantage Actor Critic (A3C) \cite{Mnih2016AsynchronousLearning}. The following objective function is maximised to optimise the parameterised policy, known as the \textit{actor}: 
\begin{align}
    J_\theta(\pi) = \mathbb{E}[A^\pi(s_t, a_t) \log\pi_\theta(a_t|s_t)] \label{eq:actor_objective}.
    \intertext{Where, one option is to use the one-step temporal difference TD(0) error for $A^\pi(s_t, a_t)$:}
    A^\pi(s_t,a_t)=\delta_t = R_t +\gamma v_w(S_{t+1}) -v_w(S_t) \label{eq:tdadv}.
\end{align}
With $v_w(\cdot)$, the \textit{critic}, which is a function of the state, parameterised as a neural network. We thus have two sets of network parameters that we need to update: the actor and the critic. For neural network frameworks, we need to maximise the objective \eqref{eq:actor_objective} which is equivalent to minimising the loss:
\begin{align}
    L_{actor}(\theta) = -\frac{1}{T}\sum_{t=0}^T A^\pi(s_t,a_t) \log \pi_\theta(a_t|s_t) . \label{eq:actor_loss}
\end{align}
For updating the critic, we can use a similar scheme as in DQN and minimise a squared error between $v_w(s_t)$ and $R_t+\gamma v_w(s_{t+1})$:
\begin{align}
    L_{critic}(w) &= \frac{1}{T}\sum_{t=0}^T\Big(R_t + \gamma [[v_w(s_{t+1})]]-v_w(s_t)\Big)^2.\label{eq:critic}
\end{align}

For our Advantage Actor Critic (A2C), we aim to make a simple agent, to contrast with our other Actor Critic; PPO. 
For A2C, the actor network is implemented with a neural network that takes as input the state and outputs four unnormalised probabilities corresponding to the probability of taking each action. The probabilities are then normalised and used for a multinomial distribution, from which we sample to get the action for the time step. 
The advantage function used is the TD(0) error, which values short-term rewards, has high bias, and low variance. This is the simplest form of the advantage function. We also included entropy regularisation, controlled by an exploratory parameter $\epsilon$. The losses then have the form:
\begin{align}
    L_{actor}(\theta) &= -\frac{1}{T}\sum_{t=0}^T\Big[\big(R_t +\mathbbm{1}[\overline{done_t}]\, \gamma v_w(s_{t+1})-v_w(s_t)\big)\log \pi_\theta(a_t|s_t)\Big] - \epsilon \mathcal{H}(\pi_\theta)\\
    L_{critic}(w) &= \frac{1}{T}\sum_{t=0}^T\Big(R_t +\mathbbm{1}[\overline{done_t}]\, \gamma [[v_w(s_{t+1})]]-v_w(s_t)\Big)^2.
\end{align}
Where we include the same notation $\mathbbm{1}[\overline{done}]$ to make the next state value equal to 0, if we are at a terminal state. Entropy regularisation (maximising the entropy) ensures that the policy stays exploratory by forcing the action distribution to be more uniform. It is known to improve the final policy \cite{Mnih2016AsynchronousLearning}. The entropy coefficient $\epsilon$ is optimised as a hyperparameter.
The A2C agent was implemented with a rollout buffer (which stores the transitions in order). The agent steps through the environment multiple times, then once the rollout buffer is full, the agent is updated with the entire buffer of experience at once with no batching. The reason no batching is done is that A2C is an on-policy algorithm. So if an update is applied to the network, the policy that collected the experience is no longer the same as the one that has now been updated. Updating a second time with the experience from the previous policy yields incorrect gradient updates. We treat the rollout length as a hyperparameter.
Finally, we include gradient clipping so that the policies or the value estimates do not change too abruptedly. 

\subsubsection{Proximal Policy Optimisation}
Actor Critic methods can lead to large updates to the policy, which can negatively affect performance. To remedy this, Schulman et al. \cite{Schulman2017ProximalAlgorithms} maximise a clipped objective for the actor instead of Equation \eqref{eq:actor_objective}, equivalent to minimise the following loss:
\begin{align}
    L^{CLIP}(\theta) = -\mathbb{E}\Big[\min\Big(\frac{\pi_\theta(a_t|s_t)}{\pi_{\theta_{old}}(a_t|s_t)}A^\pi(s_t, a_t), clip(\frac{\pi_\theta(a_t|s_t)}{\pi_{\theta_{old}}(a_t|s_t)}, 1-\eta, 1+\eta)A^\pi(s_t, a_t)\Big)\Big]\label{eq:ppoloss},
\end{align}
where $\eta$ is a hyperparameter that decides how far away the new policy can go from the old policy. If there is a large difference between the old policy and the new policy, the policy is clipped, which truncates the ratio to the range $[1-\eta, \,1+\eta]$. The critic is updated similarly to Equation \eqref{eq:critic}. This algorithm, Proximal Policy Optimisation (PPO), has yielded some of the best results over a wide range of benchmarks, such as the MuJoCo environments \cite{Schulman2017ProximalAlgorithms}.

We implemented PPO following Equation \eqref{eq:ppoloss}, the clipping value is optimised with Bayesian optimisation. Otherwise, the implementation is very similar to the A2C, for easier comparison, such as the entropy regularisation and the rollout buffer.
We also added changes to the PPO algorithm following Huang et al. \cite{Huang2022TheOptimization} who summarise 37 implementation details for PPO. We implement many, but not all of the 37 details that are described. For example, we did not include the clipped loss for the value function which, according to the authors, can hurt performance. The significant changes we implement are the normalisation of the advantages of each batch; by subtracting the mean of the advantages in the batch and dividing by the standard deviation: this reduces the variance of policy updates \cite{Andrychowicz2020WhatStudy}. We implement Generalised Advantage Estimation from Schulman et al. \cite{Schulman2015High-DimensionalEstimation}, defined with the TD($\lambda$) return of Equation \eqref{eq:lreturn}:
\begin{align}
    A_t^\lambda &=  G^\lambda_{t}-V^\pi(s_t)\\
    &= \sum_{i=t}^{t+n-1} (\gamma\lambda)^{i-t}\delta_i,
    \intertext{which we define recursively:}
    A_t^\lambda &=  \delta_t + \gamma\lambda A^\lambda_{t+1}.
\end{align}
Which enables to balance between bias and variance of advantages. We then have access to a TD($\lambda$) target for updating the critic:
\begin{align}
    G_t^\lambda = A_t^\lambda + v_w(s_t).
\end{align}
Our losses to update the networks then have the form:
\begin{align}
    L_{actor}(\theta) =& -\frac{1}{T}\sum^T_{t=0}\Big[\min\Big(\frac{\pi_\theta(a_t|s_t)}{\pi_{\theta_{old}}(a_t|s_t)}A^\lambda(s_t, a_t),
    clip(\frac{\pi_\theta(a_t|s_t)}{\pi_{\theta_{old}}(a_t|s_t)}, 1-\eta, 1+\eta)A^\lambda(s_t, a_t)\Big) \Big]\nonumber\\ 
    &- \epsilon \mathcal{H}(\pi_\theta)\\
    L_{critic}(w) =& \frac{1}{T}\sum_{t=0}^T\Big(G_t^\lambda-v_w(s_t)\Big)^2.
\end{align}
Because of the clipping of excessive policy changes, the agent can be updated in epochs and batches \cite{Schulman2017ProximalAlgorithms}. We use 50 epochs which was found to set a good compromise between computation time and performance, we optimise the rollout length and the size of the baches in the hyperparameter tuning.
\\

We use these four agents in the environment with a discount $\gamma=0.99$ as we are interested in long-term rewards rather than short-term rewards \cite{Sutton2020ReinforcementIntroduction}.

\subsection{Neural Network Architectures}
A specific architecture for the neural networks is used, that sets a good compromise between accuracy and training speed. The architecture parameters are not optimised with the other hyperparameters due to limited compute power. The input to all the networks is the observed state space $(A,Y,S)^T$, which we elaborate on in Section \ref{section:AYS}. Three layers of 256 dense Linear Units are used, intercalated with Rectified Linear Units\footnote{ReLU$(x) = \max(0, x)$} (ReLU). The number of units follows from Strnad et al., and so is the use of ReLU's, other nonlinear layers were not investigated. The last layers output four values (four actions preferences or four state-action values) or one scalar value (state value) depending on the network. Training times for agents can vary from 5 minutes to 20 minutes for 100000 time steps (\textit{frames}) of the environment on a standard GPU-equipped machine. An episode in this environment can last from 20 to 600 frames, depending on the performance of the agent.

\subsection{Optimisers and Schedulers}
For all the neural networks, we use the Adam Optimiser \cite{Kingma2014Adam:Optimization}, which is commonly used for Reinforcement Learning tasks \cite{Andrychowicz2020WhatStudy}. We also make use of learning rate schedulers, recommended for Deep RL \cite{Huang2022TheOptimization}. These enable us to reduce the learning rate geometrically throughout training to help convergence of the networks. There is a risk of underfitting, if the learning rate is annealed too fast and stops the agent from converging to a suitable optimum, which makes the decay rate tuning difficult. This can happen if the agent does not explore fast enough. To set a compromise, a geometric decay rate of 0.5 was applied at regular intervals in the training. The number of times the learning rate is decayed is treated as a hyperparameter, named \textit{decay number}. For example, a learning rate with a decay number of 4 will have a final learning rate multiplied by $0.5^4$. Ideally, we would like to tune the geometric factor of 0.5 as well, but this was found to be very difficult in practice due to limited compute power.

\subsection{Hyperparameter Tuning}
Great care was taken in tuning hyperparameters due to the difficult environment that the agents have to solve. The parameters that were tuned are summarised in Tables \ref{table:dqn} and \ref{table:ac} below. To tune the hyperparameters, we used a Bayesian optimisation tool provided by the \href{https://wandb.ai/site}{\textit{Weights and Biases}} python library. The parameters were sampled from categorical distributions or uniform distributions. For parameters that can span multiple orders of magnitude (such as learning rates), we sample from a log-uniform distribution. The full descriptions of distributions and ranges are in Appendix \ref{Tables}. The objective given to the optimisation was to maximise the mean reward over 100 000 steps. Steps are a single iteration of the environment, this is equivalent to 500-2000 episodes, depending on average episode length. Maximising the mean reward optimises for two things: the agents find a fast way to learn the environment, as well as finding a way to maximise rewards. What the mean reward doesn't account for is stability, it cannot tell the difference between an agent that learns fast to obtain high rewards but is unstable or an agent that learns slowly to obtain a high reward but with high stability. To circumvent this, at the end of the optimisation, the reward curves of the top-performing parameters were qualitatively evaluated to not select parameters that yield an unstable agent.

For consistency in the optimisation scheme, we use a fixed global random seed for all the runs. The seed controls the initialisation of the weights of the neural networks, the sampling of actions, the environment initial states, and the sampling from the replay buffer or the shuffling of batches in the rollout buffer. This could affect the optimisation, where the Bayesian optimisation learns to overfit to the seed. This was not seen to make a difference in practice, we believe, due to the many sources of randomness stemming from different parameters in each training run. The optimisation was stopped once all the parameters were seen to converge. For the more basic algorithms (DQN and A2C) where there are fewer hyperparameters, this was shorter.

We present the top results of the hyperparameter search in the following tables. 

\begin{table}[H]
\centering
\begin{tabular}{|p{4cm}|l|l|p{5cm}|}
\hline
\textbf{Parameter \newline (variable name)} & \textbf{DQN} & \textbf{DuelDDQN}& \textbf{Description} \\ \hline
Learning Rate (lr) & 0.002357 & 0.004133 & Learning rate of the Network's Adam Optimiser.\\ \hline
Exploration decay ($\rho$) &0.7052 & 0.5307 & Controls the speed at which the policy converges to greedy.\\ \hline
Batch Size & 256 & 128 & Batch of experience given to the network at once.\\ \hline
Buffer Size & 32 768 & 32 768 & Maximum number of experiences stored.\\ \hline
Periodic update ($\frac{1}{\tau}$) & $\frac{1}{0.0877}\to12$ & $\frac{1}{0.01856}\to 54 $ & Frequency of target network \newline parameter copying. \\ \hline
Learning rate decay \newline (decay number) & 6 & 10 & Number of times the learning rate is reduced by half. \\ \hline
Buffer Prioritisation ($\alpha$) & - & 0.213 & Controls the probability of \newline picking high-priority experience. \\ \hline
Importance Sampling\newline exponent ($\beta$) & - & 0.7389 & Controls the value of Importance sampling weights.\\ \hline \hline
Metric (mean rewards) & 321.471 & 330.394& Bayesian optimisation target.\\ \hline
\end{tabular}
\caption{Hyperparameter table for the DQN and DuelDDQN algorithms, top result of 32 and 102 runs of Bayesian optimisation respectively.}
\label{table:dqn}
\end{table}

\begin{table}[H]
\centering
\begin{tabular}{|p{4cm}|l|l|p{6cm}|}
\hline
\textbf{Parameter \newline (variable name)} & \textbf{A2C} & \textbf{PPO} &\textbf{Description}\\ \hline
Actor Learning Rate \newline (lr actor) & 0.0002052 & 0.0003633 & Learning rate of the Actor's Adam\newline Optimiser. \\ \hline
Critic Learning Rate\newline (lr critic) & 0.002627 & 0.004864  & Learning rate of the Critic's Adam\newline Optimiser.\\ \hline
Entropy regularisation\newline coefficient ($\epsilon$) & 0.001672 & 0.0001411  & Coefficient to the entropy term $\mathcal{H}(\pi)$.\\ \hline
Batch Size & - & 256 & Batch of experience given to the network at once.\\ \hline
Rollout length\newline (buffer size) & 32 & 2048 & Length of experience accumulated for updating the networks.\\ \hline
Learning rate decay\newline (decay number) & 4 & 200 & Number of times the learning rate is\newline reduced by half. \\ \hline
$\lambda$-return parameter ($\lambda$) & - & 0.8845 & Balances short and long-term reward. \\ \hline
Policy Clipping parameter (clip) & - & 0.2762 & Controls how far away the new policy can go from the previous policy.\\ \hline \hline 
Metric (mean rewards) & 74.339 & 128.376 & Bayesian optimisation target.\\ \hline
\end{tabular}
\caption{Hyperparameter table for the A2C and PPO algorithms, top result of 67 and 229 runs of Bayesian optimisation respectively.}
\label{table:ac}
\end{table}

We will comment on the numerical values of the hyperparameters in Chapter \ref{chapter:experiments}.

\newpage
\section{The Environment: the AYS model}\label{section:AYS}

We use the AYS model \cite{Kittel2017FromManagement} as the environment for our experiments. This is a low-complexity World-Earth system model. Here, we will summarise the important aspects of the model, but for details, please refer to the original paper. 

\subsection{Observables}
The model has three observed variables: 
\begin{itemize}
    \item Excess atmospheric carbon, $A$ in GtC\footnote{GigaTon of Carbon}
    \item Economic output, $Y$ in \$/yr
    \item Renewable knowledge stock, $S$ in GJ\footnote{GigaJoules}
\end{itemize}
This last variable is less tangible than the previous others, it represents how much knowledge humans have about renewable energy, and is in GJ to represent that it is proportional to renewable energy production. The dynamics of the model are simplified compared to the true interactions of humans and the climate. This low dimensional environment enables us to test our framework's limits as well as plot trajectories in phase space for better interpretability. We hope that future work in the field of World-Earth systems could bring new data-driven, higher dimensional environments that would model real-world interactions more accurately.

\subsection{Equations}
The system is governed by three differential equations, one for each observed variable:
\begin{align}
    \frac{dA}{dt} &= E- A/\tau_A,\\
    \frac{dY}{dt} &= \beta Y - \theta AY,\\
    \frac{dS}{dt} &= R - S/\tau_S.
\end{align}
With $R$ and $E$ the energy extracted from renewables and fossil fuels respectively. These are defined with the energy demand $U$ in GJ/year, which is proportional to the economic output:
\begin{align}
    U=\frac{Y}{\epsilon},
\end{align}
where $\epsilon=147 \,\$/GJ$, the efficiency of energy. Energy is either from renewable sources or from fossil fuel sources: 
\begin{align}
    R=(1-\Gamma)U,\\ F=\Gamma U,\\ E = F/\phi.
\end{align}
Here, $\phi$ is the fossil fuel combustion efficiency in GJ/GtC. The share of fossil fuel energy $\Gamma$ is calculated with an inverse response to the renewable knowledge: 
\begin{align}
    \Gamma = \frac{1}{1+(S/\sigma)^\rho}.
\end{align}
With $\sigma$ being the break-even knowledge: when renewable and fossil fuel costs become equal, and $\rho$ is the renewable knowledge learning rate. As knowledge on renewables ($S$) increases, $\Gamma \to 0$ and the total energy share produced by renewables increases. If $S\to 0$, then $\Gamma \to 1$ and more energy is produced from fossil fuels. The full table of parameters and their description is in Appendix \ref{Tables}.

To give a qualitative view on the equations:
\begin{itemize}
    \item Atmospheric carbon $A$ increases proportionally to fossil fuel energy production, but is reduced by natural $CO_2$ decay out of the atmosphere.
    \item Economic output $Y$ is exponential in growth, but is reduced by climate instability proportional to atmospheric carbon and the economic output.
    \item Renewable knowledge stock $S$ proportionally increases with the amount of renewable energy produced, but renewable knowledge deteriorates over time.
\end{itemize}
The interactions in the AYS are summarised in the schematic below:
\begin{figure}[H]
    \centering
    \includegraphics[scale=0.80]{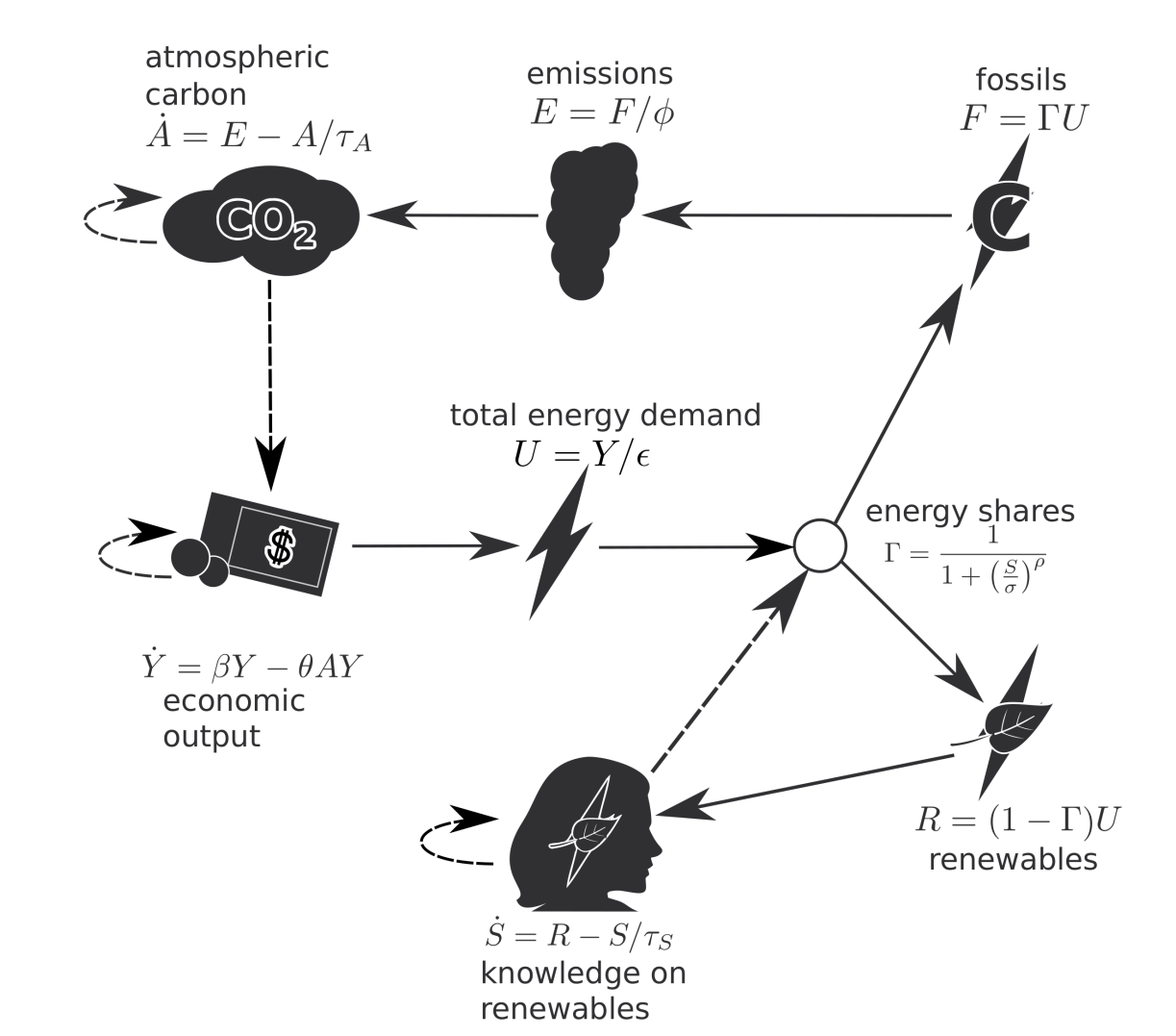}
    \caption{Schematic from Kittel et al. \cite{Kittel2017FromManagement} of the interactions in the AYS model. The $\dot{X}$ notation is equivalent to $\frac{dX}{dt}$. Line arrows are positive interactions, and dotted arrows are negative interactions.}
    \label{fig:aysmodel}
\end{figure}

\subsection{States}
The initial state is:
\begin{align}
s_{t=0} = 
    \begin{pmatrix}
    240\, GtC\\
    7\times 10^{13}\,\$/yr\\
    5\times10^{11}\, GJ\\
    \end{pmatrix},
\end{align}
which aims to represent the current state of the Earth in this model.
There are two attractors in this model: 
\begin{align}
s_b = 
    \begin{pmatrix}
    \beta/\theta\\
    \frac{\phi\beta\epsilon}{\theta \tau_A}\\
    0\\
    \end{pmatrix}
    =
    \begin{pmatrix}
    350\, GtC\\
    4.84\times10^{13}\,\$/yr\\
    0\, GJ
    \end{pmatrix}.
\end{align}
This is denoted as the \textit{Black fixed point}: roughly half of the current economic production, and this economic production is stagnant. Furthermore, in the Black fixed point, there is an excess of 350 GtC in the atmosphere with no renewable energy production: this can be seen as the scenario we wish to avoid. 
The other point we are interested in, is located at the boundaries of the state space: 
\begin{align}
    s_g = 
    \begin{pmatrix}
    0\\
    +\infty\\
    +\infty
    \end{pmatrix}.
\end{align}
Where economic growth and renewable energy knowledge grow forever, we label this the \textit{Green fixed point}, it is the ideal scenario. The dynamics of this environment do not allow for more fixed points. Therefore, any point in the space will be naturally drawn to one of these fixed points.

Just as in Kittel et al., we bound the state variables $A$, $Y$ and $S$ between 0 and 1. This avoids many numerical issues that arise from dealing with large, unbounded numbers. The scheme used to normalise the state space is the following:
\begin{align}
    \Bar{s_t} = \frac{s_{t}}{s_t + s_{t=0}}.
\end{align}
This leads to the initial state being $(0.5, 0.5, 0.5)^T$ and the Green fixed point to be $s_g=(0, 1, 1)^T$.

\subsection{Planetary Boundaries}
The model includes planetary and social boundaries, quantitative limits that should not be crossed, lest of disastrous and irreversible consequences for the biosphere and humans \cite{Rockstrom2009AHumanity, Steffen2015PlanetaryPlanet}. The planetary boundary for Atmospheric carbon is set at $A_{PB}=345\,GtC$. A minimum economic output (Social Foundation boundary) is set at $Y_{SF}=4\times10^{13}\,\$/yr$, this number is the total world Gross Domestic Product (GDP) from the year 2000. Kittel et al. admit that this is a rough estimate, as GDP is a flawed measurement of wealth. We show the phase space with the boundaries below in Figure \ref{AYS_basins}.
\begin{figure}[H]
    \centering
    \includegraphics[scale=0.6]{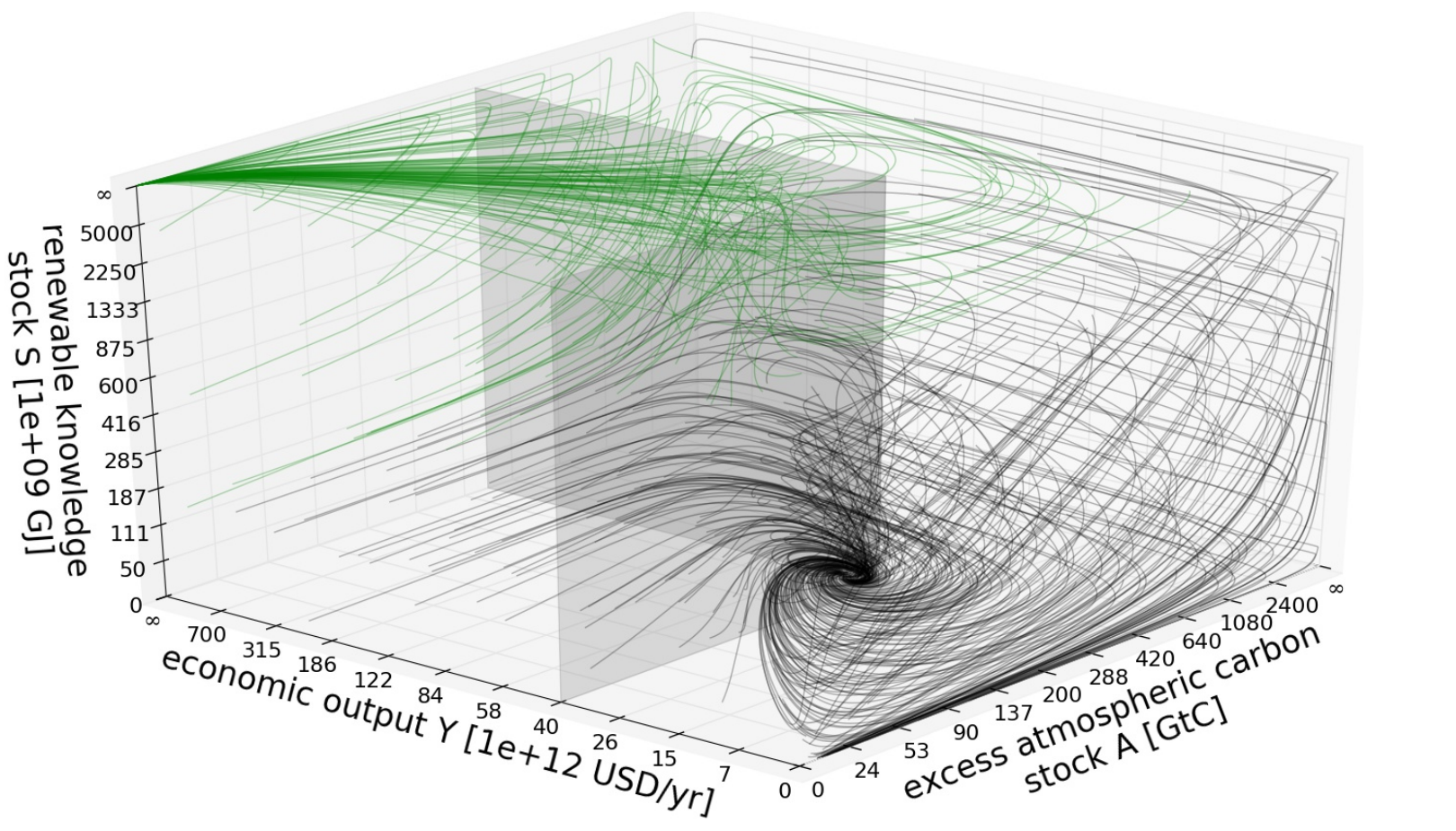}
    \caption{Phase Space of the AYS model with boundaries from Kittel et al. \cite{Kittel2017FromManagement}. Different initial conditions with their corresponding trajectories are plotted with default dynamics. They are colour-coded according to the fixed point they reach, these can be seen as the default flows of the model. The Carbon and Economic boundaries are shown as planes intersecting the space.}
    \label{AYS_basins}
\end{figure}
Observing Figure \ref{AYS_basins}, we can see the Black fixed point at the bottom, as well as the Green fixed point at the top left. The default flows of the model give intuition of the where each initial state will tend. The planetary boundaries are also represented as grey planes intersecting the phase space.

\subsection{Actions}
In Kittel et al., four actions are described:
\begin{itemize}
    \item \textit{Default}, the default parameters are used, the environment evolves as described above.
    \item \textit{DeGrowth (DG)}, the $\beta$ parameter, that quantifies the economic growth, is halved from 3\%/yr to 1.5\%/yr.
    \item \textit{Energy Transition (ET)}, the $\sigma$ parameter is reduced by $1-1/\sqrt{2} = 31.3\%$, which is equivalent to halving the break even cost (renewable energy to fossil fuel energy cost ratio).
    \item \textit{DG+ET}, a combination of the two above options.
\end{itemize}
By using these actions, the parameters of the equations above are changed, and thus the flows are changed.

\subsection{Rewards}
No reward function is described in the original paper, leading us to use the same one as in Strnad et al. \cite{Strnad2019DeepStrategies}, which we describe here. The goal for the agent is to reach the Green fixed point, $s_g$ which, is far away from the planetary boundaries in phase space. We therefore re-frame the problem as a staying away from the planetary boundaries set by the environment. We use a distance between the current agent state and the planetary boundary state, which we label:
\begin{align}
    s_{PB} &= \begin{pmatrix}
    945\,GtC\\
    4\times10^{13}\,\$/yr\\
    0\, GJ
    \end{pmatrix}\\ &\Leftrightarrow
    \begin{pmatrix}
    0.59\\
    0.37\\
    0
    \end{pmatrix} \textrm{ in normalised state space.}
\end{align} 
We get the following reward function:
\begin{align*}
    R_t =  ||s_t-s_{PB}||^2.
\end{align*}
This makes intuitive sense from a short-term perspective: we would like to maximise economic output and renewable energy knowledge while staying away from the atmospheric Carbon Planetary boundary. We call this the Planetary Boundary (PB) reward just like in Strnad et al. We use this same reward function for easy comparison with Strnad et al. in our experiments, but we will extend it and change it in various ways to test the limits of our agents.

\subsection{Episode Description}
The AYS model is a deterministic environment, to make sure the agents learn, we initialise each new episode at a random state defined by:
\begin{align}
    s_{t=0}=
    \begin{pmatrix}
    0.5 + \mathcal{U}(-0.05, 0.05)\\
    0.5 + \mathcal{U}(-0.05, 0.05)\\
    0.5\\
    \end{pmatrix}.
\end{align}
Where $\mathcal{U}$ is the uniform distribution. The reason the last variable has no noise added follows from Strnad et al., if noise is added to this variable, it dramatically reduces the ability for the agent to learn. 

The agent can select one of the four actions at each integration time step which, again from Strnad et al., we take as being equal to one, this corresponds to one action every year. Then the environment uses an Ordinary Differential Equation (ODE) numerical solver to calculate the state for the next step given the action-specific parameters. 

Following from Strnad et al., the episode terminates when the agent either:
\begin{itemize}
    \item crosses a Planetary Boundary
    \item reaches the vicinity of the Black Final Point
    \item reaches the vicinity of the Green Final Point
\end{itemize}
In the last two cases, the future discounted rewards are approximated and given to the agent. This is to compensate for the fact that an agent might receive a smaller reward if it reaches a final point too quickly due to the nature of the distance-based reward. One that reaches a final point more slowly would have more time to collect rewards.

%% file: chapters/4.experiments.tex
\section{Introduction to Experiments}
\subsection{Experimental Setup}

In this chapter, we will present the experiments and their results. We conduct five different experiments, altering only one aspect of the agent-environment interface for each experiment to draw meaningful conclusions.
\begin{itemize}
    \item First, we use the same experimental set up as in Strnad et al. \cite{Strnad2019DeepStrategies} for comparison of results, we will call this the \textit{Planetary Boundary} or \textit{PB} reward experiment as the reward function depends on the distance to the planetary boundaries. Does the tuning of hyperparameters improve performance compared to previous work? How do on-policy algorithms compare to off-policy ones in this environment? 
    \item Then, we modify the reward function so that a cost is added to using each action that implies a policy change. We want to minimise the number of interactions that the agent has with the environment. This is the \textit{Policy Cost} experiment. How is the learning of the agents affected when actions have intrinsically different values?
    \item We radically change the reward function to a much more simple scheme, with sparser rewards, naming it the \textit{Simple} reward, and to assert if the Planetary Boundary reward function is flawed. How does the learning of the agents change with sparse rewards?
    \item Next, to improve interpretability, we inject various amounts of noise in the parameters of the environment to observe how the agents react to different environment parameters at every episode. How does a noisy environment affect the learning of the agents?
    \item Lastly, we show that the current environment is only partially observable. The state at each time step does not follow the Markov property. We modify the environment such that we have \textit{Markov States}. How does the learning of the agents change when the environment becomes fully observable?
\end{itemize}
Additionally, we analyse the data we have collected to understand the agents better. Then, we use the hyperparameter tuning data to analyse the agents' sensitivity to hyperparameters.

In this Chapter, we will often refer to the DQN agent and the Double-DQN with Duelling architecture and Prioritised Importance Sampling agent as the \textit{DQN agents}. We will refer to A2C and PPO collectively as the \textit{Actor Critics}. This helps to compare trends without excessive notation. 

\subsection{Evaluation}
We compare our agents to each other with their moving average rewards. The aim is a high moving average reward with low variance. We chose to use the moving average reward across the last 50 episodes. This enables us to compare the performance of our agents with the results of Strnad et al., who use the same metric.
For the experiments, we modify the reward function in 2 of the 5 experiments. As the reward is our metric, changing it brings significant challenges for evaluation of our agents \textbf{across} different experiments. To accurately evaluate across different experiments, we have to find a different metric than the reward. We chose to use the \textit{Success Rate}. We define this metric as the number of times the agent reaches the Green fixed point during training divided by the number of episodes. This metric is independent of the type of the reward, and thus we can compare the agents across different experiments. We present the table summarising the success rates in different experiments in Section \ref{secion:summary}.

Another way to evaluate our model is by qualitatively analysing the trajectories that the agents take in the phase space. To this end, we present the phase space plots of the agents' trajectories in different experiments and describe them.

\subsection{Technical Details}
We use the four agents in 5 different experiments. For each experiment, we present the moving average episode reward of the last 50 episodes for each agent. The moving average for each agent is the mean over three seeds. The lightly shaded areas are half of the standard deviation of the three runs. To compute the average over the three seeds, the longer runs were cropped down to the shortest run for a given algorithm. The random seed controls the initialisation of neural network weights, the action sampling, the initial states of each episode and the sampling and shuffling of experience. For all the experiments, the agents were trained for 500000 frames with the parameters from the table in Chapter \ref{chapter:methods}. We use a frame limit to train the agents, as the training times can be inconsistent with an episode limit. This is because the agents who learn to control the environment fast will have longer episodes with more frames. By enforcing a frame limit rather than an episode limit, we ensure that training times are consistent across agents and seeds.
\newpage
\section{Comparison with Previous Benchmark}\label{section:PB}

\subsubsection{Results}
\begin{figure}[H]
    \centering
    \includegraphics[scale=0.65]{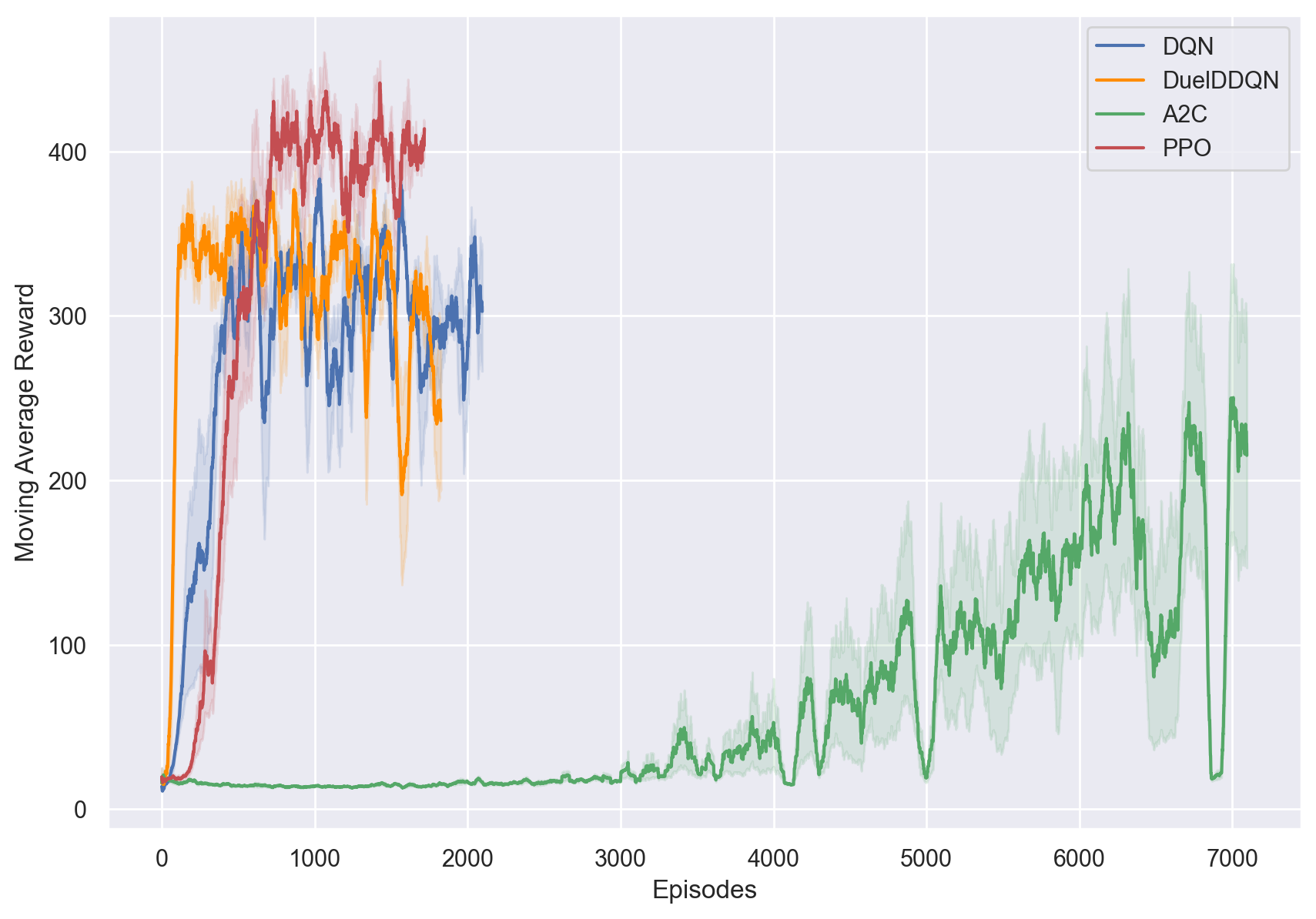}
    \caption{Moving average rewards of the four agents in the AYS environment with the \textit{Planetary Boundary} reward over 500 000 frames. The plot shows the mean and standard deviation over three different seeds. }
    \label{fig:default runs}
\end{figure}

Looking at the plot:
\begin{itemize}
    \item DQN learns fast and oscillates at an average reward of around 300, while staying relatively stable/low variance.
    \item DuelDDQN learns extremely fast but is unstable, decreasing from an average reward of 350 to 300 over time.
    \item A2C learns extremely slowly and is very high variance and unstable, achieving average rewards of 200 or so.
    \item PPO learns slowly, but then reaches average rewards of around 400 and is very stable.
\end{itemize}
We see that the best performing agent in terms of the best rewards is PPO, followed by DuelDDQN, then DQN and lastly A2C. This is in line with what we would expect, the agents with additional upgrades; PPO and DuelDDQN outperform their simpler counterparts; A2C and DQN. What is surprising is how slow A2C learns compared to the other agents. This is because A2C is not nearly as data efficient as the other agents, which all have some way of replaying experiences they have previously sampled. 

\subsubsection{Analysis}
For context, the top reward that agents can receive in this environment is 750 and the minimum is 15. Reaching the Green fixed point can yield a reward from 400 to 750, reaching the Black fixed point yields a reward of around 100. The lowest rewards (15-80) are achieved when crossing planetary boundaries. So, a mean reward of 300 can be interpreted as an agent getting to the Green fixed point once every 2 episodes. Picking actions uniformly at random yields a mean reward of 17.

The previous best benchmark set by Strnad et al. \cite{Strnad2019DeepStrategies} was a mean reward of around 250,
after 5000 episodes of training. The agent that set this benchmark was almost identical to our DuelDDQN agent, with a few differences in implementation. We can see that three of our agents comfortably outperform this both in average rewards and training speed. PPO, our best performing agent, achieves a mean reward of 400 after 800 episodes of training. The second-best agent, DuelDDQN, obtains mean rewards of 350 after around 200 episodes of learning. Our rigorous parameter tuning was effective at obtaining better rewards. 

\subsubsection{Trajectories}
Since our state space is three-dimensional, we can plot the trajectories of the agents in phase space. We colour-code the action at each time step. This gives an intuitive understanding of the behaviours of different agents. The initialisation $s_{0}=(0.5, 0.5, 0.5)^T$ is labelled \textit{current state} with a red dot, as it is an estimate of the current state of the Earth in this system.
\begin{figure}[H]
    \centering
    \begin{subfigure}[b]{0.49\textwidth}
        \centering
        \includegraphics[width=\textwidth]{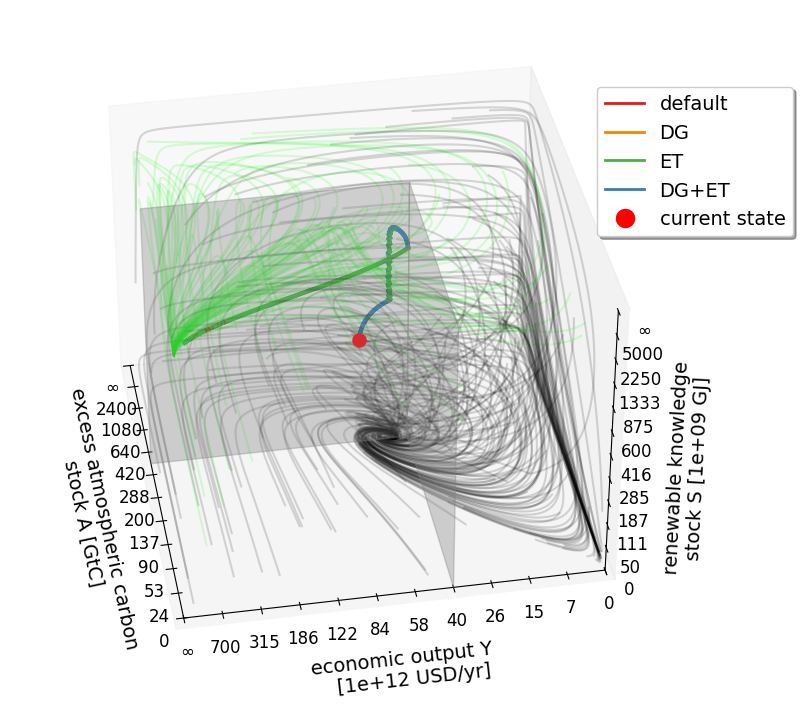}
        \caption{DQN}
        \label{fig:dqntraj}
    \end{subfigure}
    \hfill
    \begin{subfigure}[b]{0.49\textwidth}
    \centering
        \includegraphics[width=\textwidth]{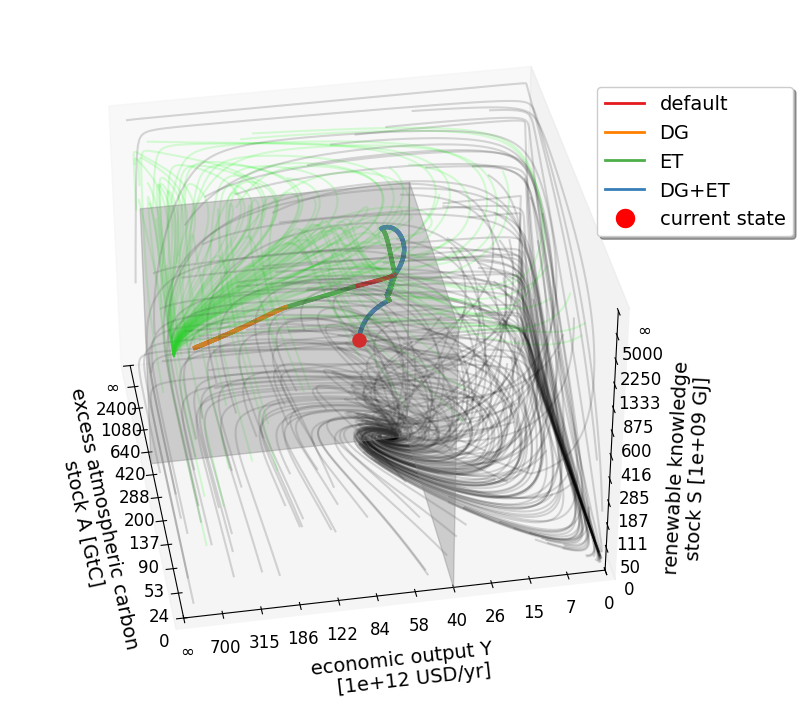}
        \caption{DuelDDQN}
        \label{fig:ddqntraj}
    \end{subfigure}
    \vskip\baselineskip
    \begin{subfigure}[b]{0.49\textwidth}
        \includegraphics[width=\textwidth]{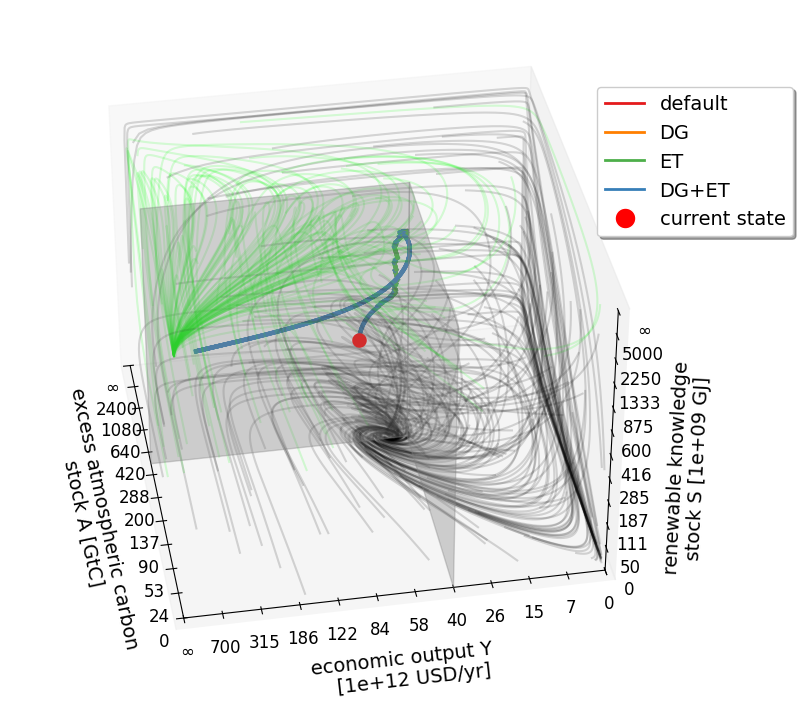}
        \caption{A2C}
        \label{fig:a2traj}
    \end{subfigure}
    \begin{subfigure}[b]{0.49\textwidth}
        \centering
        \includegraphics[width=\textwidth]{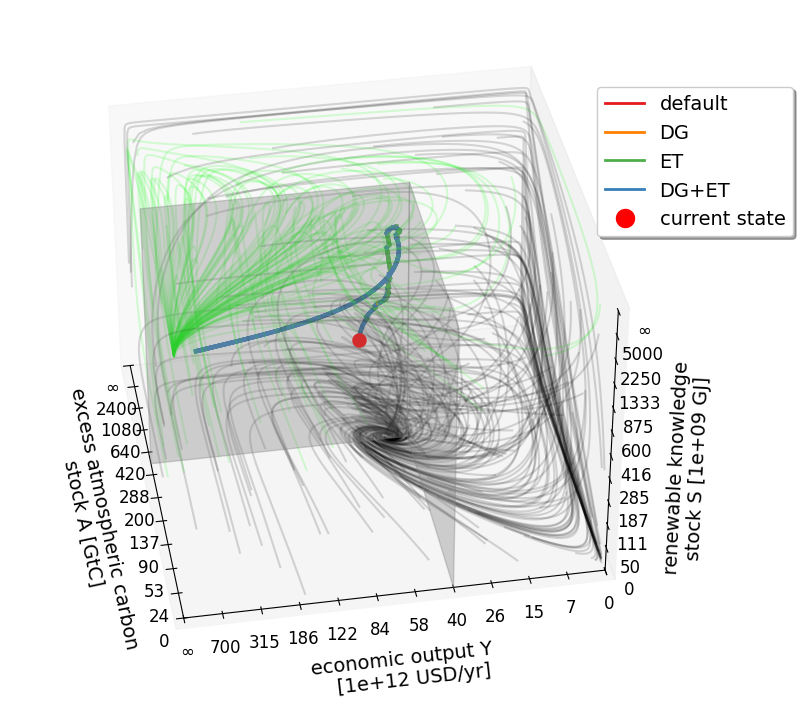}
        \caption{PPO}
        \label{fig:pptraj}
    \end{subfigure}
    \caption{Trajectory of the trained agents plotted in phase space from the $s_{0}=(0.5, 0.5, 0.5)^T$ initialisation (marked \textit{current state}). The trajectory is colour-coded with the actions of the agent at each time step.}
    \label{fig:traj}
\end{figure}

Looking at these plots, a pattern emerges. The agents pick the DG+ET action many times at the beginning. Then, once it gets close to the economic boundary, it switches off the DG option and instead just uses ET. Then, they use DG+ET again for a few more steps. Once it reaches certain renewable knowledge stock $S$, the agents start exhibiting some variance in the actions they choose. They all seem to prefer DG, or DG+ET. This has for effect that the agent converges slower to the optimal green state by slowing the economic growth. We believe this is because the agent wants to collect as much reward as possible from the environment by staying inside of it as long as possible. Strnad et al. \cite{Strnad2019DeepStrategies} addresses this by giving the agent discounted rewards at the end of the simulation for the remaining time steps, but by slowing the growth of $Y$, the agent strays further away from the atmospheric boundary (they do not follow the green flow lines) and therefore maximises its reward.

We can see that the Actor Critics learn similar trajectories. They stray as far as possible from the Atmospheric boundary. This is because the Actor Critics are \textit{on-policy}, they take into account their own stochastic action selection. Thus, they minimise the risk of crossing a boundary due to a stochastically selected action; they give themselves some leeway. This resonates well with the \textit{Cliff-Walk} example from Sutton \& Barto Chapter 6 \cite{Sutton2020ReinforcementIntroduction} that highlights the contrast between on-policy and off-policy algorithms.

The trajectory curves around the green contour lines by repeatedly using actions that change parameters of the system. This is most likely because the agent collects more reward the further it is from the point $(A_{PB}, Y_{SF}, 0)^T$, which is the lowest point at the intersection of the two planetary boundaries. Comparing this with the trajectory from Strnad et al. (Appendix \ref{graphs}), our agents appear to collect better rewards by using this curved trajectory. 
\newpage
\section{Behaviour with Different Rewards}

\subsection{Policy Cost}

To get a behaviour that is equivalent to what a human policymaker would want, the agent needs to take as few decisions as possible. For example, cutting economic growth by half, for multiple years in a row, is not realistic. Therefore, we adapt the previous reward function to include a cost to using actions that manage the parameters of the system. We set the cost of using each action to 50\% of the reward for that time step. If the double action DG+ET is used, the reward is reduced by a further 50\%, such that the reward received is 25\% of the original reward given. The reward for using no management, \textit{default} is then 100\% of its original value. This incentivises the agents to use as few management actions as possible. We summarise this below with $R_t^r$ the reward received by the agent and $R_t^{PB}$ the reward from the PB reward function in the previous experiment:
\begin{itemize}
    \item if $a_t=default ,\,R_t^r=R_t^{PB}$,
    \item if $a_t=ET  ,\,  R_t^r=0.5\times R_t^{PB}$,
    \item if $a_t=DG ,\,R_t^r=0.5\times R_t^{PB}$,
    \item if $a_t=DG+ET , \,R_t^r=0.25\times R_t^{PB}$.
\end{itemize}
These coefficients are arbitrarily chosen for the simplicity of the experiment, where we want to evaluate the agents' performance where we penalise taking more actions. Further experiments could be done with different coefficients, or consulting experts to derive more data-driven coefficients.

\subsubsection{Results}
\begin{figure}[H]
    \centering
    \includegraphics[scale=0.65]{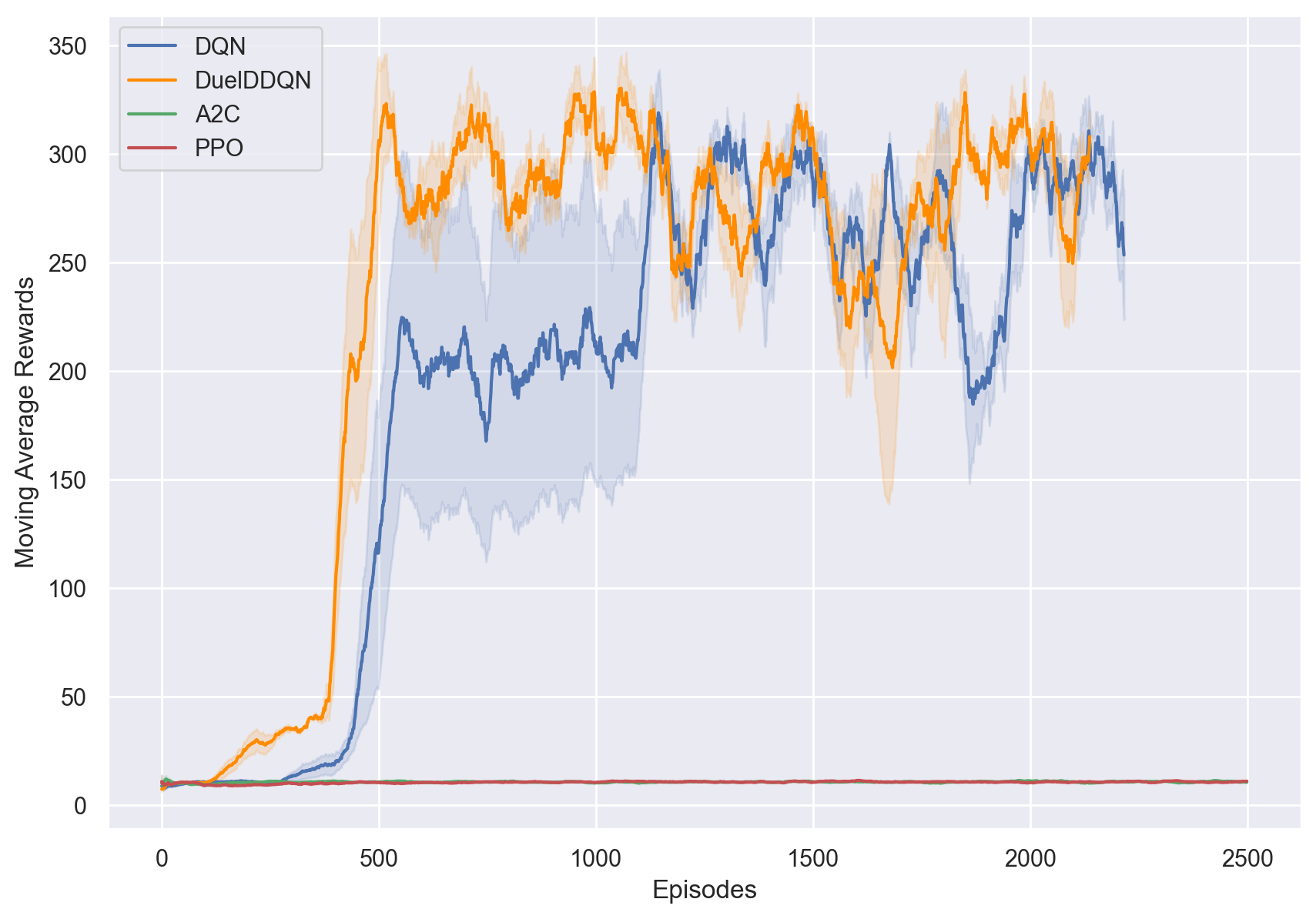}
    \caption{Moving average rewards of the four agents in the AYS environment, with the \textit{Policy Cost} reward over 500 000 frames. The plot shows the mean and standard deviation over
three different seeds. \textit{n.b:} The rewards for PPO and A2C are cropped from 10 000 episodes down to 2500 for visibility, for the full plots, please see Appendix \ref{graphs}.}
    \label{fig:cost_plot}
\end{figure}
\begin{itemize}
    \item For DQN agents, the rewards are around 50 less than for the PB reward. This is in line with the fact that the rewards available have been significantly reduced by penalising taking actions. DuelDDQN is reduced from a mean reward of around 350 in Figure \ref{fig:default runs} to 300 here, and DQN from 300 to around 250 here. The agents learn to maximise the reward with the changed reward function.
    \item The most obvious differences are in the learning of the Actor Critics; there is none of it. In 10000 episodes, they do not learn to maximise rewards as well as the DQN agents. 
\end{itemize}

\subsubsection{Trajectories}
\begin{figure}[H]
    \centering
    \begin{subfigure}[b]{0.49\textwidth}
        \centering
        \includegraphics[width=\textwidth]{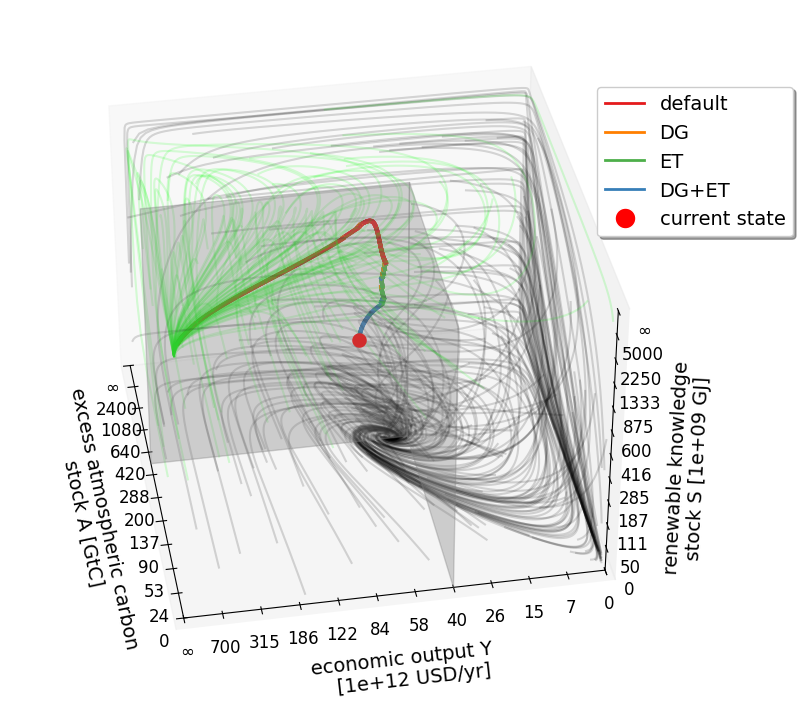}
        \caption{DQN}
        \label{fig:dqntrajcost}
    \end{subfigure}
    \hfill
    \begin{subfigure}[b]{0.49\textwidth}
    \centering
        \includegraphics[width=\textwidth]{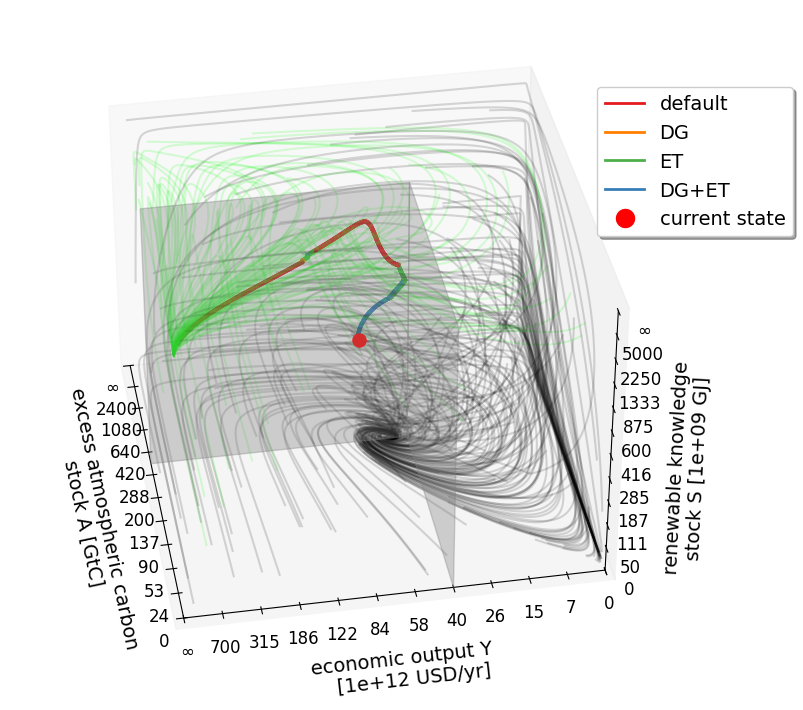}
        \caption{DuelDDQN}
        \label{fig:ddqntrajcost}
    \end{subfigure}
    \vskip\baselineskip
    \begin{subfigure}[b]{0.49\textwidth}
        \includegraphics[width=\textwidth]{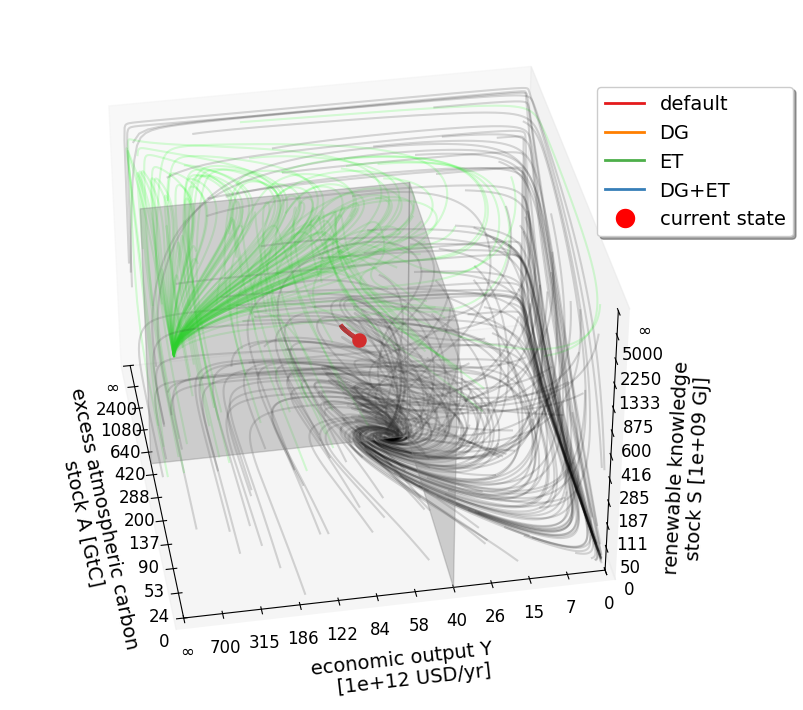}
        \caption{A2C}
        \label{fig:a2trajcost}
    \end{subfigure}
    \begin{subfigure}[b]{0.49\textwidth}
        \centering
        \includegraphics[width=\textwidth]{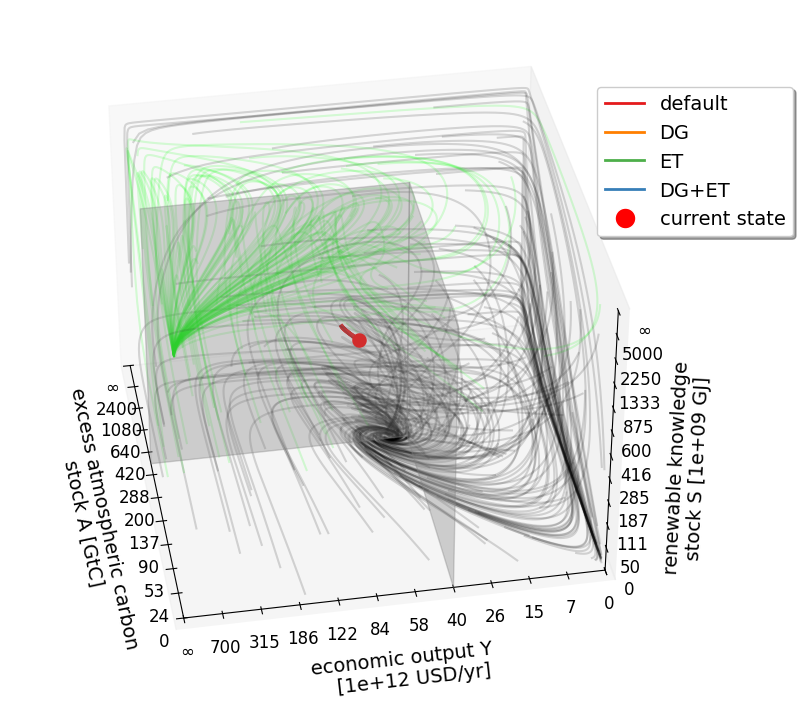}
        \caption{PPO}
        \label{fig:pptrajcost}
    \end{subfigure}
    \caption{Trajectory of agents trained with the \textit{Policy Cost} reward, plotted in phase space from the $s_{0}=(0.5, 0.5, 0.5)^T$ initialisation (labelled \textit{current state}). The trajectory is colour-coded with the actions of the agent at each time step.}
    \label{fig:trajcost}
\end{figure}

\begin{itemize}
    \item We see that, indeed, the DQN agents do what we expected them to: use the \textit{default} action as much as possible to maximise the rewards. They seem to use the double action DG+ET early on then switch to just ET, similarly to Figures \ref{fig:dqntraj}, \ref{fig:ddqntraj} with the regular PB reward. Then, they use the natural flows of the system to be carried to the Green fixed point without using any management actions. This shows that the agents in the previous experiment were using the PB reward to their advantage, as the reward function made no assumption about the relative value of each action. 
    \item The Actor Critics have learnt to only take the \textit{default} action to maximise their reward. We believe this to be due to the lack of exploration. We can theorise that because the entropy regularisation parameters are small ($\epsilon=0.001672$ and $0.0001411$ for A2C and PPO respectively), not enough exploration is conducted by the Actor Critics. They never find out that the short-term reward of not using management options is minuscule compared to the reward that can be obtained from staying in the environment longer.
\end{itemize}

A lot of exploration is required to learn an optimal trajectory, when there are intrinsically different rewards for each action. The on-policy agents struggle with this significantly. 

\subsection{Simple Reward}
From the previous two experiments, the reward function seems to be introducing a significant amount of bias in the behaviour of the agents. To verify this hypothesis, we create a much simpler reward function that only considers whether the agent reaches the Green fixed point and penalises crossing the boundaries:
\begin{align}
    r(s) = 
    \begin{cases}
    1 \quad \textrm{if} \quad s_t=s_g\\
    -1 \quad \textrm{if} \quad (A_t>A_{PB}) \vee (Y_t<Y_{SF})\\
    0 \quad \textrm{otherwise}.
    \end{cases}
\end{align}
We call this much simpler scheme the \textit{Simple} reward. This makes the environment harder to master for the agents, as the PB reward does not guide them. They only receive rewards at the end of the episode; the rewards are sparse. We expect the learning to be much more difficult for the agents, but that the trajectories obtained will be less dependent on the planetary boundaries. From the previous experiment, we can expect that the DQN agents perform better as they explore more effectively than the Actor Critics. 

\subsubsection{Results}
We show the learning curves for the agents with this different reward, but it should be noted that comparison with the other reward types is not particularly useful as they have different scales. We compare the agents with each other.
\begin{figure}[H]
    \centering
    \includegraphics[scale=0.65]{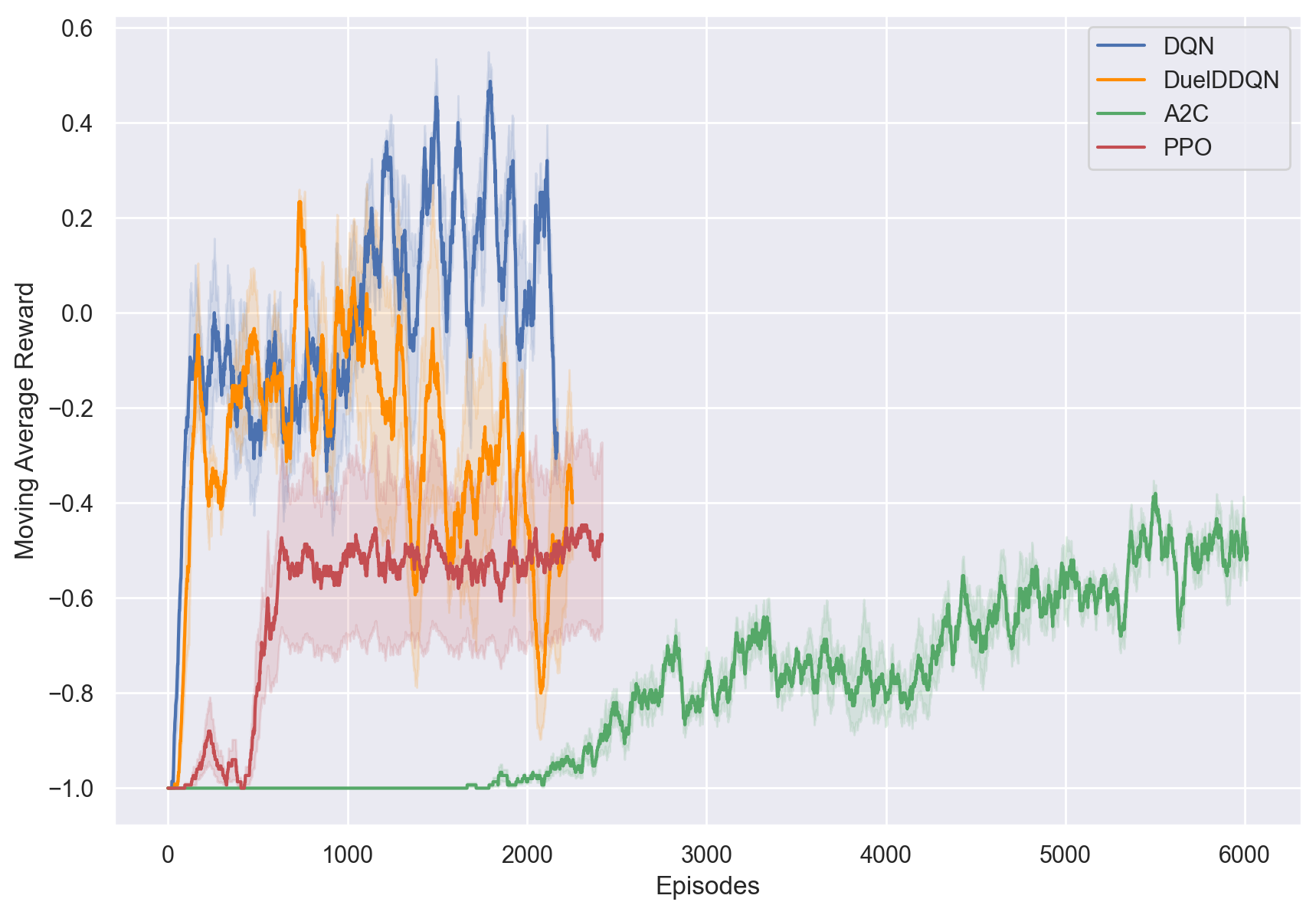}
    \caption{Moving average rewards of the four agents in the AYS environment, with the \textit{Simple} reward over 500 000 frames. The plot shows the mean and standard deviation over three different seeds.}
    \label{fig:simple_plot}
\end{figure}

We see that, surprisingly, DQN performs the best, learning an average reward of 0.2. DuelDDQN learns just as fast, but then its learning seems to collapse over time, down to an average reward of -0.5. 
PPO and A2C, perform worse than the DQN agents and learn quite slowly, particularly A2C. They both achieve mean rewards of -0.5 but seem much more stable than the DQN agents.

The collapse of the DuelDDQN's learning could be attributed to a bad tuning of hyperparameters, particularly the learning rate decay. This is sensible given how sensitive the agents are to hyperparameters, as we show later in Section \ref{section:hparam}. By radically changing the reward function, without re-tuning parameters, we have much worse performance. This is reinforced by the fact that DQN and A2C, the two agents with the fewest hyperparameters, perform better or just as well as their more complex counterparts. Re-tuning hyperparameters for this specific reward function was not further explored. 

\subsubsection{Trajectories}
\begin{figure}[H]
    \centering
    \begin{subfigure}[b]{0.49\textwidth}
        \centering
        \includegraphics[width=\textwidth]{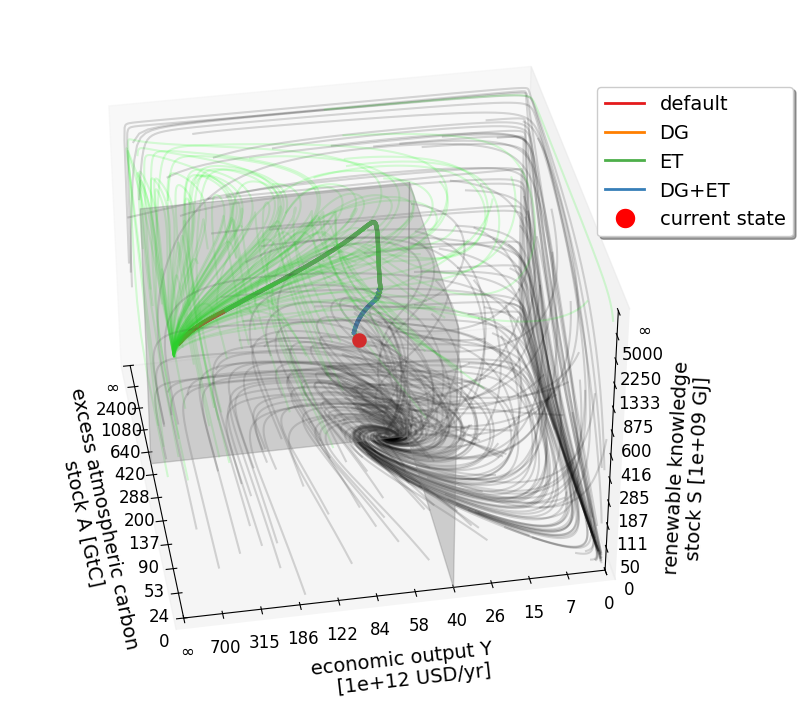}
        \caption{DQN}
        \label{fig:dqntrajsimple}
    \end{subfigure}
    \hfill
    \begin{subfigure}[b]{0.49\textwidth}
    \centering
        \includegraphics[width=\textwidth]{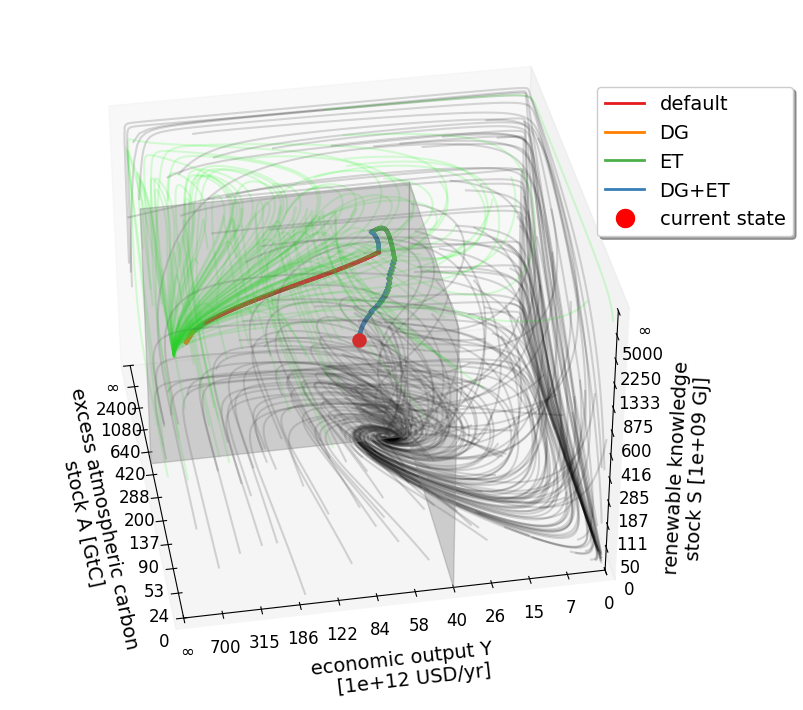}
        \caption{DuelDDQN}
        \label{fig:ddqntrajsimple}
    \end{subfigure}
    \vskip\baselineskip
    \begin{subfigure}[b]{0.49\textwidth}
        \includegraphics[width=\textwidth]{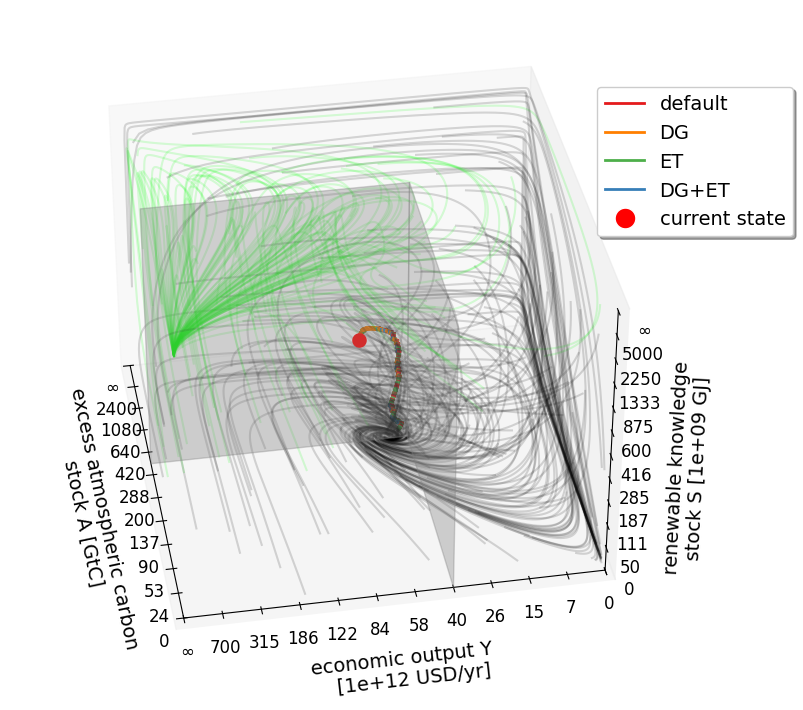}
        \caption{A2C}
        \label{fig:a2trajsimple}
    \end{subfigure}
    \begin{subfigure}[b]{0.49\textwidth}
        \centering
        \includegraphics[width=\textwidth]{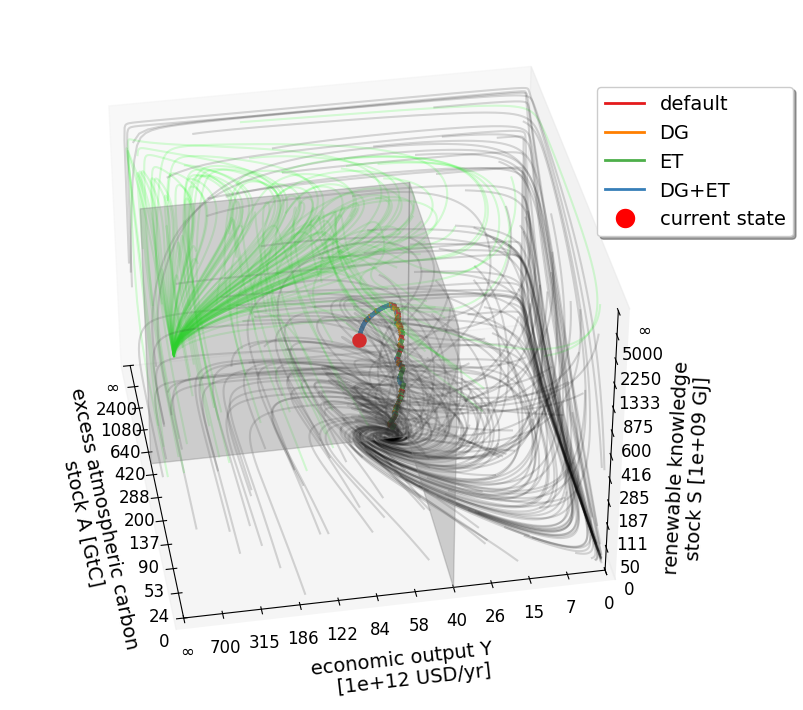}
        \caption{PPO}
        \label{fig:pptrajsimple}
    \end{subfigure}
    \caption{Trajectory of agents trained with the \textit{Simple} reward, plotted in phase space from the $s_{0}=(0.5, 0.5, 0.5)^T$ initialisation (labelled \textit{current state}). The trajectory is colour-coded with the actions of the agent at each time step.}
    \label{fig:trajsimple}
\end{figure}

\begin{itemize}
    \item For the DQN agents, they learn similar shapes of trajectories to the \textit{Policy Cost} reward from Figure \ref{fig:trajcost}. First they pick DG+ET, then they switch to just ET to reach a certain renewable knowledge stock $S$. After which, they get carried by the flows of the system towards the Green fixed point.
    \item The Actor Critics seem to pick the actions in a uniform manner, as can be seen from the variation of actions they take at each time step. PPO has slightly more structure as it picks the action DG+ET for the first few time steps, just as in the previous trajectories in Figures \ref{fig:traj} and \ref{fig:trajcost}. We believe this poor performance to be again due to lack of exploration. 
\end{itemize}

Given the similarity in trajectories (for the DQN agents) between the Policy Cost reward and the Simple reward, we can say that there is no significant bias by using the Policy Cost reward. The bias in the PB experiment (the curved trajectories) comes from the fact that the actions are treated as equal in the PB reward. 

\newpage
\section{Changing the Environment}
In this section, we modify the environment slightly without altering its core dynamics to improve interpretability.
\subsection{Environment with Noisy Parameters}
So far, given an initialisation, our environment is completely deterministic. The real world is rarely deterministic, it is often noisy. To make sure the agents are learning something meaningful about the environment, that is interpretable, we add a stochastic component to the environment. At each new episode, the parameters have a slight amount of noise added to them, we want to see how this affects the learning of the agent. This enables us to analyse how robust the agents are to different parameters so that they can be more applicable to the real world, where parameters are never perfectly known. 

\subsubsection{Implementation}
We use the agents in a modified version of the environment, where a small amount of Gaussian noise is added to each parameter of the environment on every episode. The parameters have the values given in Section \ref{section:AYS}. For each parameter $p$ in the set of parameters of $\mathcal{P}$, at each new episode:
\begin{align}
    p_{n} = p \times \varepsilon\qquad \textrm{where} \;\varepsilon\sim\mathcal{N}(1, \sigma^2_{n}) \qquad \forall p\in \mathcal{P}
\end{align}
Where $\sigma^2_n$ is a parameter that controls the noise we inject in the environment parameters. We use this scheme to introduce noise as the parameters span many orders of magnitude, so the noise injected has to scale with the parameters' magnitude. The sampled noise is independent for all the parameters, and has the correct scale for each parameter. The more we increase $\sigma^2_n$, the more noise is injected into the environment's parameters. We increase $\sigma^2_n$ for the agents to see how their learning is affected. We start $\sigma^2_n$ at $10^{-5}$ and increase is geometrically to 1 over 3000 episodes by multiplying the noise by 10 every 500 episodes. For computational stability reasons, the noise $\epsilon$ is clipped between 0.5 and 1.5 to prevent problems with the ODE solver. This is because, for certain parameter combinations, unbounded rapid exponential growth is possible. This also stops the environment from deviating too much from the original one.

We expect this to significantly negatively affect the learning of the agents over time, especially as the environment approaches noise variance of $\sigma^2_n=1$. Changing parameters of a dynamical system affects the location and size of attractors, which in turn changes how variables evolve. As a result, we include a random agent to compare with the learning agents. The random agent simply picks an action uniformly at random at every time step.

\subsubsection{Results}
\begin{figure}[H]
    \centering
    \includegraphics[scale=0.65]{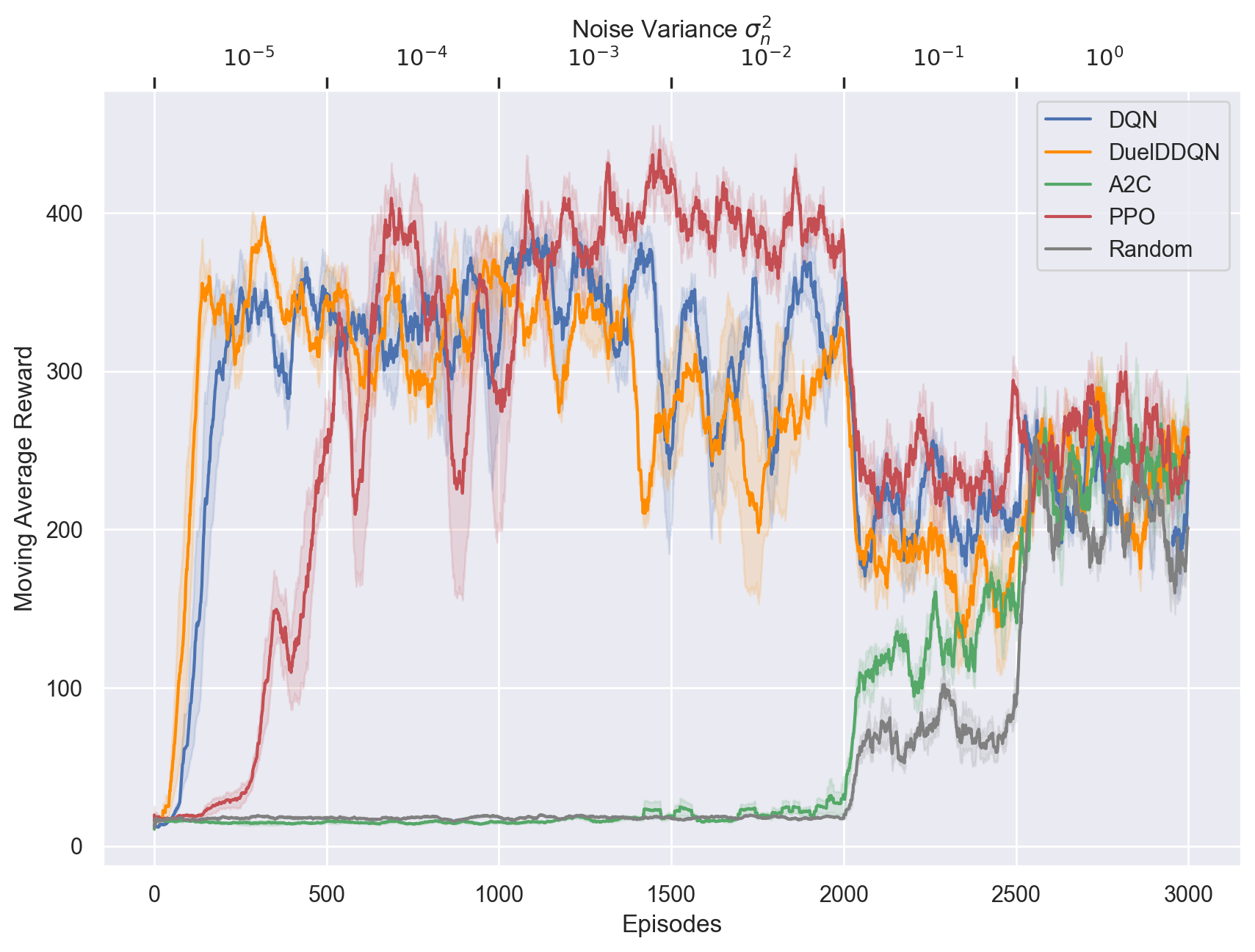}
    \caption{Moving average rewards of the five agents in the \textit{Noisy} AYS environment, with the Planetary Boundary reward over 3000 episodes. The plot shows the mean and standard deviation over three different seeds. The noise variance is initially at $10^{-5}$ and is multiplied by 10 every 500 episodes such that the noise variance is equal to 1 from episode 2500 onwards.}
    \label{fig:noisy plot}
\end{figure}
These results are very surprising considering how brittle the agents have been to hyperparameters in the previous experiments. We see here that the DQN agents have found a stable mean rewards of 350 for the first 1500 episodes, this is equivalent or better than in the noiseless environment in Figure \ref{fig:default runs}. We can also see that the DQN agents have much lower variance in their rewards. At episode 1500, when the noise variance increases from $\sigma^2_n=10^{-3}$ to $10^{-2}$, the DQN agents' mean reward collapses to around 300 for DQN and 250 for DuelDDQN. From this point on, the variance in rewards increases significantly. 

A2C performs ever so sightly better than random. PPO learns slower than in the PB experiment, taking around 1200 episodes to reach peak rewards. We believe this to be due to PPO being the most sensitive to hyperparameters out of all the agents: it took the most runs of Bayesian optimisations to tune the hyperparameters until convergence (229 runs). It then reaches a similar peak rewards of 400.

After episode 2000, all the agents converge to a mean reward of around 230, and seem relatively stable in this region. Most importantly, the Random agent also is in this range, but with a slightly lower average reward of 200. The noise variance at this point is $\sigma^2_n=1$, the agents perform barely better than random action selection once the parameters change too much.

The high stability we observe for the DQN agents when the noise variance is less than $10^{-2}$ (before 1500 episodes) is counterintuitive initially. We believe it is akin to a form of noise regularisation, a well-known Deep Learning regularisation method \cite{Goodfellow2016DeepLearning}. Adding noise to the environment enables the agent's network to find a more stable local optimum to converge to and not overfit to the environment. Verifying this hypothesis would require further investigation. We could not find any research that has explored such a phenomenon in Reinforcement Learning before. A similar phenomenon that was researched is adding noise directly to the agent's network parameters, which has been shown to improve performance, particularly in terms of exploration \cite{Plappert2017ParameterExploration}.

Adding noise seems to help the learning of the agents to some extent. If the noise added is too strong, the fundamentals of the environment are changed. Looking at the reward curves, there appears to be a sweet spot when $\sigma_n^2$ is less than $10^{-2}$. At $\sigma_n^2=10^{-2}$, the learning for the DQN agents decreases significantly.

\subsubsection{Testing the agents}
In the real world, we rarely know what the true parameters are, but they still have a fixed value. We want to see if we can apply the agents to the real world by testing them in an environment with fixed parameters sampled from the scheme above. We train a set of agents with the same setup as above, but keep the noise variance the same throughout training at $\sigma_n^2=10^{-3}$, the highest noise variance most of our agents seem to fare well with. We train the agents for 500 000 frames, the reward curves are in Appendix \ref{graphs}. We then test these agents in an environment with fixed parameters, sampled with $\sigma_n^2=10^{-3}$. We compare the agents trained here with the ones we trained in our first experiment (Section \ref{section:PB}), when noise was effectively $\sigma_n^2=0$. Are the agents trained in a noisy environment more adaptable than ones trained with a fixed environment?

\begin{table}[H]
\centering
\begin{tabular}{ccccc}
\toprule
Training Noise Variance & DQN & DuelDDQN & A2C & PPO \\ \midrule
$\sigma_n^2=0$ (PB experiment) & 358.5 $\pm$ 26.3 & \textbf{322.1 $\pm$ 62.0} &\textbf{423.8 $\pm$ 13.7} & \textbf{404.9 $\pm$ 38.2} \\
$\sigma_n^2=10^{-3}$ & \textbf{380.3 $\pm$ 36.1} & 153.9 $\pm$ 194.0 & 332.8 $\pm$ 25.16 & 271.4 $\pm$ 145.5 \\ \bottomrule
\end{tabular}
\caption{Mean rewards of running the agents in an environment with fixed parameters different from the original ones from Kittel et al. \cite{Kittel2017FromManagement}. We use agents trained with the fixed, original parameters as well as agents trained with noisy parameters (new parameters sampled at each episode). The mean and standard deviation are across 3 agents trained with different random seeds, but tested on the same environment.}
\end{table}
 
Looking at the table, the answer to our question is on average: no. Three out of four agents trained on fixed parameters do better than the ones trained with noisy parameters. Looking at the standard deviation in the last row, we see there is a lot of variation in the same agent trained with different seeds for the more complex agents DuelDDQN and PPO. One of the three seeds affected the learning of these agents significantly. This points to another hyperparameter tuning issue: the agents with fewer hyperparameters; A2C and DQN, are more adaptable than their more complex counterparts. A2C performs surprisingly well in testing, which we attribute to randomness and is not further investigated.

\subsection{Full Markov State}
Our model does not follow the Markov property. This is because the actions affect the parameters that control the differential equations of the system. Therefore, a state is not independent of the past, each state has a velocity that depends on the previous states and actions. This velocity is invisible to the agents: the previous experiments have been in a Partially Observed Markov Decision Process (POMDP).

\subsubsection{Implementation}
To obtain a full Markov state, information about the velocity must be included in the state. We use the same methods as in \textit{gym} environments\footnote{\href{https://github.com/openai/gym}{OpenAI Gym GitHub}}, we then have states of the form:
\begin{align}
s_t = 
    \begin{pmatrix}
    A_t\\
    Y_t\\
    S_t\\
    dA_t\\
    dY_t\\
    dS_t
    \end{pmatrix}
\end{align}
Where we calculate the differentials with the equations from Chapter \ref{chapter:methods} added to the velocity at the previous time step:
\begin{align}
    dA_t &= E_t- A_t/\tau_A + dA_{t-1},\\
    dY_t &= \beta(a_t) Y_t - \theta A_t Y_t + dY_{t-1},\\
    dS_t &= R_t - S_t/\tau_S + dS_{t-1}.
\end{align}
Our hypothesis is that this improves the learning of the agent. Intuitively, the information about the velocity should be very valuable to the agent to accurately evaluate actions. For example, an action that causes a high speed towards a planetary boundary would be much less valuable than one that causes the agent to move slowly towards the Green fixed point. 

\subsubsection{Results}
\begin{figure}[H]
    \centering
    \begin{subfigure}[b]{0.49\textwidth}
        \centering
        \includegraphics[width=\textwidth]{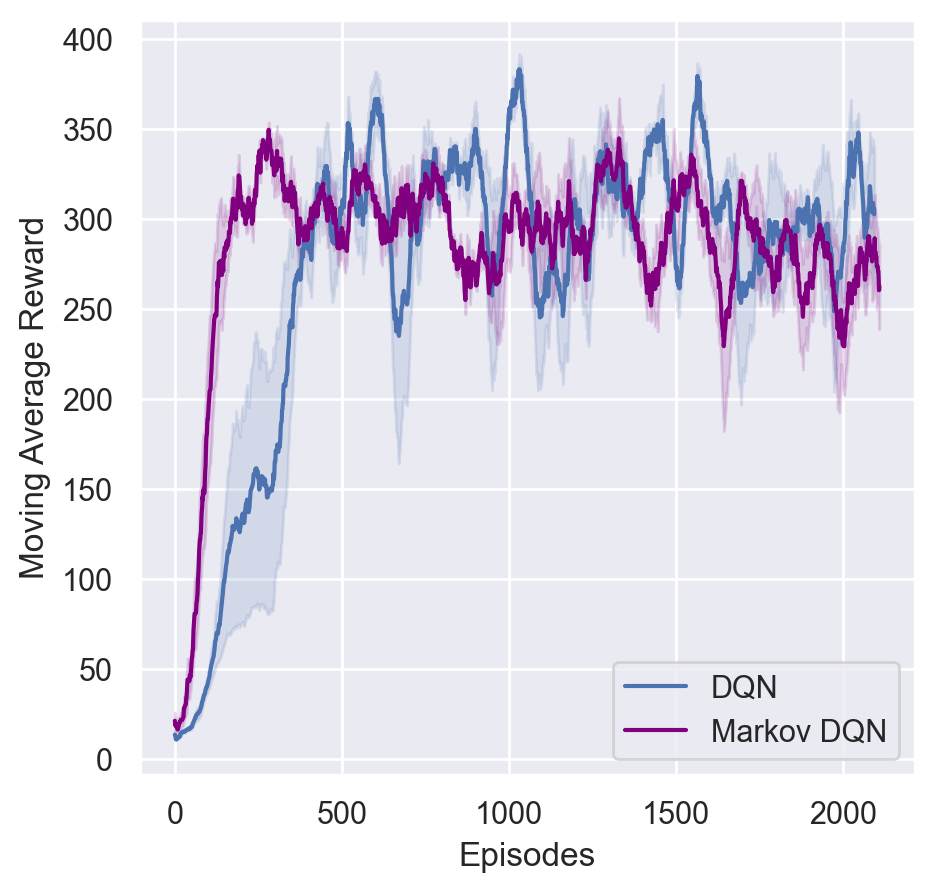}
        \caption{DQN}
        \label{fig:dqnmarkov}
    \end{subfigure}
    \hfill
    \begin{subfigure}[b]{0.49\textwidth}
    \centering
        \includegraphics[width=\textwidth]{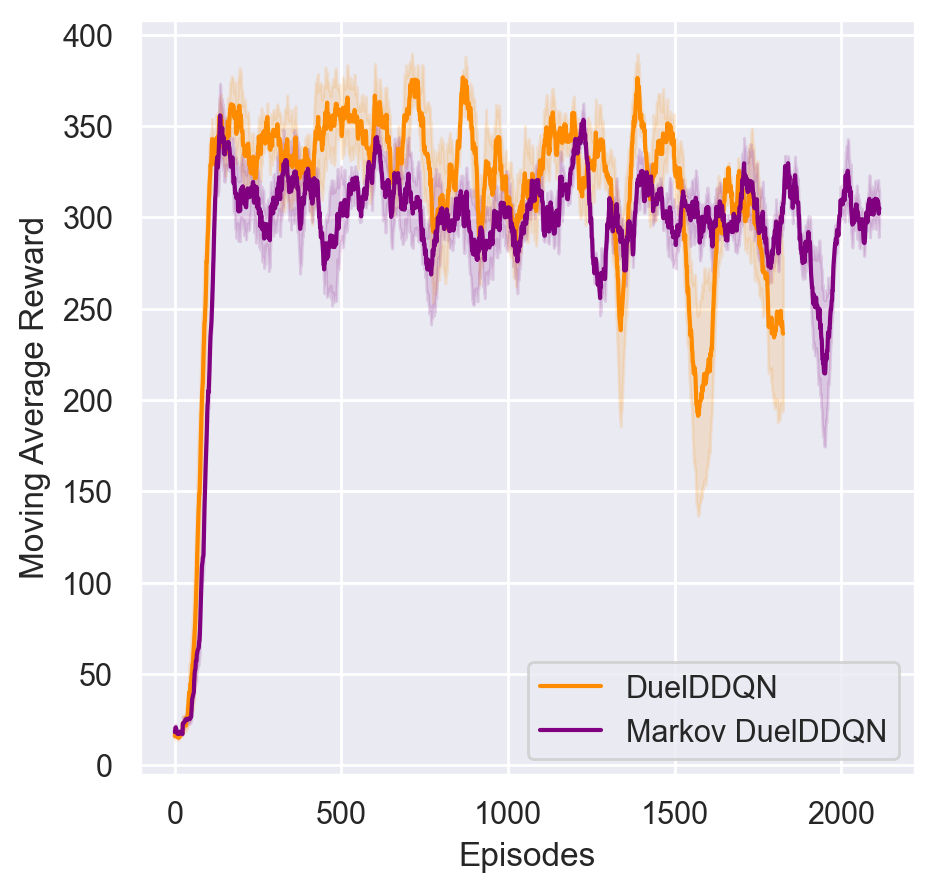}
        \caption{DuelDDQN}
        \label{fig:ddqnmarkov}
    \end{subfigure}
    \vskip\baselineskip
    \begin{subfigure}[b]{0.49\textwidth}
        \includegraphics[width=\textwidth]{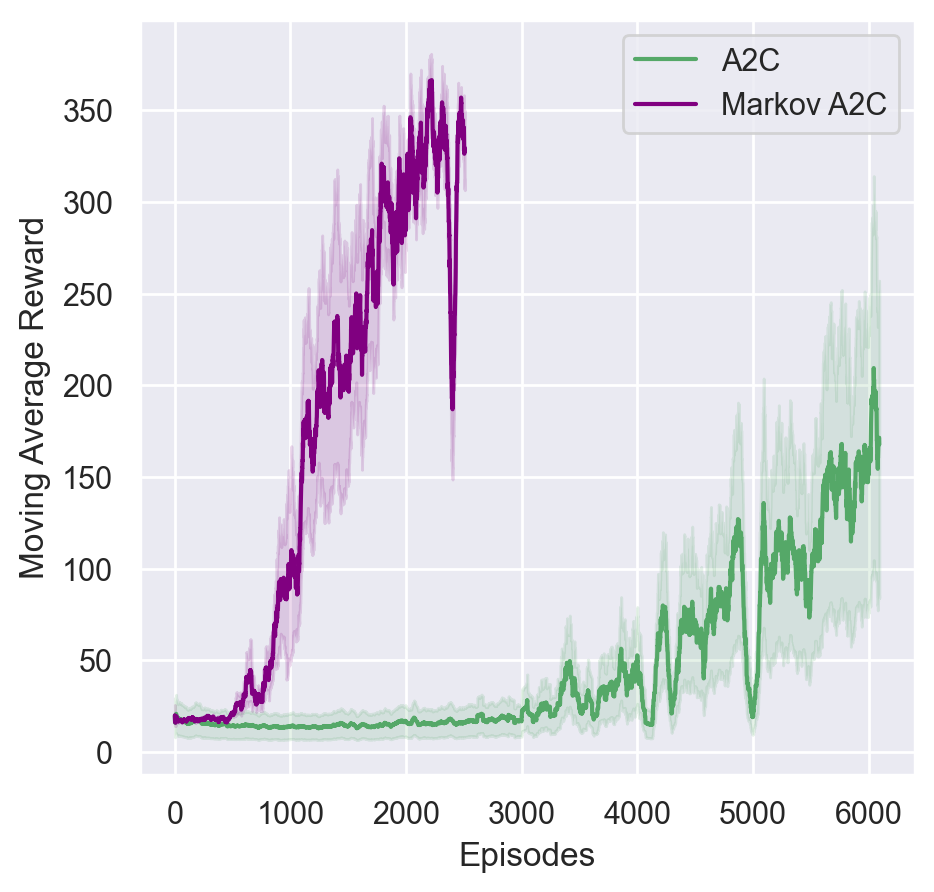}
        \caption{A2C}
        \label{fig:a2cmarkov}
    \end{subfigure}
    \begin{subfigure}[b]{0.49\textwidth}
        \centering
        \includegraphics[width=\textwidth]{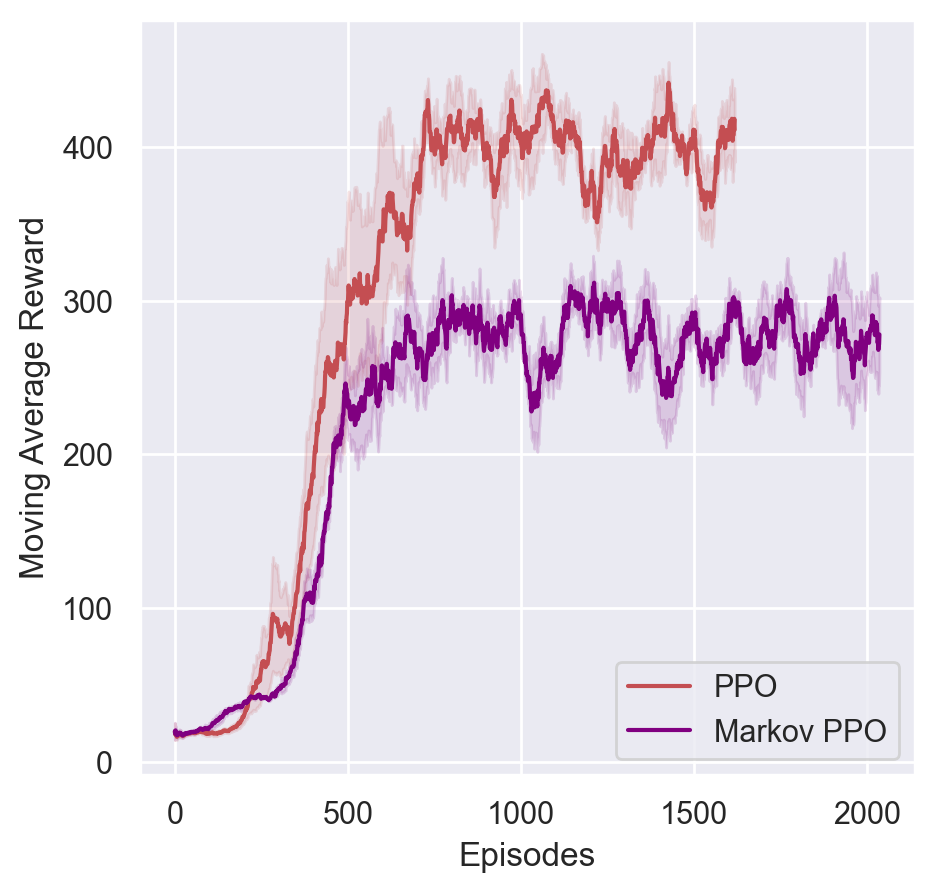}
        \caption{PPO}
        \label{fig:ppomarkov}
    \end{subfigure}
    \caption{Moving average rewards of the four agents with and without a Markov AYS environment, with the
Planetary Boundary reward over 500 000 frames. Plots show the mean and standard deviation over
three different seeds.}
    \label{fig:markov}
\end{figure}

The Markov agents are all more or less different than their non-Markov counterparts. We compare the agents to their Markov equivalent independently.
\begin{itemize}
    \item Markov DQN learns much faster than regular DQN, it then has lower variance in its mean reward. Despite this, Markov DQN achieves slightly lower rewards and its mean reward decays from around 300 down to 270 over time. DQN does not seem to decay as much.
    \item Markov DuelDDQN learns just as fast as DuelDDQN but then exhibits lower rewards at an average of 300 versus 350 for DuelDDQN. We believe that this is because the parameters used in this experiment are fine-tuned to the partial observability. Markov DuelDDQN exhibits lower variance. 
    \item For A2C, there is a very obvious difference in performance; there is much faster learning, higher peak rewards, and lower variance. We mentioned previously that sample efficiency might be a problem for A2C. Intuitively, with a full Markov state, the environment is fully observable, so the agent needs fewer samples to “understand” the environment.
    \item Here, for PPO, the mean rewards are significantly lower than the previous results, where in the PB experiment, PPO had a mean reward of 400, with Markov PPO, the agent obtains rewards closer to 280. This is again most likely due to hyperparameter sensitivity. 
\end{itemize}

It is worth noting that the DQN agents in the PB experiment have similar learning curves to the Markov DQN agents. The partial observability was not a significant hurdle for the DQN agents. For the two lower complexity agents: DQN and A2C, the full observability helps with the speed of learning, especially for A2C. For DuelDDQN and PPO, the full observability hurts the agent's learning. We conclude that full observability can help, but not if hyperparameters have been fine-tuned to partial observability.

\newpage
\section{Summary table} \label{secion:summary}
We summarise the results of the experiments in the following table. For accurate comparison, we use the \textit{Success Rate} metric, which measures the number of times an agent has reached the Green fixed point during training divided by the number of episodes.
\begin{table}[H]
\centering
\begin{tabular}{ccccc}
\toprule
\textbf{Experiment} & DQN & DuelDDQN & A2C & PPO \\ \midrule
\textbf{PB Reward} & 0.592 $\pm$ 0.026 & \textbf{0.645 $\pm$ 0.049} & 0.121 $\pm$  0.070 & 0.597 $\pm$ 0.103 \\ 
\textbf{Policy Cost Reward} & 0.493 $\pm$ 0.067 & \textbf{0.574 $\pm$ 0.020} & 0 $\pm$ 0 & 0 $\pm$ 0 \\ 
\textbf{Simple Reward} & \textbf{0.432 $\pm$ 0.034} & 0.237 $\pm$ 0.053 & 0 $\pm$ 0 & 0.017 $\pm$ 0.028 \\ 
\multicolumn{1}{c}{\begin{tabular}[c]{@{}c@{}}\textbf{Noisy Parameters}\\(increasing noise variance)\end{tabular}}   & 0.492 $\pm$ 0.035 & \textbf{0.501 $\pm$ 0.074} & 0.069 $\pm$ 0.008 & 0.432 $\pm$ 0.016 \\ 
\begin{tabular}[c]{@{}c@{}}\textbf{Noisy Parameters}\\ (fixed noise variance)\end{tabular} \newline  & 0.592 $\pm$ 0.063 & \textbf{0.596 $\pm$ 0.132}& 0.101 $\pm$ 0.028 & 0.231 $\pm$ 0.051 \\ 
\textbf{Markov State} & 0.639 $\pm$ 0.039 & \fcolorbox{black}{white}{\textbf{0.671 $\pm$ 0.021}} & 0.316 $\pm$ 0.062 & 0.386 $\pm$ 0.072 \\ 
\bottomrule
\end{tabular}
\caption{Summary of results of experiments for all agents. The agents were all trained with three seeds, we give the mean and the standard deviation of the metric. The metric is the success rate of the agents: how many times the agent has reached the Green fixed point divided by the number of episodes.}
\end{table}

We can see that the best performing models are the off-policy algorithms, DQN and DuelDDQN. DuelDDQN outperforms the other algorithms in all the experiments, except for the Simple reward, where DQN performs almost twice as well. The DQN agents also tend to have lower variance across the three seeds. It is worth noting that the success rate metric takes into account speed of training, since it is an average over the number of episodes. This explains why despite earning the highest rewards, the PPO agent does not have a higher success rate than DuelDDQN as it learns more slowly. 

\newpage
\section{Further Analysis} \label{further}

We present other trends and plots to compare between experiments.
\subsection{Initialisation}
\subsubsection{Planetary Boundaries Reward}
\begin{figure}[hbt!]
    \centering
    \begin{subfigure}[b]{0.49\textwidth}
        \centering
        \includegraphics[width=\textwidth]{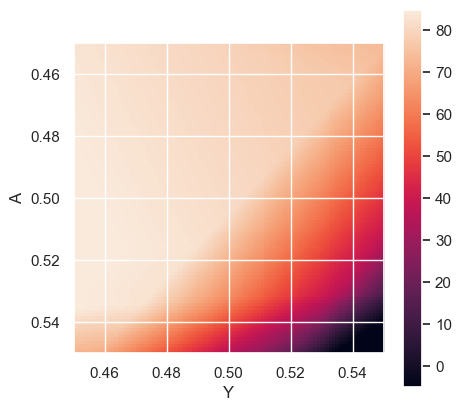}
        \caption{DQN}
        \label{fig:dqnvals}
    \end{subfigure}
    \hfill
    \begin{subfigure}[b]{0.49\textwidth}
    \centering
        \includegraphics[width=\textwidth]{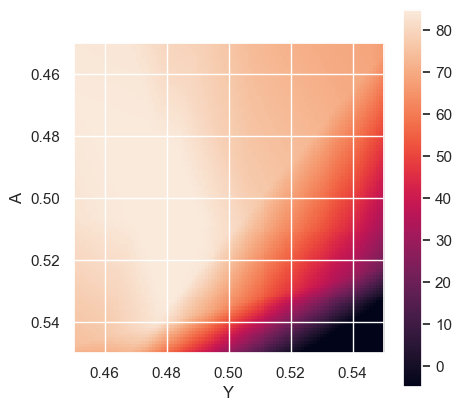}
        \caption{DuelDDQN}
        \label{fig:ddqnvals}
    \end{subfigure}
    \vskip\baselineskip
    \begin{subfigure}[b]{0.49\textwidth}
        \includegraphics[width=\textwidth]{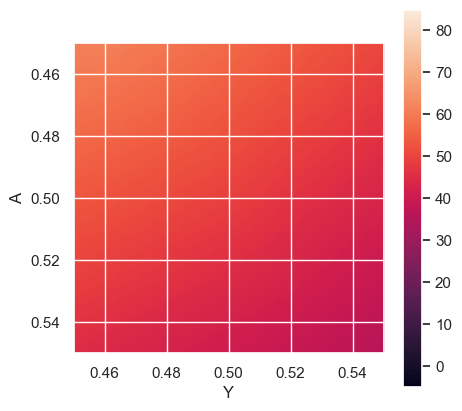}
        \caption{A2C}
        \label{fig:a2cvals}
    \end{subfigure}
    \begin{subfigure}[b]{0.49\textwidth}
        \centering
        \includegraphics[width=\textwidth]{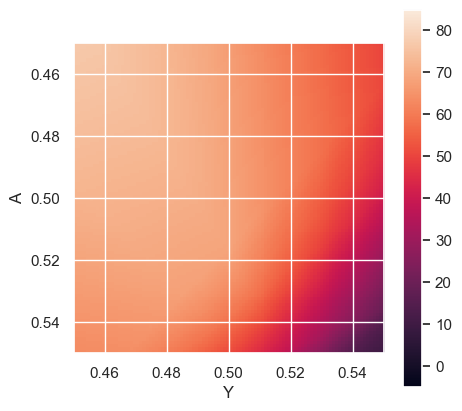}
        \caption{PPO}
        \label{fig:ppovals}
    \end{subfigure}
    \caption{State values of the possible initialisations: $(A\in[0.45,0.55], Y\in(0.45,0.55], S=0.5)$, coarse-grained for 10 000 initial states.}
    \label{fig:initvals}
\end{figure}
As described in Chapter \ref{chapter:methods}, we have random initialisation for the agents at each episode. This has a significant impact on the agent's trajectory and is believed to be the reason the agents have noisy reward curves. To understand this further, we analyse the possible initialisations. The initialisations are $(A\in[0.45,0.55], Y\in[0.45, 0.55], S=0.5)$, we present these initial states as a 2D matrix, where the states are more or less coarse-grained for computational reasons. 

First, we show the values of the initial states in Figure \ref{fig:initvals}. This is either $\max_a Q(s, a)$ for the DQN agents or simply the output of the value network for the Actor Critics. Multiple versions of each agent were trained with different seeds, representing an average is difficult to interpret for certain plots. Therefore, we choose to represent the best performing agent across the three seeds for consistency.

All the agents favour the low $A$, low $Y$ (low atmospheric carbon, low economic output) initialisations. These are the states the agent believes it can get a high reward from. The agents expect that they get little to no reward from the high $A$, high $Y$ initialisations. 

\begin{itemize}
\item DQN and DuelDDQN learn very similar values, where maximum values are around 80. We can check this estimate, the average reward varies around 1, (0.5 closer to initialisation, and 2 closer to the Green fixed point). We use a geometric series and the episode length of 600 time steps to estimate that a state that ends in the Green point has a value of $1\times\frac{1-\gamma^{600}}{1-\gamma} \approx 100$, this is a rough estimate but is in line with the state values we observe.

\item A2C's values are very low, most likely because it has not sampled the optimal states enough. A2C is updated through one-step TD, it needs to sample many trajectories for the value function to resemble the DQN values, it has sample efficiency problems. This is made evident by its slow learning, in Figure \ref{fig:default runs}.
A2C and PPO are on-policy; therefore, they update the value function with trajectory rewards they sampled (not the best-case trajectory like in DQN agents). This means that the learned value function will have lower values as it takes into account when the agent samples a “bad” action and consequently crosses a planetary boundary, into the value of the state. It is worth noting that PPO is using a $\lambda$ value of 0.8845, which means it looks at long horizon returns, which increases the value of a state since it incorporates more future rewards into the value. 
\end{itemize}
\newpage
Next, we show which of the four actions is picked by the agents first.
\begin{figure}[H]
    \centering
    \begin{subfigure}[b]{0.49\textwidth}
        \centering
        \includegraphics[width=\textwidth]{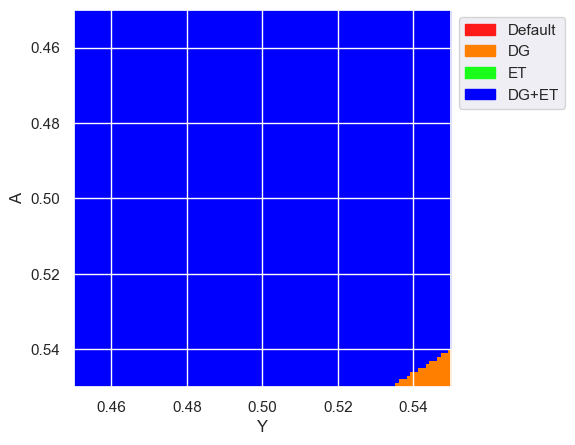}
        \caption{DQN}
        \label{fig:dqnact}
    \end{subfigure}
    \hfill
    \begin{subfigure}[b]{0.49\textwidth}
    \centering
        \includegraphics[width=\textwidth]{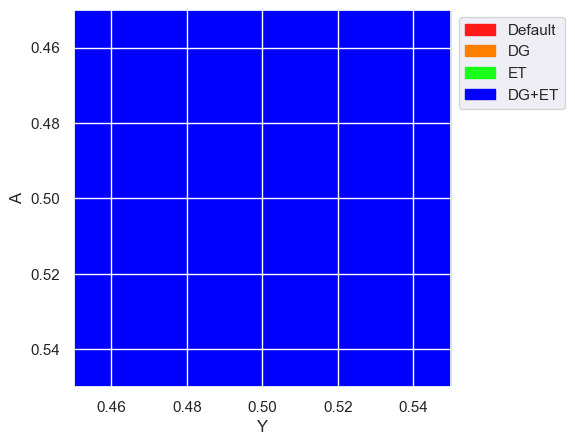}
        \caption{DuelDDQN}
        \label{fig:ddqnact}
    \end{subfigure}
    \vskip\baselineskip
    \begin{subfigure}[b]{0.49\textwidth}
        \includegraphics[width=\textwidth]{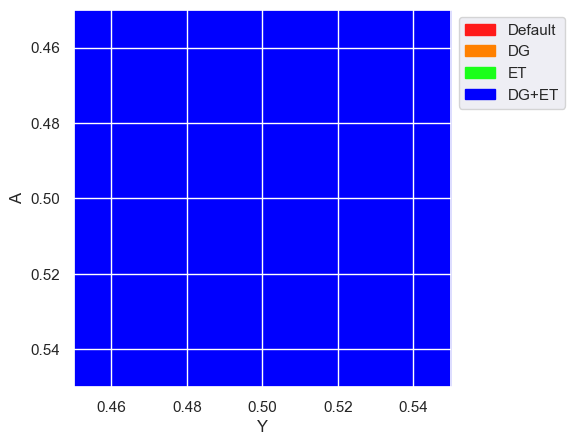}
        \caption{A2C}
        \label{fig:a2cact}
    \end{subfigure}
    \begin{subfigure}[b]{0.49\textwidth}
        \centering
        \includegraphics[width=\textwidth]{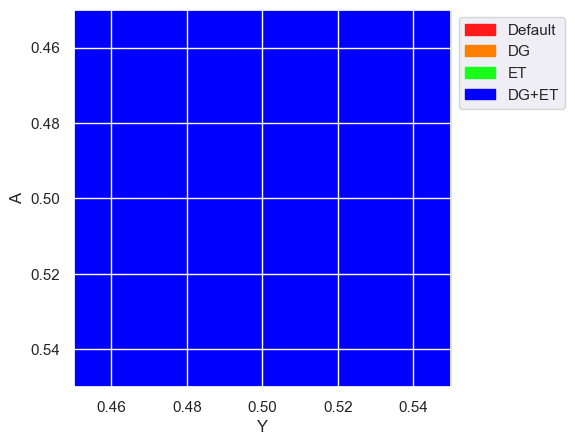}
        \caption{PPO}
        \label{fig:ppoact}
    \end{subfigure}
    \caption{The first action selected by the trained agents in the different initialisations: $(A\in[0.45,0.55], Y\in(0.45,0.55], S=0.5)$, coarse-grained for 10 000 initial states. The agents overwhelmingly prefer picking DG+ET as their first action.
    }
    \label{fig:initact}
\end{figure}
The trained agents all favour picking action DG+ET (Degrowth + Energy Transition) as the first action when starting the simulation, which reduces the growth of the economy by half as well as reduces the price of renewables.
The exception is the DQN agent who prefers to pick action ET in the high $A$, high $Y$ initialisations. This is not particularly significant as the agent expects to get very low rewards from these states anyway from Figure \ref{fig:dqnvals}. Using DG+ET as a first action makes sense in the context of Figure \ref{fig:initvals}, where the agents attribute more value to the low $A$, low $Y$ states. Action DG reduces the growth of the economic output $Y$ and action ET increases the amount of energy produced by renewables, increasing $S$, effectively reducing the increase in Atmospheric carbon. It therefore is sensible that the agent prefers to go to a low $A$, low $Y$ as soon as possible by picking the appropriate action.

Here, we present a matrix showing the final state of the trained agents at the different initialisations, \textit{n.b:} the $\epsilon$-greedy policy for the DQN agents was made fully greedy here to ensure that we get exemplary trajectories. 
\begin{figure}[H]
    \centering
    \begin{subfigure}[b]{0.49\textwidth}
        \centering
        \includegraphics[width=\textwidth]{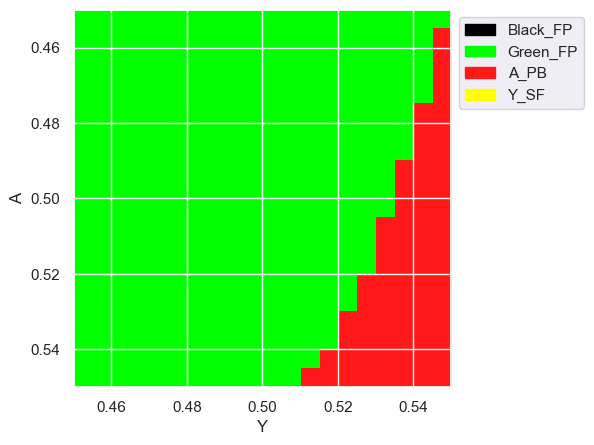}
        \caption{DQN}
        \label{fig:dqnend}
    \end{subfigure}
    \hfill
    \begin{subfigure}[b]{0.49\textwidth}
    \centering
        \includegraphics[width=\textwidth]{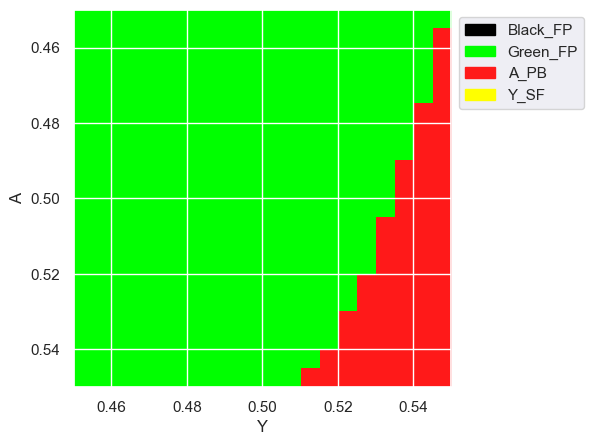}
        \caption{DuelDDQN}
        \label{fig:ddqnend}
    \end{subfigure}
    \vskip\baselineskip
    \begin{subfigure}[b]{0.49\textwidth}
        \includegraphics[width=\textwidth]{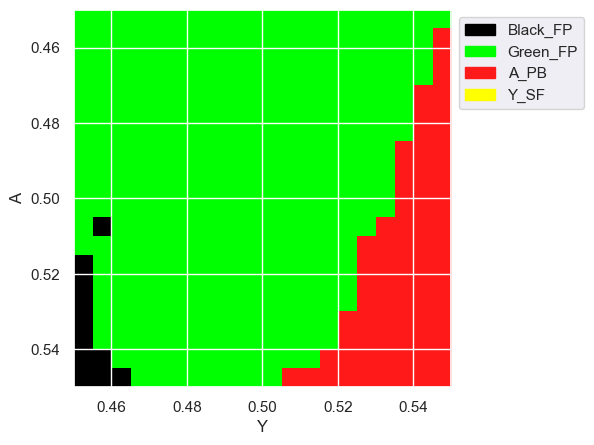}
        \caption{A2C}
        \label{fig:a2end}
    \end{subfigure}
    \begin{subfigure}[b]{0.49\textwidth}
        \centering
        \includegraphics[width=\textwidth]{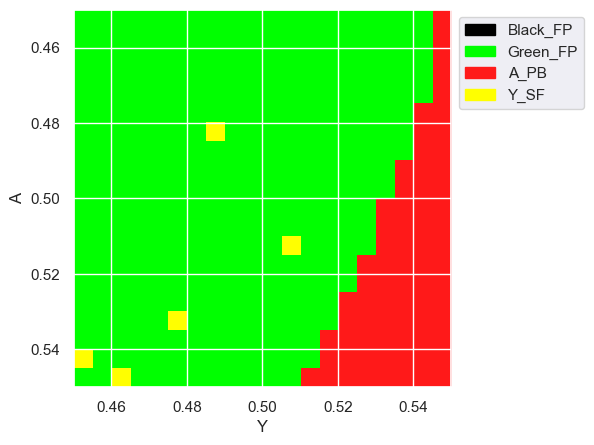}
        \caption{PPO}
        \label{fig:ppoend}
    \end{subfigure}
    \caption{Final state of the trajectory of the trained agents with the Planetary Boundary reward from the possible initialisations: $(A\in[0.45,0.55], Y\in(0.45,0.55], S=0.5)$, coarse-grained for 400 initial states.}
    \label{fig:initend}
\end{figure}
These plots resonate very well to the learned state values from Figure \ref{fig:initvals}, where high $A$, high $Y$ initialisations inevitably lead to crossing the carbon boundary $A_{PB}$. We see that almost any very high $Y$ initialisation leads to hitting the carbon boundary for all the agents.

DQN agents have identical results despite different value functions, which implies that, perhaps, this is the best performance the agents can achieve in this environment.

A2C shows a very similar pattern, except for slightly more states that end up at $A_{PB}$. Moreover, some states end up at the Black final point. PPO has a very similar matrix to the DQN agents, except for a few additional states that go to $A_{PB}$. Moreover, it has states that go to the economic boundary $Y_{SF}$ which change position depending on the seed we run the agent in, implying that this is because of the stochastic nature of the agent and not because the agent has learned something wrong. This lines up with our claim relating to Figure \ref{fig:ppovals} that the values for PPO are lower than the DQN agents, as the values are an average of stochastic trajectories.

\subsubsection{Simple Reward}
We compare with the Simple reward function end state matrices.
\begin{figure}[H]
    \centering
    \begin{subfigure}[b]{0.49\textwidth}
        \centering
        \includegraphics[width=\textwidth]{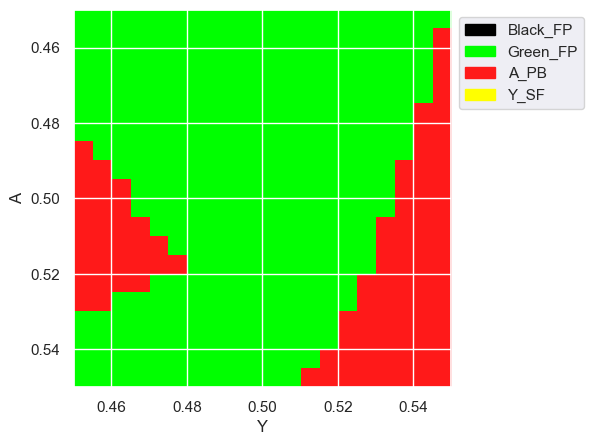}
        \caption{DQN}
        \label{fig:dqnendsimple}
    \end{subfigure}
    \hfill
    \begin{subfigure}[b]{0.49\textwidth}
    \centering
        \includegraphics[width=\textwidth]{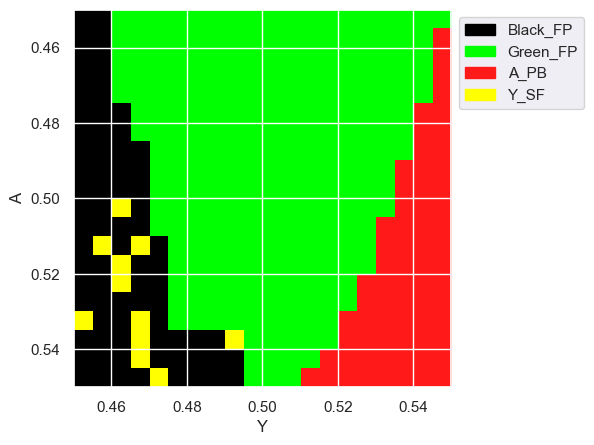}
        \caption{DuelDDQN}
        \label{fig:ddqnendsimple}
    \end{subfigure}
    \vskip\baselineskip
    \begin{subfigure}[b]{0.49\textwidth}
        \includegraphics[width=\textwidth]{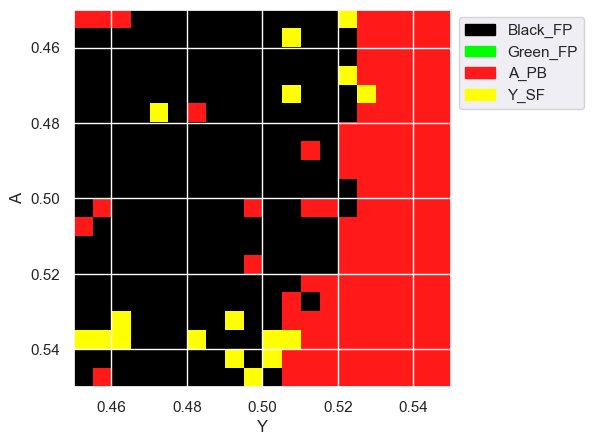}
        \caption{A2C}
        \label{fig:a2endsimple}
    \end{subfigure}
    \begin{subfigure}[b]{0.49\textwidth}
        \centering
        \includegraphics[width=\textwidth]{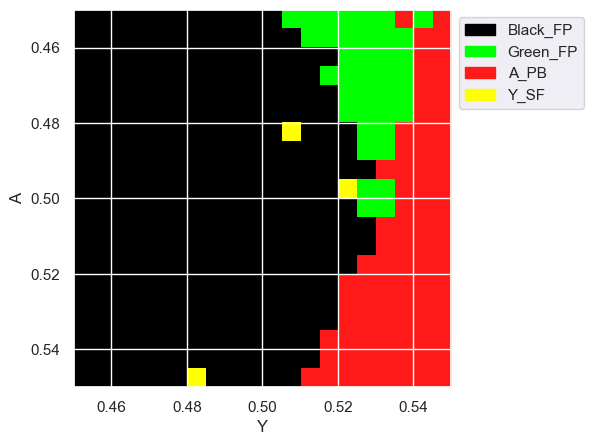}
        \caption{PPO}
        \label{fig:ppoendsimple}
    \end{subfigure}
    \caption{Final state of the trajectory of the agents trained with the \textit{Simple} reward from the possible initialisations: $(A\in[0.45,0.55], Y\in(0.45,0.55], S=0.5)$, coarse-grained for 400 initial states.}
    \label{fig:initendsimple}
\end{figure}

We can see that the learning is objectively worse than the PB reward function, where all the agents are less likely to reach the Green fixed point compared to Figure \ref{fig:initend}. As seen in the reward curves, DQN performs the best. Its end state matrix is also the closest to the Planetary Boundary experiment in Figure \ref{fig:dqntraj}. This shows that when the “hand holding” of the planetary boundaries is not present, it considerably reduces the agents' ability to learn to control the environment. This is most significant for the Actor Critics, which both do not learn to control the environment as accurately as the DQN agents. Without the guiding of the PB reward, they tend to go towards the Black final point. The edges in A2C's final state matrix are not smooth, they are jagged: indicating that the agent was perhaps not trained long enough. This is reinforced by its learning curve in Figure \ref{fig:simple_plot}, that has an upwards trend.

The sparse rewards have hurt the learning of the agents significantly, except for DQN.

\subsection{Feature Importance}
\subsubsection{Partial Observability Feature Importance}
We would like to understand how each variable of the state space influences the agent's decisions. To achieve this, we look at the importance of each feature; each state variable $A,Y$ and $S$. For feature importance, we use SHAP values (\textit{Shapley Additive Explanations}, \cite{Lundberg2017APredictions}). These values are unique for each data point, which here we take as states sampled from the environment. The SHAP values tell us how the numerical value of a feature contributes to the prediction. To collect data, we run the trained agents in the environment until we have collected 2000 states. We then randomly sample 500 states from this data to use for the SHAP values. This ensures that the states we use for feature importance are true samples from the environment that the agent would encounter.

We plot the SHAP value for the output of the network for each sampled state. For DQN agents, the network does not directly output the state value, it outputs Q-values instead. For accurate comparison, we calculate the state value: from Equation \eqref{eq:vtoq}, it is $V(s) = \mathbb{E}[Q(s,a)| \pi]$, where we take $\pi$ to be the greedy policy such that $V(s) = \max_a Q(s,a)$. We also present the maximum output of the actor networks from the Actor Critics. These represent the network's preferred action, we can interpret this as how a feature value impacts the decision of the agent.

\begin{figure}[hbt!]
    \centering
    \begin{subfigure}[b]{0.49\textwidth}
        \centering
        \includegraphics[width=\textwidth]{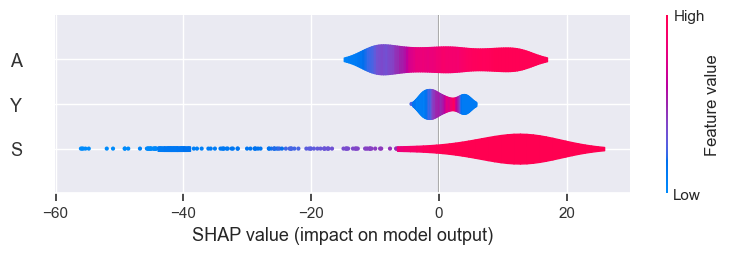}
        \caption{DQN}
        \label{fig:dqnfeatj}
    \end{subfigure}
    \hfill
    \begin{subfigure}[b]{0.49\textwidth}
    \centering
        \includegraphics[width=\textwidth]{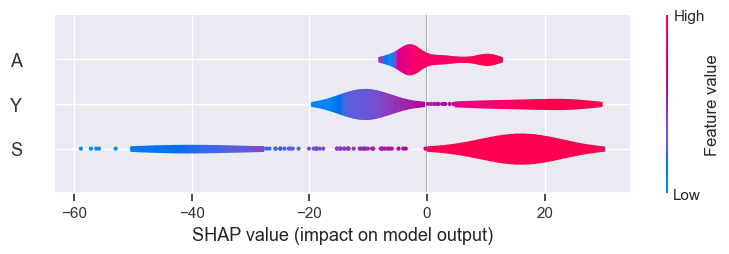}
        \caption{DuelDDQN}
        \label{fig:ddqnfeat}
    \end{subfigure}
    \vskip\baselineskip
    \begin{subfigure}[b]{0.49\textwidth}
        \includegraphics[width=\textwidth]{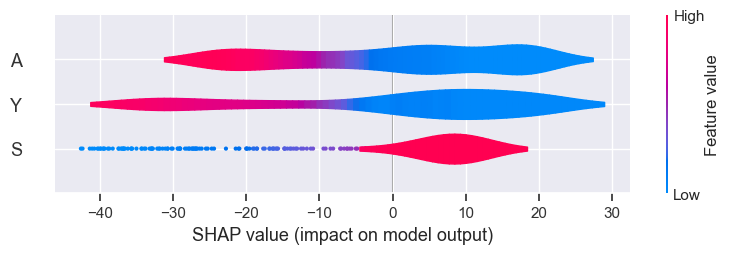}
        \caption{A2C critic}
        \label{fig:a2feat}
    \end{subfigure}
    \begin{subfigure}[b]{0.49\textwidth}
        \centering
        \includegraphics[width=\textwidth]{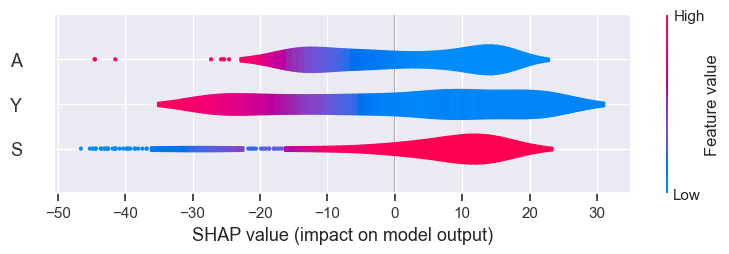}
        \caption{PPO critic}
        \label{fig:ppofeat}
    \end{subfigure}
    \vskip\baselineskip
    \begin{subfigure}[b]{0.49\textwidth}
        \includegraphics[width=\textwidth]{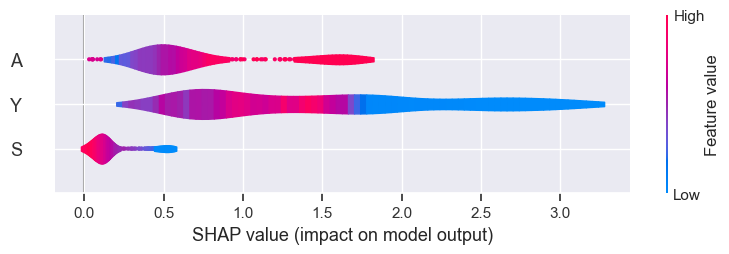}
        \caption{A2C actor}
        \label{fig:a2featact}
    \end{subfigure}
    \begin{subfigure}[b]{0.49\textwidth}
        \centering
        \includegraphics[width=\textwidth]{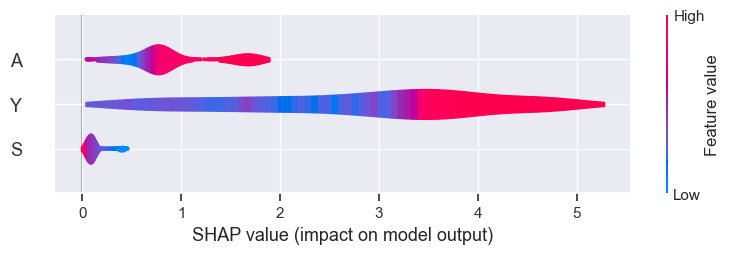}
        \caption{PPO actor}
        \label{fig:ppofeatact}
    \end{subfigure}
    \caption{Violin plots of 500 SHAP values for the state value estimation with the Planetary Boundary reward. For DQN agents, the state value was taken as $\max_a Q(s,a)$. The bottom two plots present the maximum output of the actor network (the maximum preference). \textit{n.b:} here, \textit{feature values} refers to the numerical value of a feature and not the output of the state value function.}
    \label{fig:feature}
\end{figure}

In Figure \ref{fig:feature} below, the input feature value is colour coded from blue (low input value) to red (high input value). Then the output contribution of the feature for each data point is on the horizontal axis. The plots show some similarities between the same agent types. But across agent types, they are extremely diverse. 

For DQN agents, low atmospheric carbon $A$ and low renewable knowledge stock $S$ give yield high negative values but, high $A$ and high $S$ give high positive values. $Y$ has little to no impact on the model in DQN but, in DuelDDQN, higher economic output $Y$ outputs higher values.

We see that the opposite is true for the actor critics. Low inputs of $A$ are much more valuable, and so are small values of $Y$. On the other hand, high $A$ leads to negative impact, and so does high $Y$. Higher $S$ returns higher values and vice-versa. 

The difference in the two classes of agents is most likely due to the fact that the state value in the Actor Critics is used to optimise the policy and not used directly to take actions. The actor networks takes decisions, they learn what actions to take with a stochastic policy. The actor networks for both Actor Critics show little interest in the scale of $S$. Higher $A$ causes a higher action preference. They seem to pay most attention to the amount of $Y$. Interestingly, A2C puts high importance on low $Y$ and low importance for high $Y$, PPO does the opposite. This is most likely due to the fact that A2C takes a long time to train and in the process hits the economic boundary very often and so learns to avoid low $Y$ as much as possible.

The agents seem to value an Earth in a low-economy, low-carbon regime. From the equations, this is because the growth of the economy is exponential. As a result, the energy requirements for this economy are exponential. If renewable knowledge stock does not “keep up” with economic output, the agent will inevitably cross the atmospheric carbon planetary boundary. This can also be seen in the actor network SHAP plots, high carbon or strong economy requires action. On the other hand, low $Y$ also is important for the agent's decision-making, if the economic output is too weak, it risks crossing the social foundation boundary. 

Overall, the agents' interpretation of the environment is extremely diverse, despite all reaching the Green fixed point similarly often.

\subsubsection{Markov State Feature Importance}

\begin{figure}[H]
    \centering
    \begin{subfigure}[b]{0.49\textwidth}
        \centering
        \includegraphics[width=\textwidth]{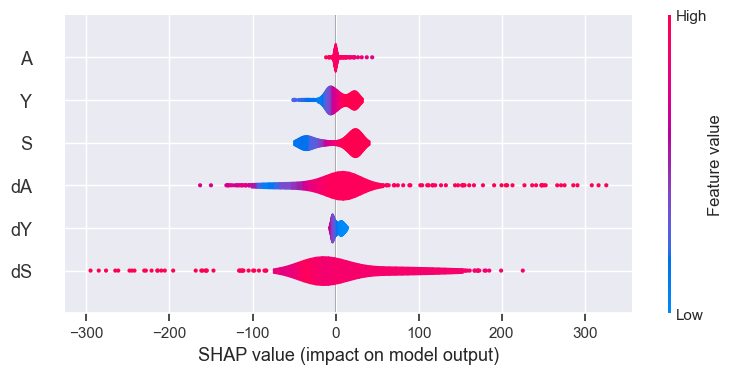}
        \label{fig:mdqnfeatj}
    \end{subfigure}
    \hfill
    \begin{subfigure}[b]{0.49\textwidth}
    \centering
        \includegraphics[width=\textwidth]{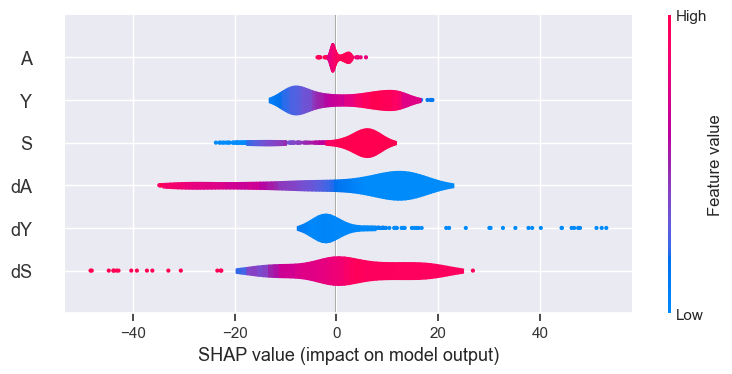}
        \caption{DuelDDQN}
        \label{fig:mddqnfeat}
    \end{subfigure}
    \vskip\baselineskip
    \begin{subfigure}[b]{0.49\textwidth}
        \includegraphics[width=\textwidth]{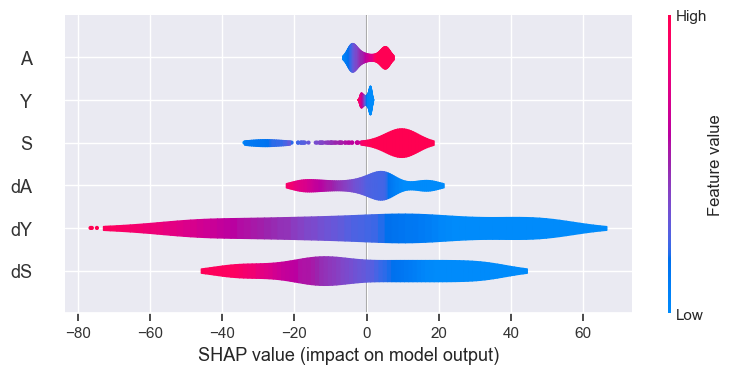}
        \caption{A2C critic}
        \label{fig:ma2feat}
    \end{subfigure}
    \begin{subfigure}[b]{0.49\textwidth}
        \centering
        \includegraphics[width=\textwidth]{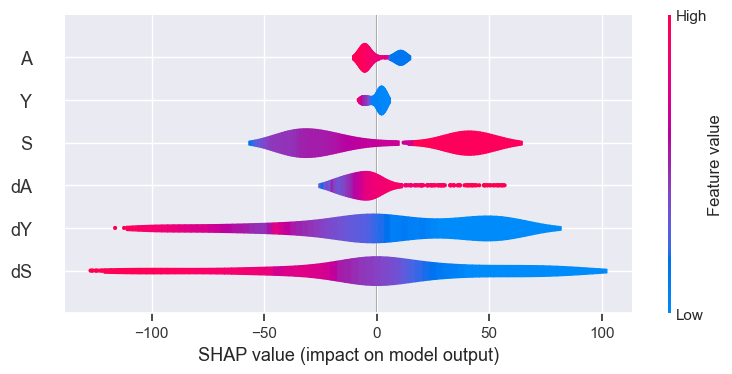}
        \caption{PPO critic}
        \label{fig:mppofeat}
    \end{subfigure}
    \vskip\baselineskip
    \begin{subfigure}[b]{0.49\textwidth}
        \includegraphics[width=\textwidth]{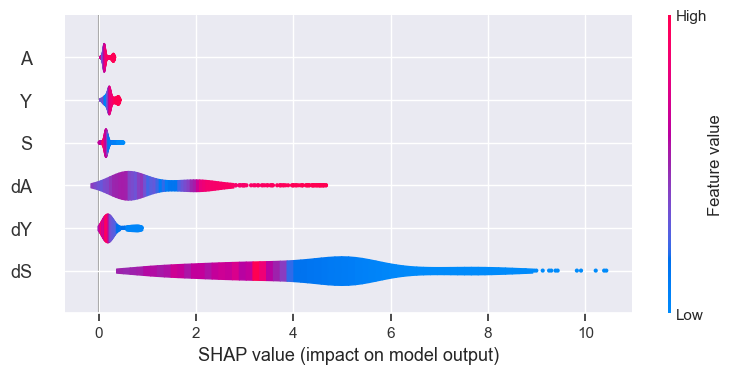}
        \caption{A2C actor}
        \label{fig:ma2featact}
    \end{subfigure}
    \begin{subfigure}[b]{0.49\textwidth}
        \centering
        \includegraphics[width=\textwidth]{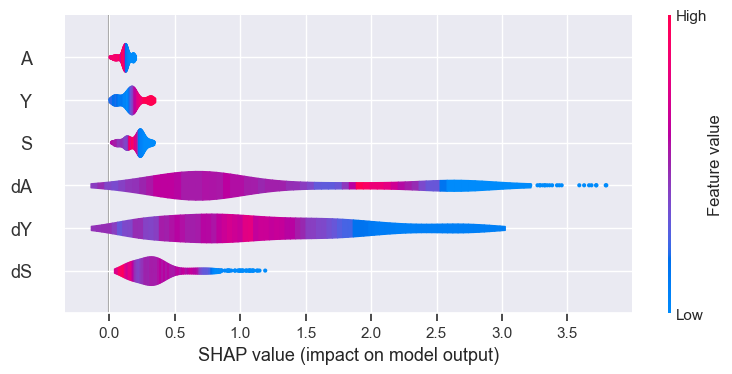}
        \caption{PPO actor}
        \label{fig:mppofeatact}
    \end{subfigure}
    \caption{Violin plots of 500 SHAP values for the value estimation with full Markov state. For Q-value based agents, the value was taken as $\max_a Q(s,a)$. The bottom two plots present the maximum output of the actor network (the maximum preference). \textit{n.b}: here, feature values refers to the numerical value of a feature and not the output of the state value function}
    \label{fig:markovfeature}
\end{figure}
To check if the additional information is being used by the agent when using a fully observed environment, we look at the feature importance plots.
We can see in these plots that the velocity is indeed extremely significant to both the state value estimation and the action selection (for the Actor Critics). For the Actor Critics, we can see in Figure \ref{fig:ma2featact} and \ref{fig:mppofeatact}, the velocity of the state variables is much more significant when it comes to taking actions than the numerical value of the state variables. 

Again, we have enormous amounts of variability in the interpretation of the environment by the agents.

\subsection{Hyperparameter Analysis}\label{section:hparam}

During the early stages of this work, it was quickly remarked that the agents are very sensitive to hyperparameters, where small changes across various parameters can lead to drastic increases or decreases in performance. As a result, we use the acquired hyperparameter tuning data to train a random forest \cite{Breiman2017ClassificationTrees} with the hyperparameters as features and the target metric as label (the mean reward across 100 000 frames). This enables us to calculate the \textit{Mean Decrease in Impurity} or \textit{Gini Importance}. This measurement sums the number of splits that include the feature proportionally to the number of samples it splits, across all the trees. It can be understood as the impact of the hyperparameter on the model performance. We also present correlation plots. these are simply the correlation between the features (hyperparameter data) and the label (metric).

\subsubsection{DQN agents}
\begin{figure}[hbt!]
    \centering
    \begin{subfigure}[b]{0.49\textwidth}
        \centering
        \includegraphics[width=\textwidth]{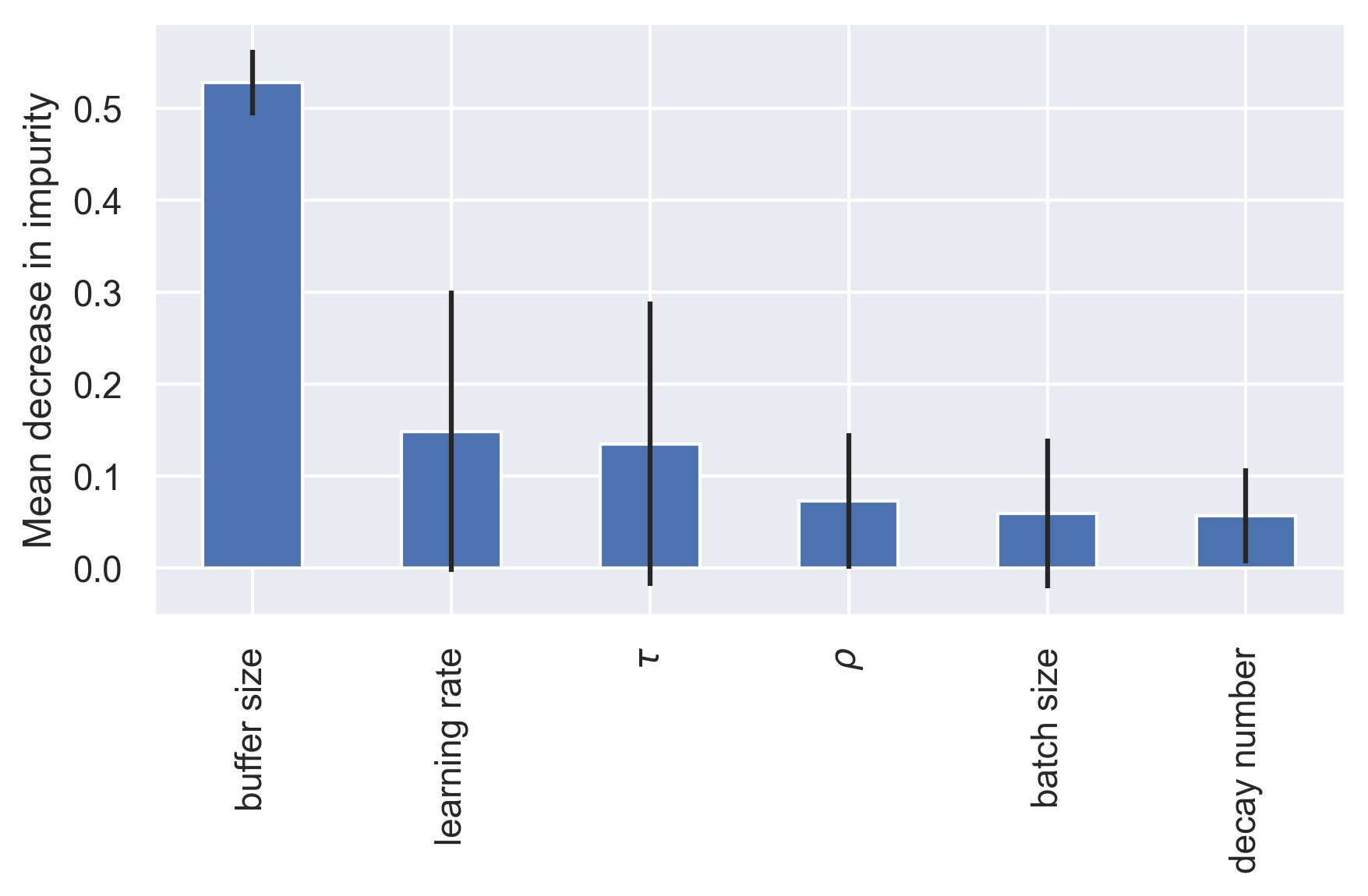}
        \caption{Feature Importance for DQN}
        \label{fig:dqnforest}
    \end{subfigure}
    \hfill
    \begin{subfigure}[b]{0.49\textwidth}
    \centering
        \includegraphics[width=\textwidth]{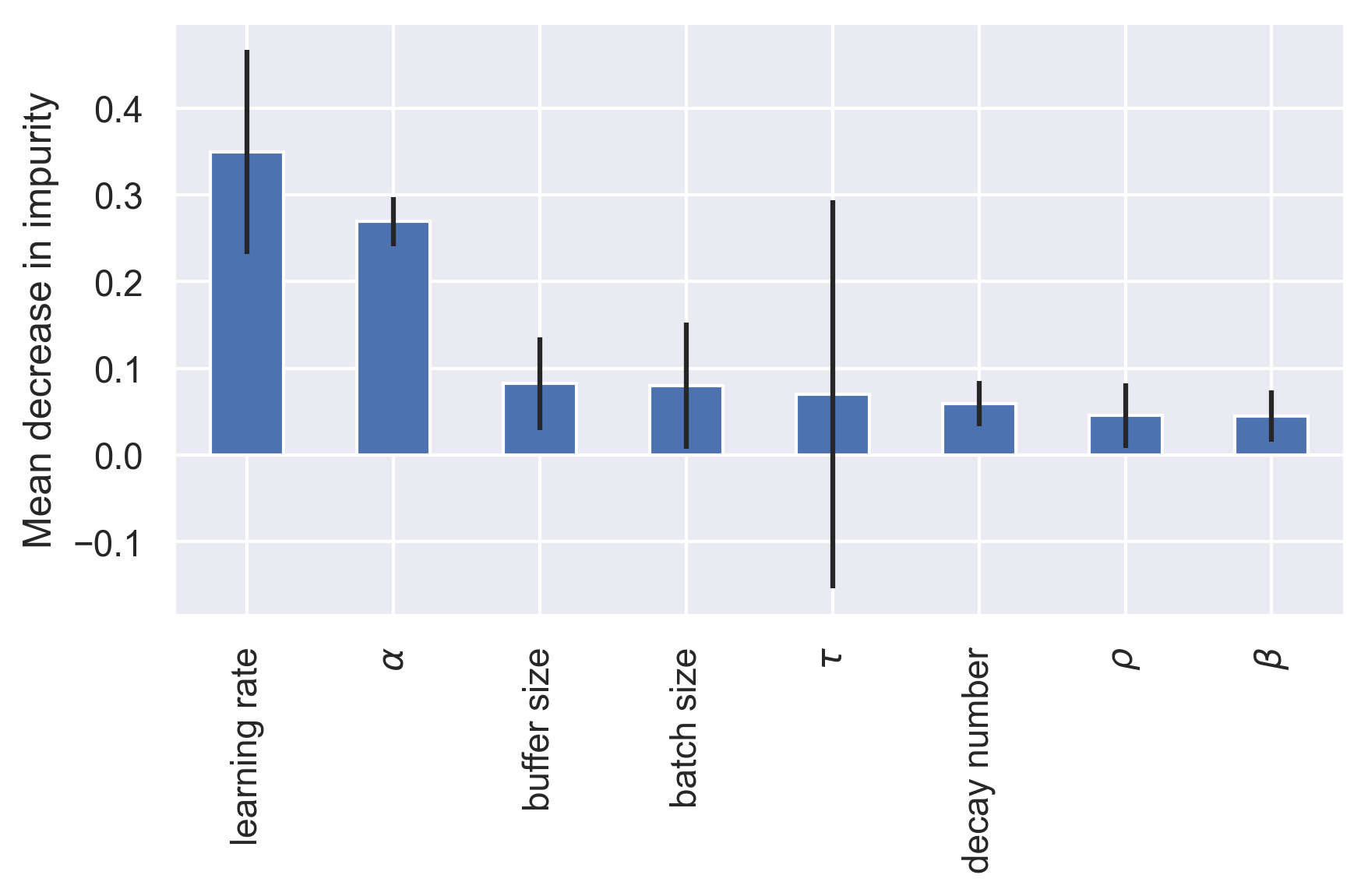}
        \caption{Feature Importance for DuelDDQN}
        \label{fig:ddqnforest}
    \end{subfigure}
    \vskip\baselineskip
    \begin{subfigure}[b]{0.49\textwidth}
        \centering
        \includegraphics[width=\textwidth]{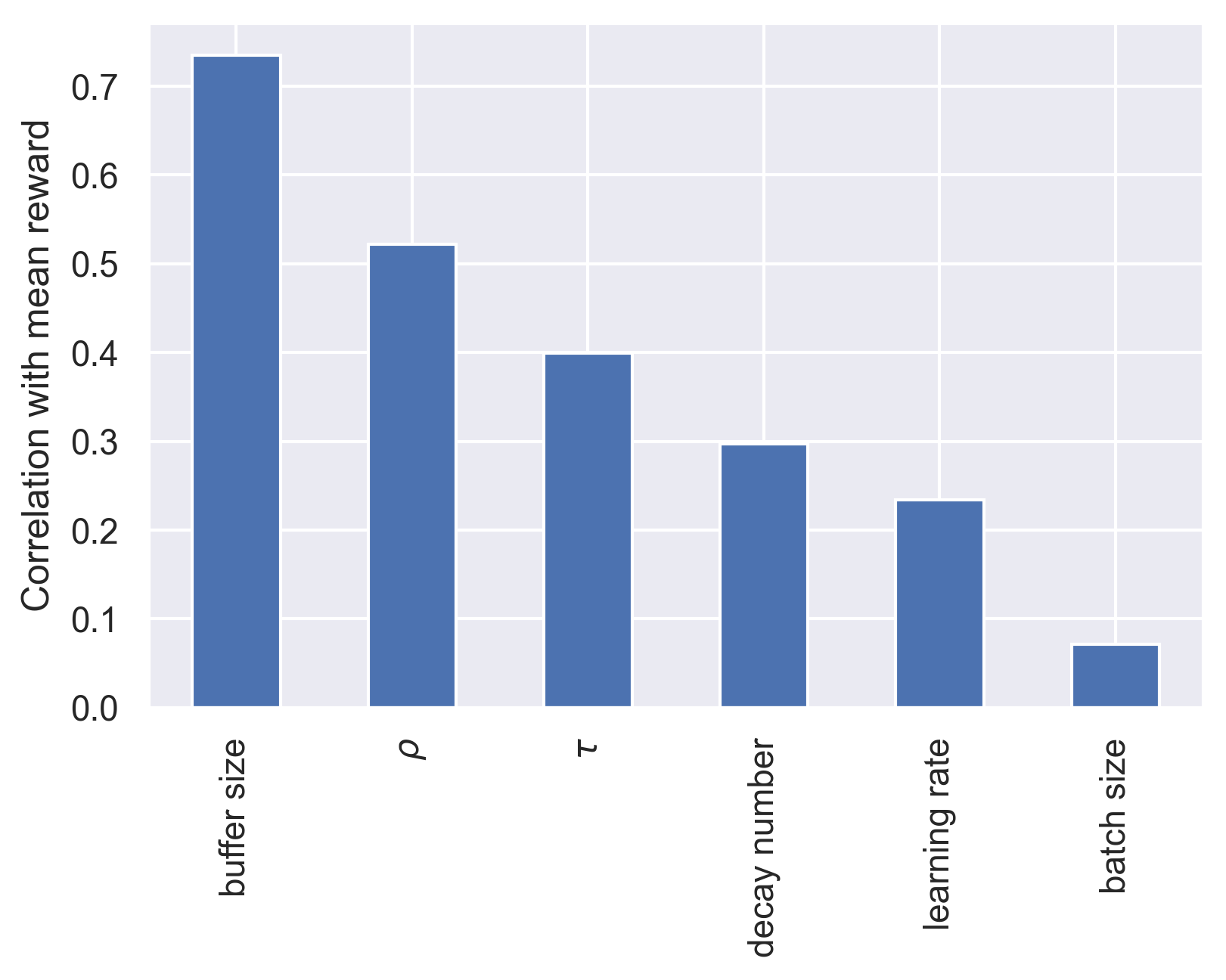}
        \caption{Correlation for DQN}
        \label{fig:dqncorr}
    \end{subfigure}
    \hfill
    \begin{subfigure}[b]{0.49\textwidth}
    \centering
        \includegraphics[width=\textwidth]{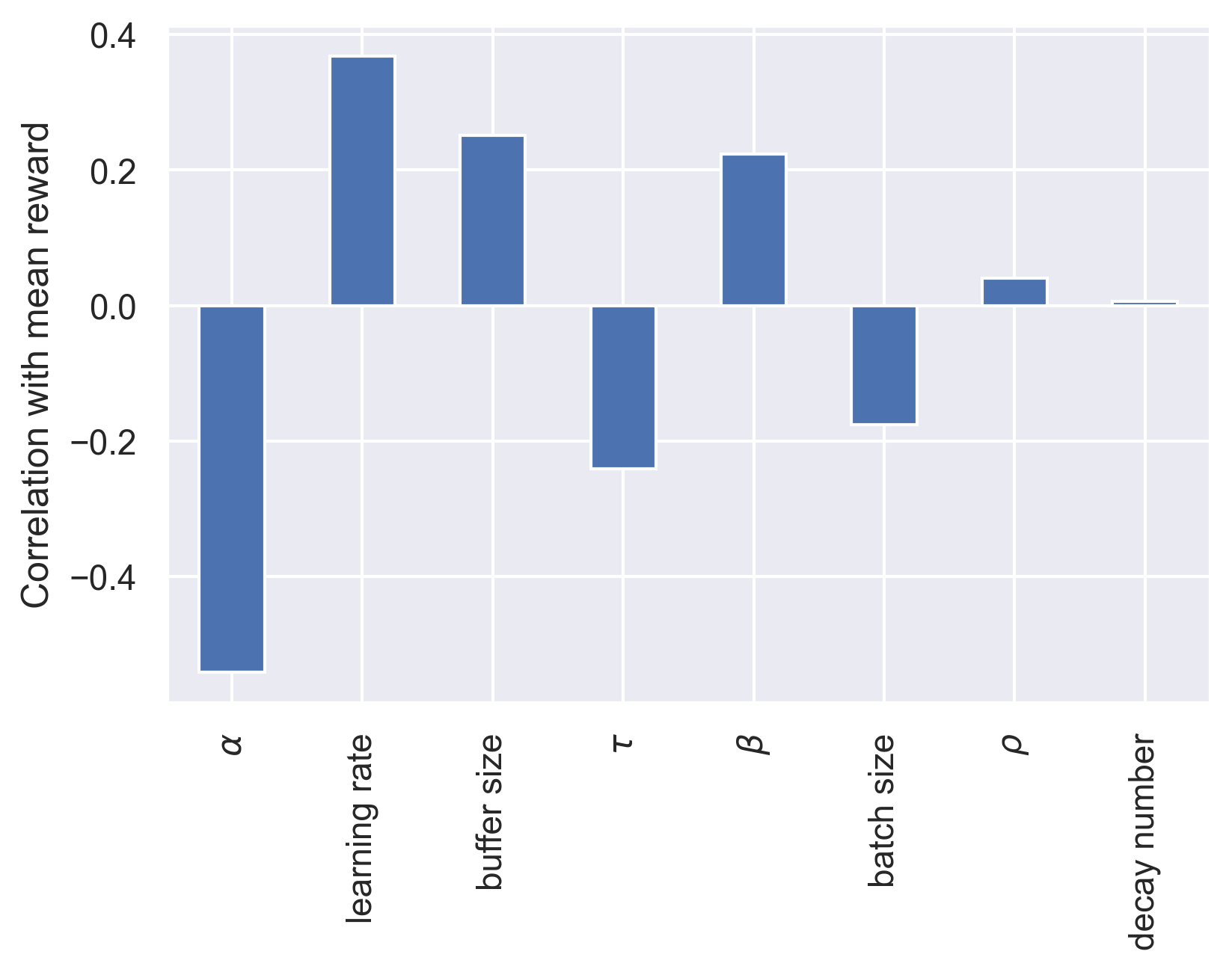}
        \caption{Correlation for DuelDDQN}
        \label{fig:ddqncorr}
    \end{subfigure}
    \caption{Plots of the feature importance sorted by mean impurity of a trained random forest (top), and correlation of features to label (bottom). Features are the hyperparameter tuning runs for the DQN agents (32 runs for DQN, 112 runs for DuelDDQN), labels are the metric: the mean reward.  The black bars are the standard deviation of the impurity across all trees in the forest.}
    \label{fig:dqnimp}
\end{figure}

In Figure \ref{fig:dqnimp} below, DQN depends heavily on the buffer size, as well as the learning rate and the target network update rate ($\tau$). We can see that the buffer size has a strong positive correlation with the mean reward. We can see for the learning rate and $\tau$ there is a high variance in the importance. $\tau$ also seems to have a strong positive correlation, updating the target network more often improves the learning of the agent. This parameter exists to regulate the overestimation bias, a common problem in Q-learning, where an agent overestimates the value of an action due to random sampling of a high reward. Overestimation bias is common in stochastic environments \cite{Sutton2020ReinforcementIntroduction}. This environment is completely deterministic given an initialisation which could indicate then that overestimation bias is less of an issue. The value found for the periodic update $\frac{1}{\tau}=12$, is very low compared to the recommended value of 10 000 \cite{VanHasselt2010DoubleQ-learning}, but closer to the value used by Strnad et al. \cite{Strnad2019DeepStrategies}, who use 100. We believe this to be due to the fact that the Bayesian optimisation was maximising the mean reward and so aims to maximise the reward as fast as possible, ignoring possible overestimation bias. The parameter controlling the exploration decay $\rho$ has a strong positive correlation with the mean reward, the DQN agent does not require extensive exploration to maximise its reward.

DuelDDQN also depends much more on the learning rate, which has a strong positive correlation with the mean reward. The $\alpha$ parameter that controls the prioritisation of the experience in the buffer has a strong negative correlation with the reward. This agrees with our findings as $\alpha$ was found to have an optimal value of 0.213 which is lower than the recommended range of 0.4-0.6 \cite{Schaul2015PrioritizedReplay}. Sampling from the buffer uniformly is beneficial to the learning of the agent. This implies that most of the experiences are valuable to the agent, there are few “high priority” experiences. Again, the $\tau$ parameter has medium importance but very high variance among the trees.  

The addition of the prioritised importance sampling buffer is a good explanation as to why the buffer size matters less for the DuelDDQN than for DQN. DuelDDQN can directly access the most relevant experiences more frequently and so does not need to store as much. More generally, we show that overestimation bias is not of critical importance in this environment and that more uniform sampling of experiences improves performance. This can explain why the performance of DQN is so similar to DuelDDQN: two of the upgrades (double-Q and PER-IS) are not very relevant to improving performance.

\subsubsection{Actor Critics}
\begin{figure}[H]
    \centering
     \begin{subfigure}[b]{0.49\textwidth}
        \includegraphics[width=\textwidth]{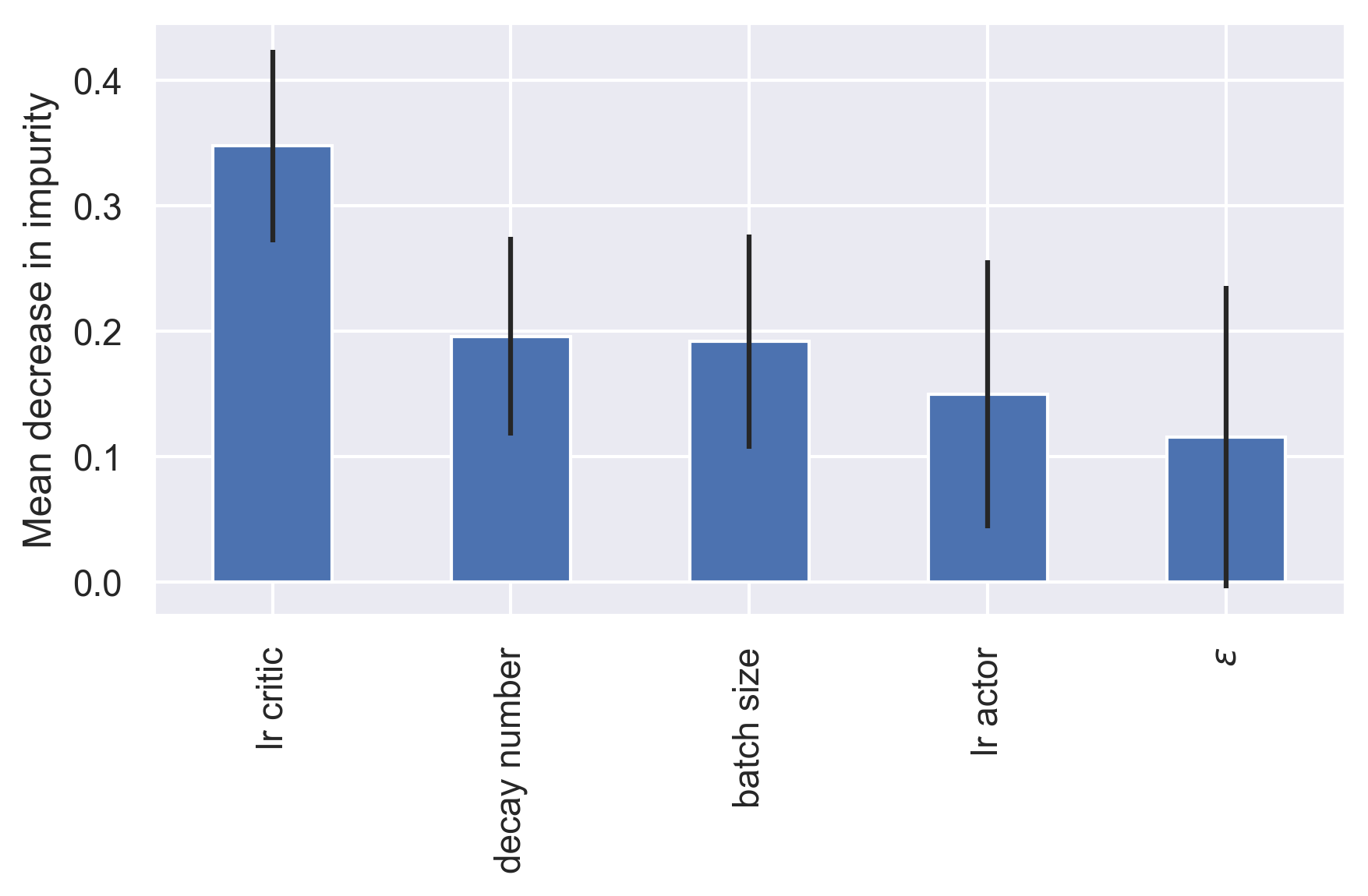}
        \caption{Feature Importance for A2C}
        \label{fig:a2cforest}
    \end{subfigure}
    \begin{subfigure}[b]{0.49\textwidth}
        \centering
        \includegraphics[width=\textwidth]{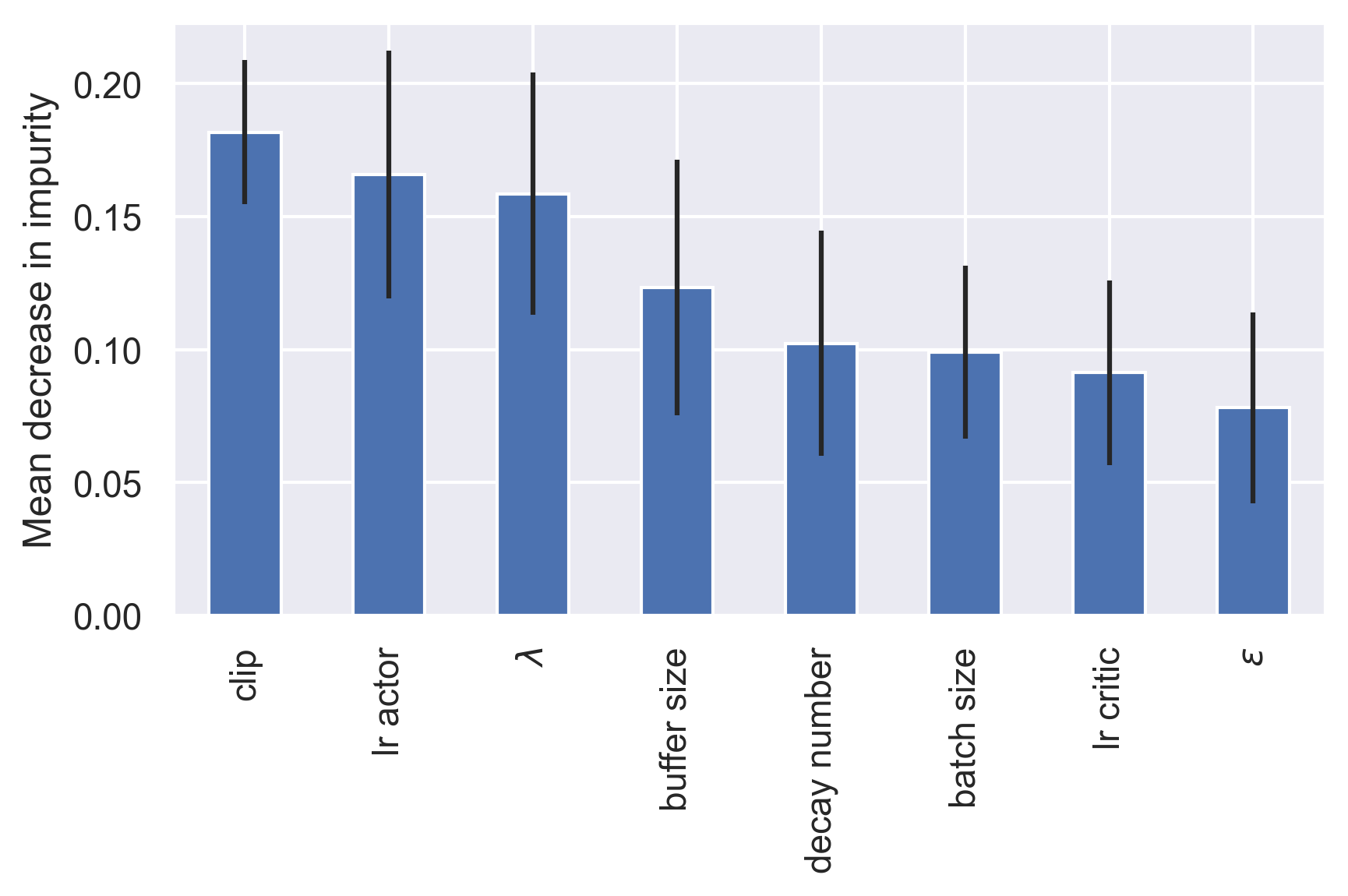}
        \caption{Feature Importance for PPO}
        \label{fig:ppoforest}
    \end{subfigure}
    \vskip\baselineskip
    \begin{subfigure}[b]{0.49\textwidth}
        \includegraphics[width=\textwidth]{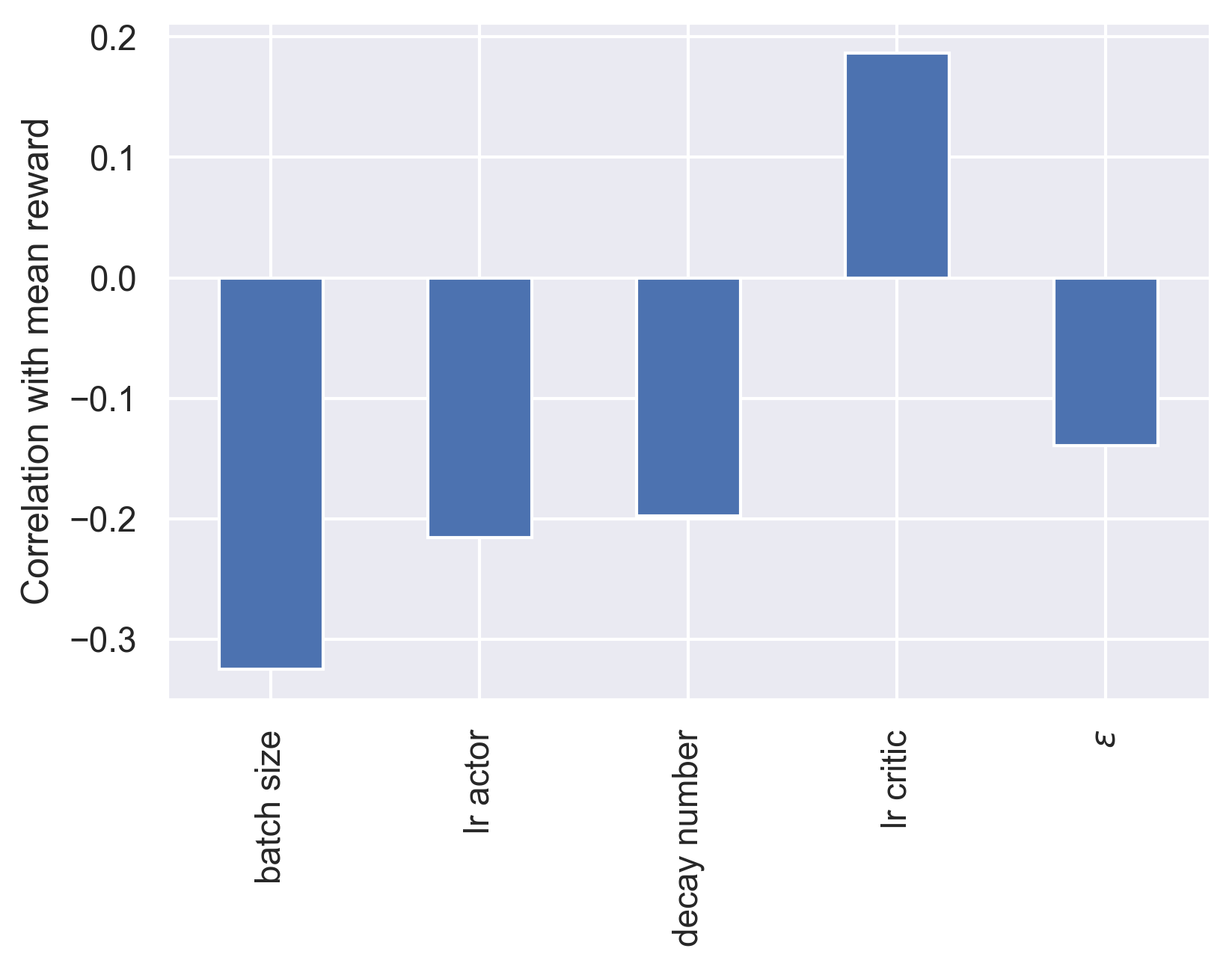}
        \caption{Correlation for A2C}
        \label{fig:a2ccorr}
    \end{subfigure}
    \begin{subfigure}[b]{0.49\textwidth}
        \centering
        \includegraphics[width=\textwidth]{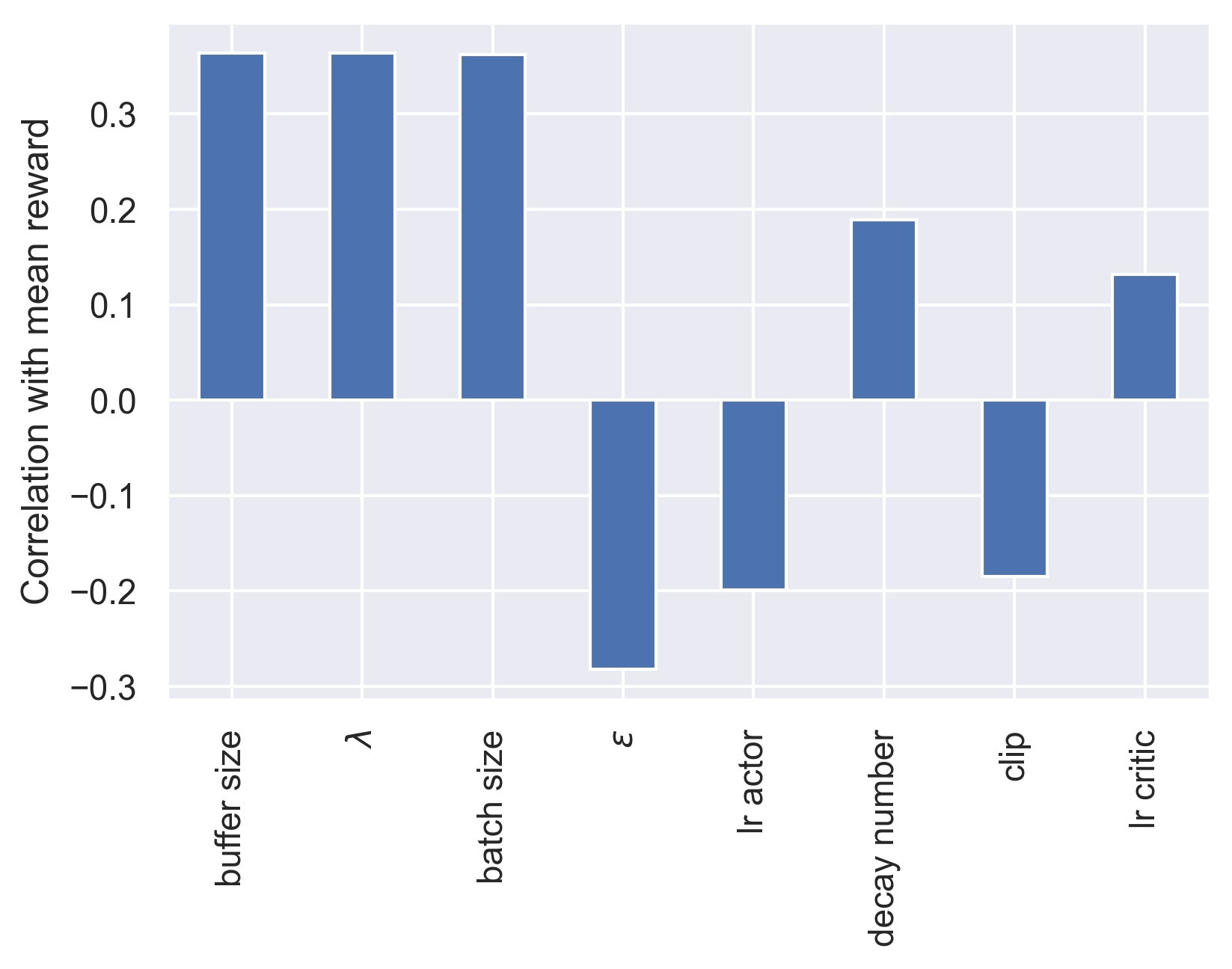}
        \caption{Correlation for PPO}
        \label{fig:ppocorr} 
    \end{subfigure}
    \caption{Plots of the feature importance sorted by mean impurity of a trained random forest (top), and correlation of features to label (bottom). Features are the hyperparameter tuning runs for the Actor Critic agents (67 runs for A2C, 229 runs for PPO), labels are the metric: the mean reward. The black bars are the standard deviation of the impurity across all trees in the forest.}
    \label{fig:acimp}
\end{figure}

For A2C, the parameters are more or less equally significant. The most important being the learning rate of the critic, which has a positive correlation with the mean reward. All the other parameters have negative correlations.

For PPO, the most significant parameter is the \textit{clip} parameter, which has small negative correlation: better rewards are obtained when the policy is not allowed to diverge too much. A significant parameter is the $\lambda$ parameter, which has a strong positive correlation. The agents learn better in a long horizon, high variance, low bias setting. Introducing long horizon returns to the DQN agents or A2C could improve their learning. For both A2C and PPO, the exploratory parameter $\epsilon$, which is the entropy regularisation coefficient, has a negative correlation, implying just as in for the DQN agents that exploration is not essential in this environment for maximising rewards.

The \textit{decay number} parameter is much more significant for the Actor Critics than for the DQN agents. The Actor Critics seem to require more decay to obtain higher rewards. The effect is clear for PPO, which requires 200 decays for better convergence.
\\

Overall, the effect of the hyperparameters is pivotal. The choice of values can significantly change the learning of the agents.

%% file: chapters/5.discussion.tex
\section{Discussion of Results}
In this chapter, we attempt to draw meaningful conclusions to the experiment. Additionally, we answer whether the agents have brought forward some aspects of climate change and if they teach us something about how to solve it. We also highlight the drawbacks of the methods we applied here. 

\subsection{Comparison with Previous Results}

These results beat the previous benchmark of the best average reward of 250 for the Duelling DDQN with Prioritised Experience Replay and Importance Sampling \cite{Strnad2019DeepStrategies}. Where we achieved a mean reward of around 400 with PPO. This is most likely due to the rigorous Bayesian hyperparameter tuning we conducted, where all the hyperparameters were tuned (except for number of layers and number of units for the neural networks). The previous work hand tuned only certain parameters. 

\subsection{What Have We Learnt?}
Overall, the agents can solve the environment and find optimal pathways to a sustainable future by using specific sequences of actions. There were significant issues with sensitivities to hyperparameters, which were very difficult to tune. The off-policy agents were much more flexible with learning the environment in different experiments, which is clear from their higher average success rate and consistency across the experiments. The on-policy agents were lacking in exploration, which significantly hurt their performance in different experiments.

The type of reward function has a drastic effect on the learning of the agents. For example, the Actor Critics were unable to learn to solve the environment when a cost was added to using the actions, whereas the DQN agents were relatively unaffected. If this framework is pushed closer to real-world application, the choice of reward function is incredibly significant for discovering strategies towards a sustainable future. 

Adding small amounts of noise made the agents more robust, which was not initially anticipated. Moreover, making the environment fully observable did help the agents, where the highest success rate was obtained by the DuelDDQN agent utilising the full Markov state. We did find that the agents can still learn equivalently well in a partially observed environment if fine-tuned to it. This is promising as it shows that the agents can learn in noisy environments where things are uncertain and not always observable, they are very adaptable. This brings them one step closer to real-world application.

The agents tell us that a small economic output World-Earth System is the easiest to steer towards a sustainable future. The system becomes much harder to control as the economic output grows faster than the renewable energy generation. If the economic output is too high, the agents are unable to control the environment and the carbon planetary boundary is crossed. The agents sometimes find that the long-term reward of reaching a sustainable future is worth more than the short-term reward of maximising only the economic output, but not always, which was shown with the Policy Cost reward experiment.

\subsection{Challenges}
A significant amount of this work was focussed on tuning hyperparameters, which we showed sometimes backfired, as we appear to have overfitted to the PB reward at the expense of the other reward functions. Much of the theory for Deep RL (and Deep Learning more generally) is very far behind the practice, and so there are only heuristics for tuning these hyperparameters. The sensitivity of the agents to hyperparameters shows the flaws of RL for this type of problem. If such a model was used for policy, a bad tuning of hyperparameters could lead to incorrect results and consequently affect policy. This could have disastrous consequences for both humans and the Earth. This could be circumvented by having an ensemble of agents guide policy and a human agent interpreting the results to take an informed decision.

There are significant challenges to interpreting these models, we showed some methods, but these are not sufficient. Analysing the decision-making of the agent is difficult, and the excessively large number of results can be overwhelming for humans to understand. Here, we used a 3-dimensional environment so that we could plot our results in phase space, but more complex World Earth System models have higher dimensionality. This makes them difficult to plot and thus to understand. 

The most challenging aspect is the accuracy of the model of the environment. In this work, we have used a basic World-Earth System model to test the framework. This model is limited by the small number of actions, variables, interactions, as well as having only two fixed points that the Earth can go to. Applying this framework to informing policy is dangerous until we have models we can trust. We need to wait for additional research on the effects and causes of climate change, and how these interact with humans, to create better models.

\section{Conclusion and Extensions}

The conclusion that a slower economy gives more opportunity to control the environment resonates with the recent Covid-19 pandemic, where $CO_2$ emissions sharply dropped as the global economy was taken to a standstill \cite{Tollefson2021CarbonDip}. As the economy restarted, the emissions came back up. This close relationship between the strength of the economy and emissions is important to consider for policy. This model also leads us to the conclusion that slowing the economy is important to give time for the renewable energy research to overtake fossil fuel energy: if the economy is left unchecked, it grows too fast for renewable energy growth to sustain it alone. This is nevertheless just a model, and a very simplistic one at that. 

There are many extensions that have not been explored in this work. Changes that were not looked at were changes in the number of actions that can be taken per year, we set this to one throughout this work, but there is no particular reason for this, apart from the easily interpretable idea of one policy per year. In this work, we focussed more on the interpretable aspect and thus aimed to leave the dynamics of the model from Kittel et al. \cite{Kittel2017FromManagement} untouched. Additional actions or continuous actions are a clear avenue for probing the environment in different ways. There is also research in \textit{Explainable Artificial Intelligence} (XAI) that could be integrated in this framework, specifically: explainable RL \cite{Juozapaitis2019ExplainableDecomposition,vanderWaa2018ContrastiveConsequences}. This would help with explainability of the agents and interpreting their decisions.

As early work, we believe this to be very promising. It appears that the bottleneck of applicability is on the side of the models more than on the side of Reinforcement Learning. Better models are needed for RL to truly show us innovative and applicable solutions, as here we are very much limited by the models' complexity.

%% file: chapters/6.appendix.tex
\appendix

\chapter{Plots} \label{graphs}

\begin{figure}
    \begin{subfigure}[b]{0.49\textwidth}
        \includegraphics[width=\textwidth]{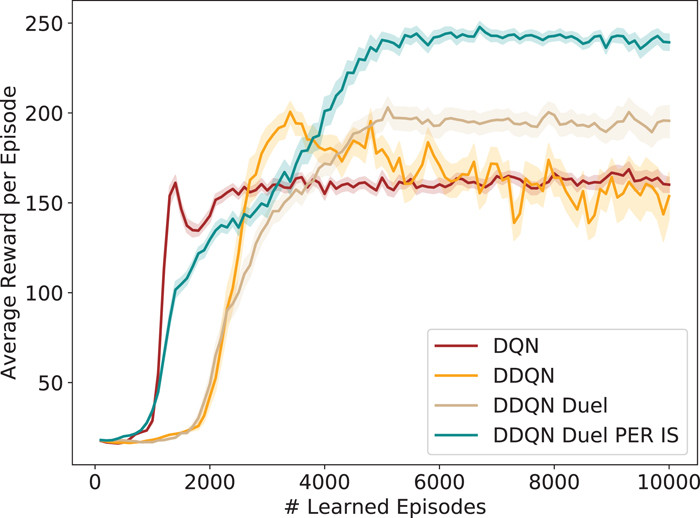}
        \caption{Moving Average Rewards}
        \label{fig:strnadplot}
    \end{subfigure}
    \begin{subfigure}[b]{0.49\textwidth}
        \centering
        \includegraphics[width=\textwidth]{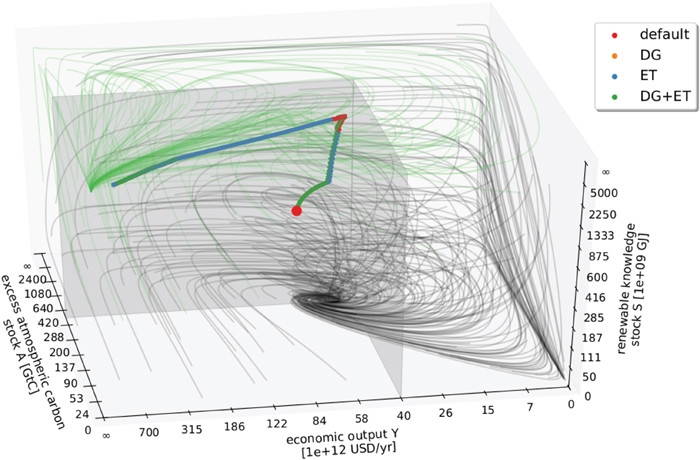}
        \caption{Trajectory}
        \label{fig:strnadtraj}
    \end{subfigure}
    \caption{(a) Plot of the reward curves from Strnad et al. \cite{Strnad2019DeepStrategies}. (b) Example trajectory from Strnad et al. \textit{n.b:} the colours for the ET and DG+ET actions are switched compared to our trajectory plots.}
    \label{fig:strnad}
\end{figure}

\begin{figure}[H]
    \centering
    \includegraphics[scale=0.65]{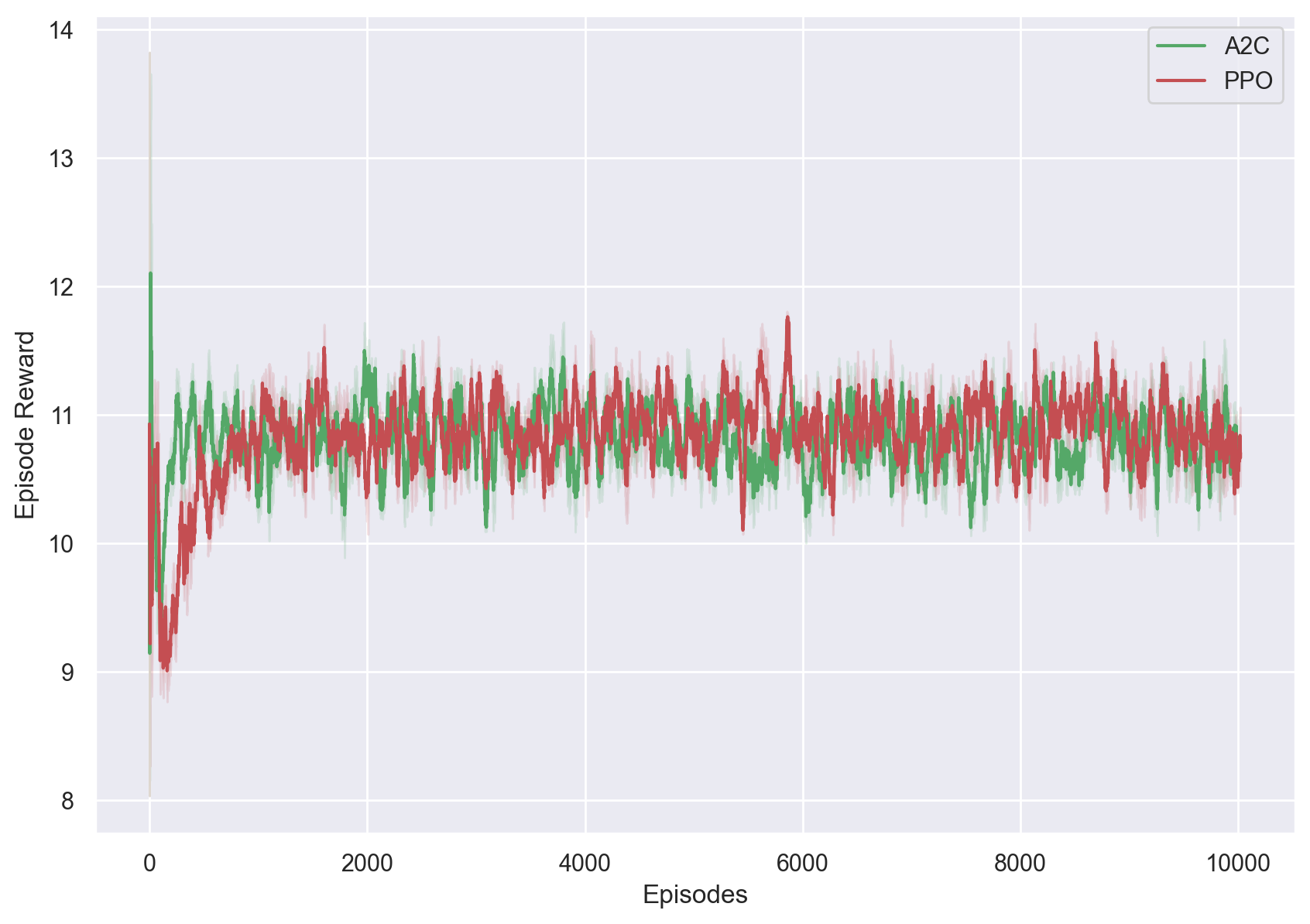}
    \caption{Moving average rewards of the two actor critic agents in the AYS environment, with the \textit{Policy Cost} reward over 500 000 frames. Plots show the mean and standard deviation over three different seeds.}
    \label{fig:app:ac extended}
\end{figure}

\begin{figure}
    \centering
    \includegraphics[scale=0.55]{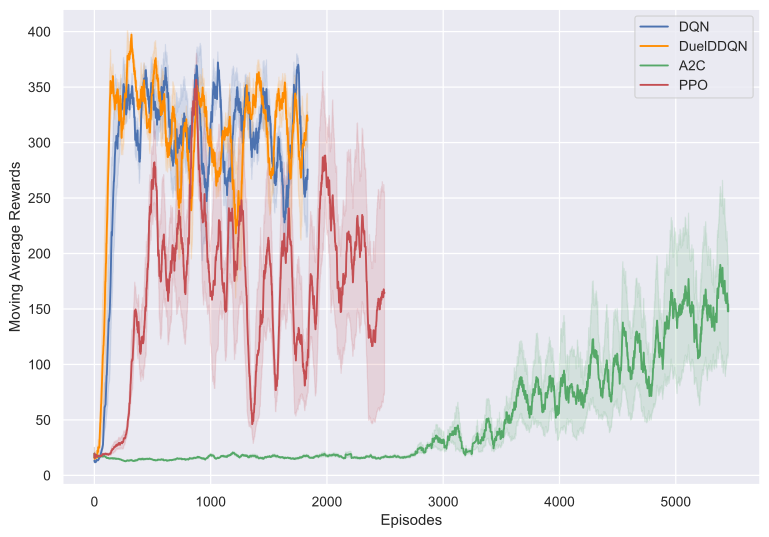}
    \caption{Moving average rewards of the agents in the noisy AYS environment with $\sigma_n^2=10^{-3}$, with the \textit{Planetary Boundaries} reward over 500 000 frames. Plots show the mean and standard deviation over three different seeds.}
    \label{fig:app:noisy learn}
\end{figure}

\chapter{Tables and Equations}\label{Tables}

The Lorenz system:
\begin{align}
    \frac{dx}{dt} &= \sigma(y-x),\\
    \frac{dy}{dt} &= x(\rho-z)-y,\\
    \frac{dz}{dt} &= xy-\beta z.
\end{align}

\begin{table}[H]
\centering
\begin{tabular}{|l|l|l|}
\hline
\textbf{Parameter} & \textbf{Value} & \textbf{Description} \\ \hline
\textbf{$\tau_A$} & 50 years & Carbon decay out of the atmosphere. \\ \hline
\textbf{$\tau_S$} & 50 years & Decay of renewable knowledge stock. \\ \hline
\textbf{$\beta$} & 3 \%/year & Economic output growth. \\ \hline
\textbf{$\sigma$} & $4\times 10^{12}$ GJ & \begin{tabular}[c]{@{}l@{}}Break-even knowledge: knowledge at which \\ fossil fuel and renewables have equal cost.\end{tabular} \\ \hline
\textbf{$\phi$} & $4.7\times 10^{10}$ GJ/GtC & Fossil fuel combustion efficiency. \\ \hline
\textbf{$\epsilon$} & 147 \$/GJ & Energy efficiency parameter. \\ \hline
\textbf{$\theta$} & $8.57\times10^{-5}$ & Temperature sensitivity parameter. \\ \hline
\textbf{$\rho$} & 2 & Learning rate of renewable knowledge. \\ \hline
\end{tabular}
\caption{Table summarising the parameters of the AYS model from Kittel et al. \cite{Kittel2017FromManagement}.}
\label{table:params}
\end{table}

\begin{table}[H]
\centering
\begin{tabular}{|p{4cm}|l|l|p{5cm}|}
\hline
\textbf{Parameter \newline (variable name)} & Distribution & Range & \textbf{Description} \\ \hline
Learning Rate (lr) & Log-Uniform & $10^{-5}-10^{-2}$ & Learning rate of the Network's Adam Optimiser.\\ \hline
Exploration decay ($\rho$) &Uniform & 0-1 & Controls the speed at which the policy converges to greedy.\\ \hline
Batch Size & Categorical & $2^{[4-11]}$ & Batch of experience given to the network at once.\\ \hline
Buffer Size & Categorical & $2^{[11-15]}$ & Maximum number of experiences stored.\\ \hline
Target Update frequency ($\tau$) & Log-Uniform & $10^{-4}-10^{-1}$ & Frequency of target network \newline parameter copying. \\ \hline
Learning rate decay \newline (decay number) & Categorical & 0-300 & Number of times the learning rate is reduced by half. \\ \hline
Buffer Prioritisation ($\alpha$) & Uniform & 0-1 & Controls the probability of \newline picking high-priority experience. \\ \hline
Importance Sampling\newline exponent ($\beta$) & Uniform & 0-1 & Controls the value of Importance sampling weights.\\ \hline 
Entropy regularisation\newline coefficient ($\epsilon$) & Log-Uniform &  $10^{-4}-0.2$ & Coefficient to the entropy term $\mathcal{H}(\pi)$.\\ \hline
Rollout length\newline (buffer size) & Categorical & $2^{[5-11]}$ & Length of experience accumulated for updating the networks\\ \hline
$\lambda$-return parameter ($\lambda$) & Uniform & 0-1 & Balances short and long-term reward. \\ \hline
Policy Clipping parameter (clip) & Uniform & 0-1 & Controls how far away the new policy can go from the previous policy.\\ \hline 
\end{tabular}
\caption{Distributions used to sample parameters for the Bayesian optimisation of the algorithms. }
\label{table:hparamsopt}
\end{table}

%% file: ref.bib
@article{Rockstrom2009AHumanity,
    title = {{A safe operating space for humanity}},
    year = {2009},
    journal = {Nature 2009 461:7263},
    author = {Rockstr{\"{o}}m, Johan and Steffen, Will and Noone, Kevin and Persson, Asa and Chapin, F. Stuart and Lambin, Eric F. and Lenton, Timothy M. and Scheffer, Marten and Folke, Carl and Schellnhuber, Hans Joachim and Nykvist, Björn and De Wit, Cynthia A. and Hughes, Terry and Van Der Leeuw, Sander and Rodhe, Henning and S{\"{o}}rlin, Sverker and Snyder, Peter K. and Costanza, Robert and Svedin, Uno and Falkenmark, Malin and Karlberg, Louise and Corell, Robert W. and Fabry, Victoria J. and Hansen, James and Walker, Brian and Liverman, Diana and Richardson, Katherine and Crutzen, Paul and Foley, Jonathan A.},
    number = {7263},
    month = {9},
    pages = {472--475},
    volume = {461},
    publisher = {Nature Publishing Group},
    url = {https://www.nature.com/articles/461472a},
    doi = {10.1038/461472a},
    issn = {1476-4687},
    pmid = {19779433}
}

@inproceedings{Lundberg2017APredictions,
    title = {{A Unified Approach to Interpreting Model Predictions}},
    year = {2017},
    booktitle = {Advances in Neural Information Processing Systems},
    author = {Lundberg, Scott M and Lee, Su-In},
    editor = {Guyon, I and Luxburg, U Von and Bengio, S and Wallach, H and Fergus, R and Vishwanathan, S and Garnett, R},
    pages = {},
    volume = {30},
    publisher = {Curran Associates, Inc.},
    url = {https://proceedings.neurips.cc/paper/2017/file/8a20a8621978632d76c43dfd28b67767-Paper.pdf}
}

@article{Kingma2014Adam:Optimization,
    title = {{Adam: A Method for Stochastic Optimization}},
    year = {2014},
    journal = {3rd International Conference on Learning Representations, ICLR 2015 - Conference Track Proceedings},
    author = {Kingma, Diederik P. and Ba, Jimmy Lei},
    month = {12},
    publisher = {International Conference on Learning Representations, ICLR},
    url = {https://arxiv.org/abs/1412.6980v9},
    doi = {10.48550/arxiv.1412.6980},
    arxivId = {1412.6980}
}

@article{Mnih2016AsynchronousLearning,
    title = {{Asynchronous Methods for Deep Reinforcement Learning}},
    year = {2016},
    journal = {33rd International Conference on Machine Learning, ICML 2016},
    author = {Mnih, Volodymyr and Badia, Adria Puigdomenech and Mirza, Lehdi and Graves, Alex and Harley, Tim and Lillicrap, Timothy P. and Silver, David and Kavukcuoglu, Koray},
    month = {2},
    pages = {2850--2869},
    volume = {4},
    publisher = {International Machine Learning Society (IMLS)},
    url = {https://arxiv.org/abs/1602.01783v2},
    isbn = {9781510829008},
    arxivId = {1602.01783}
}

@article{Bury2019ChartingModel,
    title = {{Charting pathways to climate change mitigation in a coupled socio-climate model}},
    year = {2019},
    journal = {PLOS Computational Biology},
    author = {Bury, Thomas M. and Bauchid, Chris T. and Anand, Madhur},
    number = {6},
    month = {6},
    pages = {e1007000},
    volume = {15},
    publisher = {Public Library of Science},
    url = {https://journals.plos.org/ploscompbiol/article?id=10.1371/journal.pcbi.1007000},
    isbn = {1111111111},
    doi = {10.1371/JOURNAL.PCBI.1007000},
    issn = {1553-7358},
    pmid = {31170149}
}

@article{Breiman2017ClassificationTrees,
    title = {{Classification and regression trees}},
    year = {2017},
    journal = {Classification and Regression Trees},
    author = {Breiman, Leo and Friedman, Jerome H. and Olshen, Richard A. and Stone, Charles J.},
    month = {1},
    pages = {1--358},
    publisher = {CRC Press},
    url = {https://www.taylorfrancis.com/books/mono/10.1201/9781315139470/classification-regression-trees-leo-breiman-jerome-friedman-richard-olshen-charles-stone},
    isbn = {9781351460491},
    doi = {10.1201/9781315139470/CLASSIFICATION-REGRESSION-TREES-LEO-BREIMAN-JEROME-FRIEDMAN-RICHARD-OLSHEN-CHARLES-STONE}
}

@article{IPCC2021ClimateSummary,
    title = {{Climate Change 2021: the Physical Science Basis, Technical Summary}},
    year = {2021},
    url={https://www.ipcc.ch/report/ar5/wg1/technical-summary/},
    journal={Climate Change 2022: Mitigation of Climate Change. Contribution of Working Group I to the Sixth Assessment Report of the Intergovernmental Panel on Climate Change},
    author = {{IPCC}},
    doi = {10.1017/9781009157896.002}
}

@article{Donges2017ClosingScience:,
    title = {{Closing the loop: Reconnecting human dynamics to Earth System science:}},
    year = {2017},
    journal = {https://doi.org/10.1177/2053019617725537},
    author = {Donges, Jonathan F. and Winkelmann, Ricarda and Lucht, Wolfgang and Cornell, Sarah E. and Dyke, James G. and Rockstr{\"{o}}m, Johan and Heitzig, Jobst and Schellnhuber, Hans Joachim},
    number = {2},
    month = {8},
    pages = {151--157},
    volume = {4},
    publisher = {SAGE PublicationsSage UK: London, England},
    url = {https://journals.sagepub.com/doi/10.1177/2053019617725537},
    doi = {10.1177/2053019617725537},
    issn = {2053020X}
}

@article{vanderWaa2018ContrastiveConsequences,
    title = {{Contrastive Explanations for Reinforcement Learning in terms of Expected Consequences}},
    year = {2018},
    author = {van der Waa, Jasper and van Diggelen, Jurriaan and Bosch, Karel van den and Neerincx, Mark},
    month = {7},
    url = {https://arxiv.org/abs/1807.08706v1},
    doi = {10.48550/arxiv.1807.08706},
    arxivId = {1807.08706}
}

@article{Lammers2020CorrectingDistancing,
    title = {{Correcting misperceptions of exponential coronavirus growth increases support for social distancing}},
    year = {2020},
    journal = {Proceedings of the National Academy of Sciences of the United States of America},
    author = {Lammers, Joris and Crusius, Jan and Gast, Anne},
    number = {28},
    month = {7},
    pages = {16264--16266},
    volume = {117},
    publisher = {National Academy of Sciences},
    url = {https://www.pnas.org/doi/abs/10.1073/pnas.2006048117},
    doi = {10.1073/PNAS.2006048117/SUPPL{\_}FILE/PNAS.2006048117.SAPP.PDF},
    issn = {10916490},
    pmid = {32581118}
}

@article{Vargas2019CreativityLearning,
    title = {{Creativity in Robot Manipulation with Deep Reinforcement Learning}},
    year = {2019},
    author = {Vargas, Juan Carlos and Bhoite, Malhar and Farimani, Amir Barati},
    month = {10},
    url = {https://arxiv.org/abs/1910.07459v1},
    doi = {10.48550/arxiv.1910.07459},
    arxivId = {1910.07459}
}

@book{Goodfellow2016DeepLearning,
    title = {{Deep Learning}},
    year = {2016},
    author = {Goodfellow, Ian and Bengio, Yoshua and Courville, Aaron},
    publisher = {MIT Press}
}

@article{Bury2021DeepPoints,
    title = {{Deep learning for early warning signals of tipping points}},
    year = {2021},
    journal = {Proceedings of the National Academy of Sciences of the United States of America},
    author = {Bury, Thomas M. and Sujith, R. I. and Pavithran, Induja and Scheffer, Marten and Lenton, Timothy M. and Anand, Madhur and Bauch, Chris T.},
    number = {39},
    month = {9},
    pages = {e2106140118},
    volume = {118},
    publisher = {National Academy of Sciences},
    url = {https://www.pnas.org/doi/abs/10.1073/pnas.2106140118},
    doi = {10.1073/PNAS.2106140118/SUPPL{\_}FILE/PNAS.2106140118.SAPP.PDF},
    issn = {10916490},
    pmid = {34544867}
}

@article{Strnad2019DeepStrategies,
    title = {{Deep reinforcement learning in World-Earth system models to discover sustainable management strategies}},
    year = {2019},
    journal = {Chaos: An Interdisciplinary Journal of Nonlinear Science},
    author = {Strnad, Felix M. and Barfuss, Wolfram and Donges, Jonathan F. and Heitzig, Jobst},
    number = {12},
    month = {12},
    pages = {123122},
    volume = {29},
    publisher = { AIP Publishing LLC AIP Publishing },
    url = {https://aip.scitation.org/doi/abs/10.1063/1.5124673},
    doi = {10.1063/1.5124673},
    issn = {1054-1500},
    arxivId = {1908.05567}
}

@article{VanHasselt2015DeepQ-learning,
    title = {{Deep Reinforcement Learning with Double Q-learning}},
    year = {2015},
    journal = {30th AAAI Conference on Artificial Intelligence, AAAI 2016},
    author = {Van Hasselt, Hado and Guez, Arthur and Silver, David},
    month = {9},
    pages = {2094--2100},
    publisher = {AAAI press},
    url = {https://arxiv.org/abs/1509.06461v3},
    isbn = {9781577357605},
    doi = {10.48550/arxiv.1509.06461},
    issn = {2159-5399},
    arxivId = {1509.06461}
}

@misc{VanHasselt2021DeepMindYouTube,
    title = {{DeepMind x UCL RL Lecture Series - Introduction to Reinforcement Learning - YouTube}},
    year = {2021},
    author = {Van Hasselt, Hado and Borsa, Diana and Hessel, Matteo},
    url = {https://www.youtube.com/watch?v=TCCjZe0y4Qc&}
}

@article{Moore2022DeterminantsSystem,
    title = {{Determinants of emissions pathways in the coupled climate–social system}},
    year = {2022},
    journal = {Nature 2022 603:7899},
    author = {Moore, Frances C. and Lacasse, Katherine and Mach, Katharine J. and Shin, Yoon Ah and Gross, Louis J. and Beckage, Brian},
    number = {7899},
    month = {2},
    pages = {103--111},
    volume = {603},
    publisher = {Nature Publishing Group},
    url = {https://www.nature.com/articles/s41586-022-04423-8},
    doi = {10.1038/s41586-022-04423-8},
    issn = {1476-4687},
    pmid = {35173331}
}

@misc{EdwardN.Lorenz1963DeterministicFlow,
    title = {{Deterministic Nonperiodic Flow}},
    year = {1963},
    booktitle = {Journal of the Atmospheric Sciences Volume 20 Issue 2},
    author = {{Edward N. Lorenz}},
    url = {https://journals.ametsoc.org/view/journals/atsc/20/2/1520-0469_1963_020_0130_dnf_2_0_co_2.xml?tab_body=pdf},
    doi = {https://doi.org/10.1175/1520-0469(1963)020<0130:DNF>2.0.CO;2}
}

@inproceedings{VanHasselt2010DoubleQ-learning,
    title = {{Double Q-learning}},
    year = {2010},
    booktitle = {Advances in Neural Information Processing Systems},
    author = {Van Hasselt, Hado},
    editor = {Lafferty, J and Williams, C and Shawe-Taylor, J and Zemel, R and Culotta, A},
    pages = {},
    volume = {23},
    publisher = {Curran Associates, Inc.},
    url = {https://proceedings.neurips.cc/paper/2010/file/091d584fced301b442654dd8c23b3fc9-Paper.pdf}
}

@article{Wang2015DuelingLearning,
    title = {{Dueling Network Architectures for Deep Reinforcement Learning}},
    year = {2015},
    journal = {33rd International Conference on Machine Learning, ICML 2016},
    author = {Wang, Ziyu and Schaul, Tom and Hessel, Matteo and Van Hasselt, Hado and Lanctot, Marc and De Frcitas, Nando},
    month = {11},
    pages = {2939--2947},
    volume = {4},
    publisher = {International Machine Learning Society (IMLS)},
    url = {https://arxiv.org/abs/1511.06581v3},
    isbn = {9781510829008},
    doi = {10.48550/arxiv.1511.06581},
    arxivId = {1511.06581}
}

@article{Kamien2012DynamicManagement,
    title = {{Dynamic Optimization, Second Edition: The Calculus of Variations and Optimal Control in Economics and Management}},
    year = {2012},
    author = {Kamien, Morton I and Schwartz, Nancy L},
    publisher = {Dover Publications},
    url = {https://books.google.com/books/about/Dynamic_Optimization.html?id=0IoGUn8wjDQC},
    isbn = {048648856X}
}

@article{Bellman1957DynamicStudy,
    title = {{Dynamic Programming: A Rand Corporation Research Study}},
    year = {1957},
    author = {Bellman, Richard},
    pages = {363},
    publisher = {Princeton Unversity Press},
    isbn = {9780691079516}
}

@inproceedings{Juozapaitis2019ExplainableDecomposition,
    title = {{Explainable Reinforcement Learning via Reward Decomposition}},
    year = {2019},
    author = {Juozapaitis, Zoe and Koul, Anurag and Fern, Alan and Erwig, Martin and Doshi-Velez, Finale},
    url = {https://web.engr.oregonstate.edu/~erwig/papers/ExplainableRL_XAI19.pdf}
}

@article{Levy2017Exponential-growthOverconfidence,
    title = {{Exponential-growth bias and overconfidence}},
    year = {2017},
    journal = {Journal of Economic Psychology},
    author = {Levy, Matthew R. and Tasoff, Joshua},
    month = {2},
    pages = {1--14},
    volume = {58},
    publisher = {North-Holland},
    doi = {10.1016/J.JOEP.2016.11.001},
    issn = {0167-4870}
}

@article{Banerjee2021Exponential-growthCOVID-19,
    title = {{Exponential-growth prediction bias and compliance with safety measures related to COVID-19}},
    year = {2021},
    journal = {Social science {\&} medicine (1982)},
    author = {Banerjee, Ritwik and Bhattacharya, Joydeep and Majumdar, Priyama},
    month = {1},
    volume = {268},
    publisher = {Soc Sci Med},
    url = {https://pubmed.ncbi.nlm.nih.gov/33130402/},
    doi = {10.1016/J.SOCSCIMED.2020.113473},
    issn = {1873-5347},
    pmid = {33130402}
}

@article{Walker2018FormalStages/subseries,
    title = {{Formal ratification of the subdivision of the Holocene Series/Epoch (Quaternary System/Period): two new Global Boundary Stratotype Sections and Points (GSSPs) and three new stages/subseries}},
    year = {2018},
    journal = {International Union of Geological Sciences},
    author = {Walker, Mike and Head, Martin J and Berkelhammer, Max and Bjorck, Svante and Cheng, Hai and Cwynar, Les and Fisher, David and Gkinis, Vasilios and Long, Antony and Lowe, John and Newnham, Rewi and Rasmussen, Sune Olander and Weiss, Harvey},
    edition = {2018/12/01},
    number = {4},
    month = {12},
    pages = {213--223},
    volume = {41},
    publisher = {International Union of Geological Sciences},
    url = {http://episodes.org/journal/view.html?doi=10.18814/epiiugs/2018/018016},
    doi = {10.18814/epiiugs/2018/018016},
    issn = {0705-3797},
    language = {eng}
}

@article{Kittel2017FromManagement,
    title = {{From lakes and glades to viability algorithms: Automatic classification of system states according to the Topology of Sustainable Management}},
    year = {2017},
    journal = {European Physical Journal: Special Topics},
    author = {Kittel, Tim and M{\"{u}}ller-Hansen, Finn and Koch, Rebekka and Heitzig, Jobst and Deffuant, Guillaume and Mathias, Jean Denis and Kurths, Jürgen},
    number = {14-15},
    month = {6},
    pages = {3133--3152},
    volume = {230},
    publisher = {Springer Science and Business Media Deutschland GmbH},
    url = {https://arxiv.org/abs/1706.04542v4},
    doi = {10.48550/arxiv.1706.04542},
    issn = {19516401},
    arxivId = {1706.04542}
}

@article{Crutzen2002GeologyMankind,
    title = {{Geology of mankind}},
    year = {2002},
    journal = {Nature 2002 415:6867},
    author = {Crutzen, Paul J.},
    number = {6867},
    month = {1},
    pages = {23--23},
    volume = {415},
    publisher = {Nature Publishing Group},
    url = {https://www.nature.com/articles/415023a},
    doi = {10.1038/415023a},
    issn = {1476-4687},
    pmid = {11780095}
}

@article{Schulman2015High-DimensionalEstimation,
    title = {{High-Dimensional Continuous Control Using Generalized Advantage Estimation}},
    year = {2015},
    journal = {4th International Conference on Learning Representations, ICLR 2016 - Conference Track Proceedings},
    author = {Schulman, John and Moritz, Philipp and Levine, Sergey and Jordan, Michael I. and Abbeel, Pieter},
    month = {6},
    publisher = {International Conference on Learning Representations, ICLR},
    url = {https://arxiv.org/abs/1506.02438v6},
    isbn = {1506.02438v6},
    doi = {10.48550/arxiv.1506.02438},
    arxivId = {1506.02438}
}

@article{Mnih2015Human-levelLearning,
    title = {{Human-level control through deep reinforcement learning}},
    year = {2015},
    journal = {Nature 2015 518:7540},
    author = {Mnih, Volodymyr and Kavukcuoglu, Koray and Silver, David and Rusu, Andrei A. and Veness, Joel and Bellemare, Marc G. and Graves, Alex and Riedmiller, Martin and Fidjeland, Andreas K. and Ostrovski, Georg and Petersen, Stig and Beattie, Charles and Sadik, Amir and Antonoglou, Ioannis and King, Helen and Kumaran, Dharshan and Wierstra, Daan and Legg, Shane and Hassabis, Demis},
    number = {7540},
    month = {2},
    pages = {529--533},
    volume = {518},
    publisher = {Nature Publishing Group},
    url = {https://www.nature.com/articles/nature14236},
    doi = {10.1038/nature14236},
    issn = {1476-4687},
    pmid = {25719670}
}

@inproceedings{Krizhevsky2012ImageNetNetworks,
    title = {{ImageNet Classification with Deep Convolutional Neural Networks}},
    year = {2012},
    booktitle = {Advances in Neural Information Processing Systems},
    author = {Krizhevsky, Alex and Sutskever, Ilya and Hinton, Geoffrey E},
    editor = {Pereira, F and Burges, C J and Bottou, L and Weinberger, K Q},
    pages = {},
    volume = {25},
    publisher = {Curran Associates, Inc.},
    url = {https://proceedings.neurips.cc/paper/2012/file/c399862d3b9d6b76c8436e924a68c45b-Paper.pdf}
}

@article{Wang2017IntegratedEconomics,
    title = {{Integrated assessment models of climate change economics}},
    year = {2017},
    journal = {Integrated Assessment Models of Climate Change Economics},
    author = {Wang, Zheng and Wu, Jing and Liu, Changxin and Gu, Gaoxiang},
    month = {1},
    pages = {1--181},
    publisher = {Springer Singapore},
    isbn = {9789811039454},
    doi = {10.1007/978-981-10-3945-4/COVER}
}

@article{Schonger2021IntuitionComplexity,
    title = {{Intuition and exponential growth: bias and the roles of parameterization and complexity}},
    year = {2021},
    journal = {Mathematische Semesterberichte},
    author = {Schonger, Martin and Sele, Daniela},
    number = {2},
    month = {10},
    pages = {221},
    volume = {68},
    publisher = {Springer},
    url = {/pmc/articles/PMC8386158/ /pmc/articles/PMC8386158/?report=abstract https://www.ncbi.nlm.nih.gov/pmc/articles/PMC8386158/},
    doi = {10.1007/S00591-021-00306-7},
    issn = {14321815},
    pmid = {34795462}
}

@article{Marot2020L2RPN:Design,
  author    = {Antoine Marot and
               Benjamin Donnot and
               Gabriel Dulac{-}Arnold and
               Adrian Kelly and
               A{\"{\i}}dan O'Sullivan and
               Jan Viebahn and
               Mariette Awad and
               Isabelle Guyon and
               Patrick Panciatici and
               Camilo Romero},
  title     = {Learning to run a Power Network Challenge: a Retrospective Analysis},
  journal   = {CoRR},
  volume    = {abs/2103.03104},
  year      = {2021},
  url       = {https://arxiv.org/abs/2103.03104},
  eprinttype = {arXiv},
  eprint    = {2103.03104},
  bibsource = {dblp computer science bibliography, https://dblp.org}
}

@article{Watkins1989LearningRewards,
    title = {{Learning from delayed rewards}},
    year = {1989},
    author = {Watkins, Christopher J. C. H.},
    publisher = {King's College, Cambridge United Kingdom}
}

@article{Berkes1998LinkingSustainability,
    title = {{Linking social and ecological systems for resilience and sustainability}},
    year = {1998},
    journal = {Linking Social and Ecological Systems},
    author = {Berkes, Fikret and Folke, Carl},
    editor = {Berkes, Fikret and Folke, Carl},
    pages = {13--20},
    volume = {1},
    publisher = {Cambridge University Press},
    url = {https://books.google.com/books/about/Linking_Social_and_Ecological_Systems.html?id=XixuNvX2zLwC},
    address = {Cambridge, UK},
    isbn = {0521785626},
    issn = {1102-4941},
    pmid = {21679024}
}

@article{Degrave2022MagneticLearning,
    title = {{Magnetic control of tokamak plasmas through deep reinforcement learning}},
    year = {2022},
    journal = {Nature 2022 602:7897},
    author = {Degrave, Jonas and Felici, Federico and Buchli, Jonas and Neunert, Michael and Tracey, Brendan and Carpanese, Francesco and Ewalds, Timo and Hafner, Roland and Abdolmaleki, Abbas and de las Casas, Diego and Donner, Craig and Fritz, Leslie and Galperti, Cristian and Huber, Andrea and Keeling, James and Tsimpoukelli, Maria and Kay, Jackie and Merle, Antoine and Moret, Jean Marc and Noury, Seb and Pesamosca, Federico and Pfau, David and Sauter, Olivier and Sommariva, Cristian and Coda, Stefano and Duval, Basil and Fasoli, Ambrogio and Kohli, Pushmeet and Kavukcuoglu, Koray and Hassabis, Demis and Riedmiller, Martin},
    number = {7897},
    month = {2},
    pages = {414--419},
    volume = {602},
    publisher = {Nature Publishing Group},
    url = {https://www.nature.com/articles/s41586-021-04301-9},
    doi = {10.1038/s41586-021-04301-9},
    issn = {1476-4687},
    pmid = {35173339}
}

@article{Silver2016MasteringSearch,
    title = {{Mastering the game of Go with deep neural networks and tree search}},
    year = {2016},
    journal = {Nature 2016 529:7587},
    author = {Silver, David and Huang, Aja and Maddison, Chris J. and Guez, Arthur and Sifre, Laurent and Van Den Driessche, George and Schrittwieser, Julian and Antonoglou, Ioannis and Panneershelvam, Veda and Lanctot, Marc and Dieleman, Sander and Grewe, Dominik and Nham, John and Kalchbrenner, Nal and Sutskever, Ilya and Lillicrap, Timothy and Leach, Madeleine and Kavukcuoglu, Koray and Graepel, Thore and Hassabis, Demis},
    number = {7587},
    month = {1},
    pages = {484--489},
    volume = {529},
    publisher = {Nature Publishing Group},
    url = {https://www.nature.com/articles/nature16961},
    doi = {10.1038/nature16961},
    issn = {1476-4687},
    pmid = {26819042}
}

@article{Plappert2017ParameterExploration,
    title = {{Parameter Space Noise for Exploration}},
    year = {2017},
    journal = {6th International Conference on Learning Representations, ICLR 2018 - Conference Track Proceedings},
    author = {Plappert, Matthias and Houthooft, Rein and Dhariwal, Prafulla and Sidor, Szymon and Chen, Richard Y. and Chen, Xi and Asfour, Tamim and Abbeel, Pieter and Andrychowicz, Marcin},
    month = {6},
    publisher = {International Conference on Learning Representations, ICLR},
    url = {https://arxiv.org/abs/1706.01905v2},
    doi = {10.48550/arxiv.1706.01905},
    arxivId = {1706.01905}
}

@article{Steffen2015PlanetaryPlanet,
    title = {{Planetary boundaries: Guiding human development on a changing planet}},
    year = {2015},
    journal = {Science},
    author = {Steffen, Will and Richardson, Katherine and Rockstr{\"{o}}m, Johan and Cornell, Sarah E. and Fetzer, Ingo and Bennett, Elena M. and Biggs, Reinette and Carpenter, Stephen R. and De Vries, Wim and De Wit, Cynthia A. and Folke, Carl and Gerten, Dieter and Heinke, Jens and Mace, Georgina M. and Persson, Linn M. and Ramanathan, Veerabhadran and Reyers, Belinda and S{\"{o}}rlin, Sverker},
    number = {6223},
    month = {2},
    volume = {347},
    publisher = {American Association for the Advancement of Science},
    url = {http://dx.doi.},
    doi = {10.1126/SCIENCE.1259855},
    issn = {10959203},
    pmid = {25592418}
}

@inproceedings{Sutton1999PolicyApproximation,
    title = {{Policy Gradient Methods for Reinforcement Learning with Function Approximation}},
    year = {1999},
    booktitle = {Advances in Neural Information Processing Systems},
    author = {Sutton, Richard S and McAllester, David and Singh, Satinder and Mansour, Yishay},
    editor = {Solla, S and Leen, T and M{\"{u}}ller, K},
    pages = {},
    volume = {12},
    publisher = {MIT Press},
    url = {https://proceedings.neurips.cc/paper/1999/file/464d828b85b0bed98e80ade0a5c43b0f-Paper.pdf}
}

@article{Mansfield2020PredictingLearning,
    title = {{Predicting global patterns of long-term climate change from short-term simulations using machine learning}},
    year = {2020},
    journal = {npj Climate and Atmospheric Science 2020 3:1},
    author = {Mansfield, L. A. and Nowack, P. J. and Kasoar, M. and Everitt, R. G. and Collins, W. J. and Voulgarakis, A.},
    number = {1},
    month = {11},
    pages = {1--9},
    volume = {3},
    publisher = {Nature Publishing Group},
    url = {https://www.nature.com/articles/s41612-020-00148-5},
    doi = {10.1038/s41612-020-00148-5},
    issn = {2397-3722}
}

@article{Schaul2015PrioritizedReplay,
    title = {{Prioritized Experience Replay}},
    year = {2015},
    journal = {4th International Conference on Learning Representations, ICLR 2016 - Conference Track Proceedings},
    author = {Schaul, Tom and Quan, John and Antonoglou, Ioannis and Silver, David},
    month = {11},
    publisher = {International Conference on Learning Representations, ICLR},
    url = {https://arxiv.org/abs/1511.05952v4},
    isbn = {1511.05952v4},
    arxivId = {1511.05952}
}

@article{Schulman2017ProximalAlgorithms,
  title={Proximal Policy Optimization Algorithms},
  author={John Schulman and Filip Wolski and Prafulla Dhariwal and Alec Radford and Oleg Klimov},
  journal={ArXiv},
  year={2017},
  volume={abs/1707.06347}
}

@article{Watkins1992Q-learning,
    title = {{Q-learning}},
    year = {1992},
    journal = {Machine Learning 1992 8:3},
    author = {Watkins, Christopher J. C. H. and Dayan, Peter},
    number = {3},
    month = {5},
    pages = {279--292},
    volume = {8},
    publisher = {Springer},
    url = {https://link.springer.com/article/10.1007/BF00992698},
    doi = {10.1007/BF00992698},
    issn = {1573-0565}
}

@article{Hessel2017Rainbow:Learning,
    title = {{Rainbow: Combining Improvements in Deep Reinforcement Learning}},
    year = {2017},
    journal = {32nd AAAI Conference on Artificial Intelligence, AAAI 2018},
    author = {Hessel, Matteo and Modayil, Joseph and Van Hasselt, Hado and Schaul, Tom and Ostrovski, Georg and Dabney, Will and Horgan, Dan and Piot, Bilal and Azar, Mohammad and Silver, David},
    month = {10},
    pages = {3215--3222},
    publisher = {AAAI press},
    url = {https://arxiv.org/abs/1710.02298v1},
    isbn = {9781577358008},
    doi = {10.48550/arxiv.1710.02298},
    issn = {2159-5399},
    arxivId = {1710.02298}
}

@article{Lu2021ReinforcementSensing,
    title = {{Reinforcement Learning for Adaptive Video Compressive Sensing}},
    year = {2021},
    author = {Lu, Sidi and Yuan, Xin and Katsaggelos, Aggelos K and Shi, Weisong},
    month = {5},
    url = {https://arxiv.org/abs/2105.08205v1},
    doi = {10.48550/arxiv.2105.08205},
    arxivId = {2105.08205}
}

@book{Sutton2020ReinforcementIntroduction,
    title = {{Reinforcement Learning: An Introduction}},
    year = {2020},
    author = {Sutton, Richard S and Barto, Andrew G},
    edition = {2},
    url = {http://incompleteideas.net/book/the-book.html}
}

@article{Williams1992SimpleLearning,
    title = {{Simple statistical gradient-following algorithms for connectionist reinforcement learning}},
    year = {1992},
    journal = {Machine Learning 1992 8:3},
    author = {Williams, Ronald J.},
    number = {3},
    month = {5},
    pages = {229--256},
    volume = {8},
    publisher = {Springer},
    url = {https://link.springer.com/article/10.1007/BF00992696},
    doi = {10.1007/BF00992696},
    issn = {1573-0565}
}

@article{Levin2013Social-ecologicalImplications,
    title = {{Social-ecological systems as complex adaptive systems: modeling and policy implications}},
    year = {2013},
    journal = {Environment and Development Economics},
    author = {Levin, Simon and Xepapadeas, Tasos and Cr{\'{e}}pin, Anne Sophie and Norberg, Jon and De Zeeuw, Aart and Folke, Carl and Hughes, Terry and Arrow, Kenneth and Barrett, Scott and Daily, Gretchen and Ehrlich, Paul and Kautsky, Nils and M{\"{a}}ler, Karl Göran and Polasky, Steve and Troell, Max and Vincent, Jeffrey R. and Walker, Brian},
    number = {2},
    month = {4},
    pages = {111--132},
    volume = {18},
    publisher = {Cambridge University Press},
    url = {https://www.cambridge.org/core/journals/environment-and-development-economics/article/socialecological-systems-as-complex-adaptive-systems-modeling-and-policy-implications/C02DE8F7767B295C3289F51E83D845B4},
    doi = {10.1017/S1355770X12000460},
    issn = {1355-770X}
}

@techreport{IPCC2021SummaryPolicymakers,
    title = {{Summary for Policymakers}},
    year = {2022},
    journal={Climate Change 2022: Mitigation of Climate Change. Contribution of Working Group III to the Sixth Assessment Report of the Intergovernmental Panel on Climate Change},
    author = {{IPCC}},
    url={https://www.ipcc.ch/srccl/chapter/summary-for-policymakers/},
    doi = {10.1017/9781009157896.001}
}

@article{Nitzbon2017SustainabilityModel,
    title = {{Sustainability, collapse and oscillations in a simple World-Earth model}},
    year = {2017},
    journal = {Environmental Research Letters},
    author = {Nitzbon, Jan and Heitzig, Jobst and Parlitz, Ulrich},
    number = {7},
    month = {7},
    pages = {074020},
    volume = {12},
    publisher = {IOP Publishing},
    url = {https://iopscience.iop.org/article/10.1088/1748-9326/aa7581 https://iopscience.iop.org/article/10.1088/1748-9326/aa7581/meta},
    doi = {10.1088/1748-9326/AA7581},
    issn = {1748-9326}
}

@article{Rolnick2019TacklingLearning,
    title = {{Tackling Climate Change with Machine Learning}},
    year = {2019},
    journal = {ACM Computing Surveys},
    author = {Rolnick, David and Donti, Priya L. and Kaack, Lynn H. and Kochanski, Kelly and Lacoste, Alexandre and Sankaran, Kris and Ross, Andrew Slavin and Milojevic-Dupont, Nikola and Jaques, Natasha and Waldman-Brown, Anna and Luccioni, Alexandra Sasha and Maharaj, Tegan and Sherwin, Evan D. and Mukkavilli, S. Karthik and Kording, Konrad P. and Gomes, Carla P. and Ng, Andrew Y. and Hassabis, Demis and Platt, John C. and Creutzig, Felix and Chayes, Jennifer and Bengio, Yoshua},
    number = {2},
    month = {6},
    volume = {55},
    publisher = {Association for Computing Machinery},
    url = {https://arxiv.org/abs/1906.05433v1},
    doi = {10.48550/arxiv.1906.05433},
    issn = {15577341},
    arxivId = {1906.05433}
}

@article{Watkins1992TechnicalQ-Learning,
    title = {{Technical Note Q,-Learning}},
    year = {1992},
    author = {Watkins, Christopher J. C. H. and Dayan, Peter},
    pages = {279--292},
    volume = {8}
}

@inproceedings{Huang2022TheOptimization,
    title = {{The 37 Implementation Details of Proximal Policy Optimization}},
    year = {2022},
    booktitle = {ICLR Blog Track},
    author = {Huang, Shengyi and Dossa, Rousslan Fernand Julien and Raffin, Antonin and Kanervisto, Anssi and Wang, Weixun},
    url = {https://iclr-blog-track.github.io/2022/03/25/ppo-implementation-details/}
}

@article{Waters2016TheHolocene,
    title = {{The Anthropocene is functionally and stratigraphically distinct from the Holocene}},
    year = {2016},
    journal = {Science},
    author = {Waters, Colin N. and Zalasiewicz, Jan and Summerhayes, Colin and Barnosky, Anthony D. and Poirier, Clément and Ga{\l}uszka, Agnieszka and Cearreta, Alejandro and Edgeworth, Matt and Ellis, Erle C. and Ellis, Michael and Jeandel, Catherine and Leinfelder, Reinhold and McNeill, J. R. and Richter, Daniel De B. and Steffen, Will and Syvitski, James and Vidas, Davor and Wagreich, Michael and Williams, Mark and Zhisheng, An and Grinevald, Jacques and Odada, Eric and Oreskes, Naomi and Wolfe, Alexander P.},
    number = {6269},
    month = {1},
    volume = {351},
    publisher = {American Association for the Advancement of Science},
    url = {https://www.science.org/doi/10.1126/science.aad2622},
    doi = {10.1126/SCIENCE.AAD2622/ASSET/FD0FB267-5A01-40AD-8E8D-EAF5DC632D27/ASSETS/GRAPHIC/351{\_}AAD2622{\_}F7.JPEG},
    issn = {10959203},
    pmid = {26744408}
}

@article{Lenton2008TippingSystem,
    title = {{Tipping elements in the Earth's climate system}},
    year = {2008},
    journal = {Proceedings of the national Academy of Sciences},
    author = {Lenton, Timothy M and Held, Hermann and Kriegler, Elmar and Hall, Jim W and Lucht, Wolfgang and Rahmstorf, Stefan and Schellnhuber, Hans Joachim},
    number = {6},
    pages = {1786--1793},
    volume = {105},
    publisher = {National Acad Sciences}
}

@article{Steffen2018TrajectoriesAnthropocene,
    title = {{Trajectories of the Earth System in the Anthropocene}},
    year = {2018},
    journal = {Proceedings of the National Academy of Sciences of the United States of America},
    author = {Steffen, Will and Rockstr{\"{o}}m, Johan and Richardson, Katherine and Lenton, Timothy M. and Folke, Carl and Liverman, Diana and Summerhayes, Colin P. and Barnosky, Anthony D. and Cornell, Sarah E. and Crucifix, Michel and Donges, Jonathan F. and Fetzer, Ingo and Lade, Steven J. and Scheffer, Marten and Winkelmann, Ricarda and Schellnhuber, Hans Joachim},
    number = {33},
    month = {8},
    pages = {8252--8259},
    volume = {115},
    publisher = {National Academy of Sciences},
    url = {https://www.pnas.org/doi/abs/10.1073/pnas.1810141115},
    doi = {10.1073/PNAS.1810141115/SUPPL{\_}FILE/PNAS.1810141115.SAPP.PDF},
    issn = {10916490},
    pmid = {30082409}
}

@article{Andrychowicz2020WhatStudy,
    title = {{What Matters In On-Policy Reinforcement Learning? A Large-Scale Empirical Study}},
    year = {2020},
    author = {Andrychowicz, Marcin and Raichuk, Anton and Sta{\'{n}}czyk, Piotr and Orsini, Manu and Girgin, Sertan and Marinier, Raphael and Hussenot, Léonard and Geist, Matthieu and Pietquin, Olivier and Michalski, Marcin and Gelly, Sylvain and Bachem, Olivier},
    month = {6},
    url = {https://arxiv.org/abs/2006.05990v1},
    doi = {10.48550/arxiv.2006.05990},
    arxivId = {2006.05990}
}

@article{Menard2021WhenInequality,
    title = {{When conflicts get heated, so does the planet: coupled social-climate dynamics under inequality}},
    year = {2021},
    journal = {Proceedings of the Royal Society B},
    author = {Menard, Jyler and Bury, Thomas M and Bauch, Chris T and Anand, Madhur},
    number = {1959},
    volume = {288},
    publisher = {The Royal Society},
    url = {https://royalsocietypublishing.org/doi/10.1098/rspb.2021.1357},
    doi = {10.1098/RSPB.2021.1357},
    issn = {14712954},
    pmid = {34521252}
}

@article{580874,
  author={Tsitsiklis, J.N. and Van Roy, B.},
  journal={IEEE Transactions on Automatic Control}, 
  title={An analysis of temporal-difference learning with function approximation}, 
  year={1997},
  volume={42},
  number={5},
  pages={674-690},
  doi={10.1109/9.580874}
  }

@book{meadows1972limits,
  title={Limits to Growth},
  author={Meadows, D.H.},
  isbn={9780451066176},
  url={https://books.google.co.uk/books?id=VVyePQAACAAJ},
  year={1972},
  publisher={Penguin Group (USA) Incorporated}
}

@article{Tollefson2021CarbonDip,
    title = {{Carbon emissions rapidly rebounded following COVID pandemic dip}},
    year = {2021},
    journal = {Nature},
    author = {Tollefson, Jeff},
    month = {11},
    publisher = {Springer Science and Business Media LLC},
    doi = {10.1038/D41586-021-03036-X},
    issn = {0028-0836}
}

@misc{ESM,
    title={GFDL Earth-System Models},
    author={GFDL},
    url={https://www.gfdl.noaa.gov/earth-system-models/}
}

@article{MOTESHARREI201490,
title = {Human and nature dynamics (HANDY): Modeling inequality and use of resources in the collapse or sustainability of societies},
journal = {Ecological Economics},
volume = {101},
pages = {90-102},
year = {2014},
issn = {0921-8009},
doi = {https://doi.org/10.1016/j.ecolecon.2014.02.014},
url = {https://www.sciencedirect.com/science/article/pii/S0921800914000615},
author = {Safa Motesharrei and Jorge Rivas and Eugenia Kalnay}
}
